\documentclass[preprint,12pt]{elsarticle}
\usepackage{float}
\usepackage{graphicx}
\usepackage{soul, color}
\usepackage[acronym]{glossaries}
\usepackage{adjustbox}
\usepackage{booktabs}
\usepackage{amssymb}
\usepackage{tikz}
\usepackage{subcaption}
\usepackage{pdflscape}
\usepackage{comment}
\usepackage{multirow}
\usepackage{enumitem}

\usetikzlibrary{positioning, arrows.meta, backgrounds, fit, calc}
\newacronym{gnn}{GNN}{Graph Neural Network}
\newacronym{pf}{PF}{Power Flow}
\newacronym{opf}{OPF}{Optimal Power Flow}
\newacronym{se}{SE}{State Estimation}
\newacronym{ml}{ML}{Machine Learning}
\newacronym{mpnn}{MPNN}{Message Passing Neural Network}
\newacronym{gat}{GAT}{Graph Attention Network}
\newacronym{ac}{AC}{Alternating Current}
\newacronym{dc}{DC}{Direct Current}
\newacronym{aspl}{ASPL}{Average Shortest Path Length}
\newacronym{eda}{EDA}{Exploratory Data Analysis}
\newacronym{pyg}{PyG}{PyTorch Geometric}
\newacronym{gcn}{GCN}{Graph Convolutional Network}
\newacronym{gin}{GIN}{Graph Isomorphism Network}
\newacronym{gine}{GINE}{Graph Isomorphism Network with Edge Features}
\newacronym{resgated}{ResGated}{Residual Gated Graph Convolutional Network}
\newacronym{sage}{GraphSAGE}{Graph SAmple and aggreGatE}
\newacronym{gsp}{GSP}{Graph Signal Processing}
\newacronym{effrank}{ERank}{Effective Rank}
\newacronym{dnns}{DNNs}{Deep Neural Networks}
\newacronym{gnns}{GNNs}{Graph Neural Networks}
\newacronym{gru}{GRU}{Gated Recurrent Unit}
\newacronym{ipm}{IPM}{Interior Point Methods}
\newacronym{kcl}{KCL}{Kirchhoff's Current Law}
\newacronym{kclmae}{KCL-MAE}{Mean Absolute Power Imbalance}
\newacronym{kclmax}{KCL-Max}{Maximum Power Imbalance}
\newacronym{kvl}{KVL}{Kirchhoff's Voltage Law}
\newacronym{mlp}{MLP}{Multi-Layer Perceptron}
\newacronym{grnn}{GRNN}{Graph Recurrent Neural Network}
\newacronym{mse}{MSE}{Mean Squared Error}
\newacronym{mva}{MVA}{Megavolt-Ampere}
\newacronym{nr}{NR}{Newton--Raphson}
\newacronym{acopf}{AC-OPF}{AC Optimal Power Flow}
\newacronym{sqp}{SQP}{Sequential Quadratic Programming}
\newacronym{mmd}{MMD}{Maximum Mean Discrepancy}
\newacronym{acpf}{AC-PF}{AC Power Flow}
\newacronym{dcpf}{DC-PF}{DC Power Flow}
\newacronym{pu}{p.u.}{Per-unit}
\newacronym{relu}{ReLU}{Rectified Linear Unit}
\newacronym{silu}{SiLU}{Sigmoid Linear Unit}
\newacronym{trmae}{TRMAE}{Thresholded Relative Mean Absolute Error}
\newacronym{wls}{WLS}{Weighted Least Squares}
\newacronym{qkv}{QKV}{Query-Key-Value}
\newacronym{maxae}{MaxAE}{Maximum Absolute Error}
\newacronym{nn}{NN}{Neural Network}

\begin{document}

\title{Benchmarking Graph Neural Networks for AC Power Flow: The Role of Inductive Biases and Physical Constraints}

\tnotetext[funding]{%
This work was supported by the HP2C-DT project (European Union NextGenerationEU/PRTR, TED2021-130351B-C21), the BLOSSOMS project (grant PID2024-158530OB-I00, funded by MICIU/AEI/10.13039/501100011033 and ERDF/EU), and the GRAPHS4SEC project
(grant PCI2023-145974-2, funded by MICIU/AEI/10.13039/501100011033). This work was also partially funded by ICREA.
}

\author[upc,padua]{Federico Clerici}
\author[citcea]{Ferran Bohigas-Daranas}
\author[bnn]{Hamid Latif-Martínez}
\author[citcea]{Eduardo Prieto-Araujo}
\author[citcea]{Oriol Gomis-Bellmunt}
\author[bnn]{Pere Barlet-Ros}

\address[upc]{Universitat Politècnica de Catalunya \textperiodcentered\ BarcelonaTech (UPC), Barcelona, Spain}

\address[padua]{Department of Mathematics, University of Padua, Padua, Italy}

\address[citcea]{Centre d'Innovació Tecnològica en Convertidors Estàtics i Accionaments (CITCEA-UPC), Universitat Politècnica de Catalunya \textperiodcentered\ BarcelonaTech (UPC), Barcelona, Spain}

\address[bnn]{Barcelona Neural Networking Center (BNN), Universitat Politècnica de Catalunya \textperiodcentered\ BarcelonaTech (UPC), Barcelona, Spain}

\tnotetext[1]{\textit{Code and data availability}: The code and experimental datasets will be made publicly available upon formal publication of the paper.}

\begin{abstract}
Fast and accurate models for the~\gls{acpf} problem are becoming increasingly important as modern energy grids require more frequent status analyses.~\glspl{gnn} have emerged as a natural choice due to their ability to directly exploit the topology of electrical grids. However, despite the growing number of proposed architectures, the influence of their architectural design choices on predictive performance remains poorly understood, as existing proposals often rely on different graph representations and evaluation settings.

This paper presents a unified benchmark of representative~\gls{gnn} architectures for~\gls{acpf} prediction under a common experimental framework. All models are evaluated using identical data preprocessing, training procedures, and comparable parameter budgets, isolating the impact of different message-passing mechanisms and inductive biases. The benchmark is complemented by a diagnostic analysis of the best-performing architectures and a systematic ablation study to quantify the contribution of their principal design components.

Experimental results on the IEEE 14-bus and IEEE 118-bus datasets show that anisotropic message-passing architectures provide the best trade-off between predictive accuracy and physical consistency under the parameter and topology constraints studied. Furthermore, the ablation study reveals that edge features and global context aggregation are key to accurately capturing the physical behavior of the electrical grids, whereas model depth yields an optimal capacity allocation under this budget at 2–3 layers, beyond which additional depth offers no performance gain under parameter constraints. These findings demonstrate that appropriate architectural inductive biases are more influential than model capacity for topologically bottlenecked grids such as IEEE-118, providing practical guidelines for designing graph-based models for power systems applications.
\end{abstract}

\glsresetall 


\maketitle

\section{Introduction}\label{sec:introduction}

The transformation of modern energy distribution networks is increasing the volume and frequency of the computations required for their operation. The growing penetration of renewable generation, the electrification of transportation and industry, and the expansion of distributed energy resources require increasingly frequent analyses of the electrical network to guarantee an efficient and secure operation. Among these analyses, the~\gls{acpf} represents one of the fundamental blocks, as it determines the steady-state electrical variables from a specific operating condition and serves as the base for higher-level applications such as~\gls{opf}. While \gls{dc} approximations of this problem rely on straightforward linear systems, the full \gls{acpf} is highly non-linear and requires iterative numerical solvers.

Although numerical techniques can solve the~\gls{acpf} with high accuracy, their iterative process arises as a limitation in applications that require a high number of evaluations, as their solving time depends on the size of the grid, posing a computational bottleneck. This has motivated the development of data-driven models capable of approximating solutions at significantly lower cost. In recent years,~\glspl{gnn} have gained traction due to their design, which can naturally operate on graph-structured data, allowing the topology of the electrical grid to be directly incorporated into the learning process.

Although~\glspl{gnn} have been widely adopted for this task~\cite{donon2020neural, lin2024powerflownet, piloto2024canos, abelezele2025empirical}, it remains unclear \textit{which} architectural choices are most effective and \textit{why}. Existing studies primarily focus on proposing new architectures or improving the predictive accuracy on specific datasets. However, comparisons are often performed under different graph representations, training settings, parameter budgets, and evaluation settings, making it difficult to isolate the contribution of individual architecture design choices. As a result, there is still limited understanding of how different message-passing mechanisms and inductive biases influence predictive accuracy, physical consistency, and how these models interact with power grids.

The~\gls{acpf} calculates the operating point of the network, which represents a direct translation of Ohm's and Kirchhoff's Laws over the grid topology. Since it is the foundation upon which~\gls{opf} and~\gls{se} are built, it provides a controlled environment for evaluating graph learning architectures.

Building on this premise, this paper presents a unified benchmark of representative message-passing~\glspl{gnn} for~\gls{acpf} prediction. Rather than proposing yet another architecture, our objective is to systematically evaluate how different message-passing mechanisms and architectural inductive biases affect model performance under a common experimental framework, providing a tool and analysis that help practitioners to better understand the application of \glspl{gnn} to modern power grid challenges.

To systematically investigate the role of architectural design choices, the proposed benchmark is structured into three complementary phases:

\begin{itemize}
    \item \textbf{Phase 1 -- Broad architecture benchmark.} A representative set of message-passing~\glspl{gnn} are evaluated under a unified experimental framework using identical topology representations, training protocols, and comparable parameter budgets to ensure a fair comparison across different architectural paradigms.
    \item \textbf{Phase 2 -- Architectural diagnostics.} The best-performing architectures are further analyzed to investigate how their structural properties relate to predictive accuracy and physical consistency. This phase complements the quantitative evaluation performed in Phase 1 with analyses that help explain the observed performance.
    \item \textbf{Phase 3 -- Ablation study.} We systematically dissect the top-performing architecture by isolating and removing key structural components. This quantifies their individual contributions to both predictive accuracy and physical constraint adherence, revealing the most critical design principles.
\end{itemize}

The main contributions of this work are summarized as follows:

\begin{itemize}
\vspace{-0.5em}
    \item We present a controlled benchmark of representative \gls{gnn} message-passing mechanisms for \gls{acpf}, evaluating them within a common heterogeneous graph representation, aggregation operator, training protocol, and comparable parameter budget.

    \item We systematically analyze the architectural properties that most strongly influence prediction accuracy and physical consistency, combining structural error diagnostics and targeted ablation studies to identify the contributions of individual message-passing mechanisms and inductive biases.

    \item We derive practical design guidelines for graph-based models in power systems within in-distribution load variations on fixed topologies, identifying the inductive biases that are most effective for capturing the underlying electrical behavior.
\end{itemize}

\paragraph{Scope and Boundary Conditions}
To maintain a sharp focus on model expressivity and physical fidelity, this work operates under two explicit boundaries.
First, as computational speedups over classical solvers (e.g., Newton--Raphson) are already well-documented \cite{donon2020neural, lin2024powerflownet, piloto2024canos}, we prioritize diagnosing architectural failure modes, and physical law violations over raw execution speed.
Second, we evaluate models on static grid topologies with varying power injections; this controlled baseline isolates architectural inductive biases from the confounding effects of dynamic topological shifts (such as $N-1$ line contingencies).

The remainder of this paper is organized as follows. Section~\ref{sec:related_work} reviews the related literature. Section~\ref{sec:methodology} presents the proposed methodology. Section~\ref{sec:experimental_setup} describes the datasets and experimental setup. Section~\ref{sec:results} presents the experimental results, which are discussed in Section~\ref{sec:discussion}. Finally, Section~\ref{sec:conclusion} concludes the paper and outlines future research directions.

\section{Related Work}\label{sec:related_work}

The application of~\gls{ml} to power system analysis has evolved from conventional feed-forward neural networks to graph-based models that explicitly exploit the topology of the electrical grid. The design of~\glspl{gnn}, which allows them to naturally exploit the topological information governing electrical networks~\cite{liao2021review,ringsquandl2021power}, has made them one of the dominant paradigms for learning-based power system analysis. While this evolution has led to substantial improvements in predictive accuracy and computational efficiency, most authors focus on proposing new architectures rather than systematically analyzing the architectural design choices that motivate these new solutions. This section reviews the evolution of learning-based solutions for power system analysis before discussing existing benchmarking efforts and the research gap addressed in this work.

\subsection{From Dense Networks to Graph-Based Power Flow}

The growing scale and complexity of power systems demand increasingly fast solutions for~\gls{acpf},~\gls{opf}, and~\gls{se}. While conventional iterative solvers provide accurate baselines, their computational burden in high-volume applications has motivated the use of~\gls{ml} to approximate physical states with significantly lower inference times~\cite{paucar2002artificial,mosbah2015multilayer}.

Early data-driven approaches relied on conventional feed-forward~\glspl{nn}. However, because dense~\glspl{nn} treat electrical grids as fixed-size vectors, they ignore the underlying network topology and local physical interactions, fundamentally limiting their ability to generalize across varying grid configurations~\cite{ringsquandl2021power}. To overcome these limitations,~\glspl{gnn} have emerged as the dominant learning paradigm~\cite{liao2021review}. By representing buses and transmission lines as nodes and edges, these models explicitly incorporate relational inductive biases, restricting information exchange to electrically connected components.

Early frameworks like GraphNeuralSolver~\cite{donon2020neural} demonstrated that message-passing architectures could accurately approximate~\gls{acpf} solutions with significantly fewer parameters and faster inference times than conventional~\glspl{nn} \cite{piloto2024canos, ghamizi2024opf}. Subsequent research has rapidly expanded this paradigm by incorporating heterogeneous graph representations, edge-aware message passing (e.g., using line impedances), and attention mechanisms~\cite{lin2024powerflownet,piloto2024canos,arowolo2025towards}. Beyond standard~\gls{acpf}, these graph-based models have also been successfully deployed for~\gls{opf}, acting both as surrogate models and as accelerators for numerical optimization~\cite{ghamizi2024opf,diehl2019warm}.

\subsection{Existing GNN Benchmarks and the Case for Architectural Analysis}

Recently, several works have systematically explored the use of \glspl{gnn} in power systems, focusing on empirical performance and the identification of effective configurations. However, these studies provide limited insight into how architectural choices interact with the underlying physical structure of power networks. In particular, it remains unclear how key design decisions---such as spatial aggregation operators, message-passing expressiveness, and the integration of edge-level electrical parameters---impact the ability of these models to capture the continuous, physics-driven nature of the \gls{acpf} problem.

Recent benchmarking efforts have highlighted this gap but largely address it from a data-centric perspective (Table~\ref{tab:benchmark_comparison}). For instance, PowerGraph~\cite{varbella2024powergraph} and PF$\Delta$~\cite{rivera2025pf} introduce extensive, multi-task datasets supporting contingency analysis and diverse operating conditions. While these works successfully standardize evaluation scenarios and highlight the superiority of attention-based models over isotropic convolutions, they evaluate a limited set of architectures. Crucially, they do not perform targeted ablations or link performance differences to physical properties like \gls{kcl} satisfaction and line impedance sensitivity.

Other studies are mainly focused on architectural novelties or scaling behaviors rather than controlled comparisons. Arowolo and Cremer~\cite{arowolo2025towards} demonstrate that hybrid heterogeneous architectures and multi-scale attention improve \gls{acopf} generalization, but the lack of ablation leaves the isolated contribution of each component unclear. Similarly, while Talebi and Zhou~\cite{talebi2025graph} confirm the benefits of attention-based models on varying loads, their comparison relies on homogeneous graphs without edge conditioning and does not control for parameter budgets, conflating architectural quality with raw model capacity~\cite{dwivedi2023benchmarking}. Finally, the most recent and complete work comes from Abelezele et al. \cite{abelezele2025empirical}, which compares 6 homogeneous \glspl{mpnn} for the \gls{opf} problem, partially aligning with this work. However, the results are mainly reported as statistical and physical metrics, together with training time. No controlled parameter setting, analysis of the nature of the errors, or ablation tests were provided. From a data perspective, Liu et al.~\cite{liu2026scaling} establish predictable scaling laws for \gls{ml} in \gls{opf}, but they focus on how performance evolves for a fixed architecture class rather than investigating which design choices are optimal, without proposing it as a benchmarking paper.

Collectively, these works confirm the computational viability of \glspl{gnn}, yet they reveal a consistent analytical gap. Historically driven by electrical engineering objectives, existing literature treats neural architectures largely as black boxes---prioritizing execution speedups and minimal prediction errors over mechanistic, computer science explanations for \textit{why} specific graph-theoretic mechanisms succeed or fail.

This work bridges these disciplines by comprehensively exploring \gls{gnn} behavior on the \gls{acpf} problem from an architectural standpoint. To this end, our experimental design moves beyond standard accuracy metrics: we introduce a controlled broad benchmark across the full message-passing taxonomy, conduct an in-depth analysis of error nature---specifically evaluating error propagation as a function of topological distance from the slack bus---and execute targeted ablation studies that isolate the contribution of individual inductive biases to physical constraint satisfaction.

\begin{table}[H]
\centering
\caption{Comparison of existing \gls{gnn} benchmarks for \gls{acpf}. \checkmark = explicitly addressed; $\circ$ = partially addressed; $\times$ = not addressed.}
\label{tab:benchmark_comparison}
\small
\begin{adjustbox}{width=\textwidth}
\begin{tabular}{lccccccccc}
\toprule
\textbf{Work} & \textbf{Tasks} & \textbf{Hetero.} & \textbf{Edge} & \textbf{Param.} & \textbf{Arch.} & \textbf{Targeted} & \textbf{Physics} & \textbf{Error} & \textbf{Data} \\
 & & \textbf{Graph} & \textbf{Cond.} & \textbf{Control} & \textbf{Taxonomy} & \textbf{Ablation} & \textbf{Metrics} & \textbf{Diagnostics} & \textbf{Variation} \\
\midrule
PowerGraph \cite{varbella2024powergraph} & \gls{acpf}, \gls{opf} & $\times$ & $\circ$ & $\times$ & $\circ$ & $\times$ & $\times$ & $\times$ & $?$ \\
PF$\Delta$ \cite{rivera2025pf} & \gls{acpf} & \checkmark & $\circ$ & $\times$ & $\times$ & $\times$ & \checkmark & $\times$ & load + topo + gen \\
Arowolo et al. \cite{arowolo2025towards} & \gls{opf} & \checkmark & \checkmark & $\times$ & $\times$ & $\times$ & $\circ$ & $\times$ & $\pm$10\% load \\
Talebi et al. \cite{talebi2025graph} & \gls{acpf} & $\times$ & $\times$ & $\times$ & $\circ$ & $\times$ & $\times$ & $\times$ & $\pm$40\% load \\
Abelezele et al. \cite{abelezele2025empirical} & \gls{opf} & $\times$ & $\circ$ & $\times$ & \checkmark & $\times$ & \checkmark & $\times$ & $?$ \\
\midrule
\textbf{This work} & \textbf{\gls{acpf}} & \checkmark & \checkmark & \checkmark & \checkmark & \checkmark & \checkmark & \checkmark & $\pm$80\% load \\
\bottomrule
\end{tabular}
\end{adjustbox}
\end{table}

\section{Methodology}\label{sec:methodology}

The objective of this work is to systematically compare representative \glspl{gnn} under a unified framework. Although these architectures differ considerably in their computational mechanisms, many of them can be integrated within the \gls{mpnn} \cite{gilmer2017neural} framework or closely related message-passing formulations. This abstraction provides a common terminology to describe how information is propagated among neighboring nodes and for identifying the main architectural choices.

\subsection{The Message-Passing Framework}
\label{sec:mpnn_framework}
While \glspl{gnn} broadly refer to \glspl{nn} operating over graph-structured data, \glspl{mpnn} represent a standardized mathematical blueprint for architectures that iteratively exchange information between neighboring nodes. At each layer $\ell$, the update is decomposed into message generation and aggregation, followed by a node update.

\begin{enumerate}
    \item \textbf{Message Generation and Aggregation:} For a target node $i$ (e.g., a specific bus), a message function $\phi^{(\ell)}$ computes interactions with each physical neighbor $j \in \mathcal{N}(i)$, optionally incorporating continuous edge features $\mathbf{e}_{ij}$ (e.g., line impedances). These messages are then pooled using a permutation-invariant aggregation operator (e.g., \texttt{sum}, \texttt{mean}, or \texttt{max}) to form a single aggregated neighborhood message $\mathbf{m}_i^{(\ell)}$:
    \begin{equation}
        \mathbf{m}_i^{(\ell)} = \text{AGGREGATE}\left( \left\{ \phi^{(\ell)}\!\left(\mathbf{h}_i^{(\ell-1)}, \mathbf{h}_j^{(\ell-1)}, \mathbf{e}_{ij}\right) \mid j \in \mathcal{N}(i) \right\} \right)
    \end{equation}

    \item \textbf{Node Update:} The aggregated message is passed to an update function $\psi^{(\ell)}$ (such as a neural network or a gated recurrent unit), which combines it with the node's previous state $\mathbf{h}_i^{(\ell-1)}$ to produce the new embedding $\mathbf{h}_i^{(\ell)}$:
    \begin{equation}
        \mathbf{h}_i^{(\ell)} = \psi^{(\ell)}\!\left(\mathbf{h}_i^{(\ell-1)}, \mathbf{m}_i^{(\ell)}\right)
    \end{equation}
\end{enumerate}

By casting diverse architectures (e.g., \gls{gcn}) as specific instantiations of $\phi$, $\psi$, and the aggregator, we can pinpoint exactly where their computational mechanics differ. This common terminology ensures that our benchmark isolates and evaluates their distinct relational inductive biases in a way that is highly interpretable for physical network analysis.

\subsection{Message Passing Taxonomy}
\label{sec:mp_tax}
Following and extending the taxonomy proposed in \cite{abelezele2025empirical}, we evaluate the architectures along three main axes: the domain of the convolution operation, the treatment of neighbor importance, and the integration of edge attributes \cite{cao2020comprehensive}.

\subsubsection*{Domain of Operation}
\begin{itemize}
    \item \textbf{Spatial models:} Spatial \glspl{gnn} define graph convolutions directly on the original graph structure by iteratively exchanging messages along existing edges \cite{hamilton2017inductive}. Each node updates its embedding by aggregating information from its one-hop neighborhood through operations such as summation, averaging, or attention-weighted pooling. This localized approach natively supports inductive learning, allowing the model to generalize to entirely unseen graphs without relying on global structural assumptions.

    \item \textbf{Spectral models:} Spectral \glspl{gnn} define convolutions through the eigendecomposition of the graph Laplacian \cite{bruna2013spectral, defferrard2016convolutional}. Although this provides a rigorous connection with graph signal processing, the learned filters depend on the Laplacian eigenstructure of the graph, limiting the generalization across topologies. Since this favors learning topology-specific spectral patterns rather than local information-flow rules \cite{liao2021review}, spectral models are excluded from the benchmark.
\end{itemize}

\subsubsection*{Treatment of Neighbor Importance}
\begin{itemize}
    \item \textbf{Isotropic models:} These models apply the same transformation across neighboring nodes, with message weights determined independently of the node features. Examples include the degree-normalized weighting of \gls{gcn} \cite{kipf2016semi}, or the uniform sum aggregation of \gls{gin} \cite{xu2018powerful}.

    \item \textbf{Anisotropic models:} These architectures use feature-dependent filters to assign different importance to individual edges. Attention-based \gls{gat} \cite{velivckovic2017graph} and gated \gls{resgated} \cite{bresson2017residual} are representative examples, enabling directional and adaptive information flow.
\end{itemize}

\subsubsection*{Integration of Edge Attributes}

\begin{itemize}
    \item \textbf{Topology-only models:} These architectures utilize edges as a structural routing mechanism to define the local neighborhood, passing messages based on the binary presence of a connection in the adjacency matrix. While computationally efficient, these models (such as the standard \gls{gcn} or GraphSAGE) are inherently limited in physical domains because they ignore the multi-dimensional properties of the connections.
    \item \textbf{Edge-conditioned models:} These architectures explicitly incorporate continuous edge features into the message-passing function \cite{battaglia2018relational, gilmer2017neural}. Instead of treating edges only as pathways, the message sent from node $j$ to node $i$ is actively transformed by the feature vector of the edge connecting them (e.g., using a neural network conditioned on the edge attributes). This capability is particularly well-suited for domains where interactions are governed by connection parameters.
\end{itemize}

\begin{table}[htpb]
    \centering
    \renewcommand{\arraystretch}{1.2}
    \caption{Taxonomy of Graph Neural Network Architectures Across Design Axes}
    \label{tab:gnn_classification}
    \begin{adjustbox}{width=\textwidth}
    \begin{tabular}{@{}lllcl@{}}
        \toprule
        \textbf{Architecture} & \textbf{Domain} & \textbf{Treatment} & \textbf{Native Edge Support} & \textbf{Core Aggregation Mechanism} \\
        \midrule
        \textbf{Spectral CNN} \cite{bruna2013spectral} & Spectral & Isotropic & No & Eigendecomposition of the Graph Laplacian \\
        \textbf{ChebNet} \cite{defferrard2016convolutional}      & Spectral & Isotropic & No & Chebyshev polynomial filter approximation \\
        \midrule
        \textbf{\gls{gcn}} \cite{kipf2016semi}      & Spatial\textsuperscript{1} & Isotropic   & No                      & Degree-normalized symmetric sum \\
        \textbf{GraphSAGE} \cite{hamilton2017inductive}      & Spatial                    & Isotropic   & No                      & Uniform sampling with mean/max pooling \\
        \textbf{\gls{gin}} \cite{xu2018powerful}      & Spatial                    & Isotropic   & No                   & Sum aggregation with multi-layer perceptrons \\
        \textbf{GINE} \cite{hu2019strategies}     & Spatial                    & Isotropic   & Yes                        & Sum aggregation with multi-layer perceptrons \\
        \midrule
        \textbf{\gls{gat}} \cite{velivckovic2017graph}     & Spatial & Anisotropic & No\textsuperscript{2} & Single or multi-head attention scores \\
        \textbf{\gls{resgated}} \cite{bresson2017residual} & Spatial & Anisotropic & No                      & Residual connections with edge gating \\
        \textbf{Graph Transformer} \cite{dwivedi2023benchmarking}   & Spatial & Anisotropic & Yes                     & Multi-head attention with positional/edge embeddings \\
        \bottomrule
    \end{tabular}
    \end{adjustbox}
    \smallskip

    \noindent\footnotesize
    \textsuperscript{1}\gls{gcn} originates from spectral graph theory but functions spatially in its localized first-order approximation.\\
    \textsuperscript{2}The original \gls{gat} model utilizes edges strictly for topological masking to define neighborhoods, though extended implementations integrate continuous edge attributes into the attention coefficients.
\end{table}

\subsection{Detailed Architecture Specifications}
\subsubsection*{Shared vs. Independent Weights}
\label{sec:inductive_bias}
Classical~\gls{acpf} solvers such as the~\gls{nr} method iteratively refine the system state until convergence. Motivated by this iterative structure, we compare two architectural strategies: stacking layers with independent parameters and repeatedly applying the same learnable transformation with shared weights~\cite{lin2024powerflownet}. The latter can be interpreted as a recurrent inductive bias, where each iteration performs another refinement step using the same set of parameters.

\subsubsection*{Modeling Self-Information in Heterogeneous Power Grids}
\label{sec:self_loops}
In graph-based learning, each node should preserve part of its own representation while aggregating information from its neighbors. In homogeneous graphs, this is commonly achieved through self-loops. However, directly introducing self-loops is not straightforward in heterogeneous power-grid representations, particularly for relations connecting different node types \cite{bose2026learning}.
To avoid modifying the graph topology, architectures requiring an explicit self-contribution, such as~\gls{gcn},~\gls{gat}, and the Graph Transformer, incorporate it through a dedicated root-weight transformation~\cite{hamilton2017inductive}. This preserves the node's own features while keeping the original network structure unchanged. The remaining evaluated architectures already include self-information through their native message-passing formulations. This choice allows the comparison to focus on differences between message-passing architectures rather than on changes to the underlying graph representation.

\subsection{Heterogeneous Message Passing Pipeline}
\label{sec:shared_template}
To rigorously evaluate different relational inductive biases, all \gls{gnn} architectures are embedded within a standardized three-stage heterogeneous pipeline \cite{bohigas2026unified}: type-specific input encoding, a message-passing backbone, and node-specific output decoding. This controlled design isolates performance variations strictly to the evaluated message-passing formulations.

\begin{figure}[H]
\centering
\resizebox{0.65\linewidth}{!}{
\begin{tikzpicture}[
    node distance=1.6cm and 2.2cm,
    every node/.style={font=\small},
    block/.style={draw, rounded corners, minimum width=2.8cm, minimum height=1cm, align=center},
    smallblock/.style={draw, rounded corners, minimum width=2.5cm, minimum height=0.8cm, align=center},
    arrow/.style={->, thick}
]

\node[smallblock, fill=orange!10] (edge_enc) {\textbf{Edge Encoder}};
\node[block, fill=teal!10, below=1cm of edge_enc] (edge_norm) {\textbf{Layer Norm (E)}};
\node[block, fill=gray!10, below=1cm of edge_norm] (edge_merge) {\textbf{Edge Embeddings}};

\node[smallblock, fill=blue!10, right=1.5cm of edge_enc] (pq) {\textbf{PQ Encoder}};
\node[smallblock, fill=green!10, right=0.6cm of pq] (pv) {\textbf{PV Encoder}};
\node[smallblock, fill=purple!10, right=0.6cm of pv] (slack) {\textbf{Slack Encoder}};

\node[block, fill=gray!5, above=1.5cm of pv, inner sep=0.3cm] (input_graph) {
    \textbf{Input Graph} \\[2ex] 
    \begin{tikzpicture}[scale=0.6, every node/.style={circle, draw, minimum size=0.4cm, inner sep=0pt}]
        \node[fill=purple!20] (s)  at (0, 1.2) {};
        \node[fill=green!20]  (v)  at (1.5, 1) {};
        \node[fill=blue!20]   (q1) at (0.5, 0) {};
        \node[fill=blue!20]   (q2) at (2, 0.2) {};
        
        \draw[thick, orange!80!black] (s) -- (v);
        \draw[thick, orange!80!black] (s) -- (q1);
        \draw[thick, orange!80!black] (v) -- (q1);
        \draw[thick, orange!80!black] (v) -- (q2);
        \draw[thick, orange!80!black] (q1) -- (q2);
    \end{tikzpicture}
};

\draw[arrow] (input_graph.south) -- (edge_enc.north);
\draw[arrow] (input_graph.south) -- (pq.north);
\draw[arrow] (input_graph.south) -- (pv.north);
\draw[arrow] (input_graph.south) -- (slack.north);

\node[block, fill=teal!10, below=1cm of pv] (norm) {\textbf{Layer Norm (N)}};
\node[block, fill=gray!10, below=1cm of norm] (merge) {\textbf{Node Embeddings}};

\node[block, fill=orange!10, below=3.2cm of merge, xshift=-1.8cm] (indep) {\textbf{Independent} \\ \textbf{Layer}};
\node[block, fill=yellow!20, below=3.2cm of merge, xshift=1.8cm] (shared) {\textbf{Shared Layer}};
\node[font=\normalsize\bfseries, at={($(indep)!0.5!(shared)$)}] (or) {\textit{or}};

\node[block, fill=purple!10, below=1.2cm of or] (update) {\textbf{Update Step}};

\node[block, fill=gray!20, below=1.8cm of merge, xshift=4.4cm] (pooling) {\textbf{Global Pooling} \\ Pool + MLP};

\coordinate (bb_top) at ($(merge.south) + (0, -0.4)$);
\coordinate (bb_bottom) at ($(update.south) + (0, -0.9)$);
\coordinate (bb_left) at ($(or) + (-5.6, 0)$);
\coordinate (bb_right) at ($(or) + (7.4, 0)$);

\begin{scope}[on background layer]
  \node[draw=gray, dashed, rounded corners, inner sep=0pt, fill=yellow!10,
        fit=(bb_top)(bb_bottom)(bb_left)(bb_right)] (backbone) {};
  
  \node[draw=black, dotted, thick, rounded corners, inner sep=0.3cm, fill=green!15, 
        fit=(indep)(shared)(or)] (layer_box) {};
\end{scope}

\node[below right=0.15cm and 0.2cm of backbone.north west] (backbone_title) {\small\textit{\textbf{Message-Passing Backbone}}};

\node[block, fill=red!10, below=2.1cm of update, xshift=-1.5cm] (theta) {\textbf{$V_a$ Head}};
\node[block, fill=red!10, below=2.1cm of update, xshift=1.5cm] (vhead) {\textbf{$V_m$ Head}};

\node[below=0.6cm of theta, font=\large] (theta_out) {$\hat{V_a}$};
\node[below=0.6cm of vhead, font=\large] (v_out) {$\hat{V_m}$};

\coordinate (outcenter) at ($(theta_out)!0.5!(v_out)$);
\node[below=0.4cm of outcenter, align=center] (bustext) {For each bus \\ type PV/PQ};

\draw[arrow] (edge_enc.south) -- (edge_norm.north);
\draw[arrow] (edge_norm.south) -- (edge_merge.north);

\draw[arrow] (pq.south) -- (norm.north west);
\draw[arrow] (pv.south) -- (norm.north);
\draw[arrow] (slack.south) -- (norm.north east);
\draw[arrow] (norm.south) -- (merge.north);

\coordinate (split_point) at ($(merge.south) + (0, -1.2)$);

\draw[thick] (merge.south) -- (split_point);
\draw[arrow] (split_point) -- (layer_box.north);

\draw[thick] (edge_merge.south) -- (edge_merge.south |- bb_top);
\draw[arrow] (edge_merge.south |- bb_top) -- (edge_merge.south |- layer_box.west) -- (layer_box.west);

\coordinate (resid_x) at ($(or) + (-4.8, 0)$);
\draw[arrow] (split_point) -| (resid_x) |- node[pos=0.25, left, xshift=-0.1cm, align=right] {\textbf{Residual} \\ \textbf{connection}} (update.west);

\draw[arrow] (split_point) -| (pooling.north);

\coordinate (root_x) at ($(or) + (6.4, 0)$);

\draw[thick] (merge.east) -- (merge.east -| root_x);
\draw[arrow] (merge.east -| root_x) |- node[pos=0.31, right, xshift=0.1cm, align=left] {\textbf{Root} \\ \textbf{connection}} ($(update.east) + (0, -0.2)$);

\draw[arrow] (pooling.south) |- ($(update.east) + (0, 0.2)$);

\coordinate (loop_start) at ($(backbone.east |- update.south) + (0, 0.4)$);
\coordinate (loop_end) at ($(backbone.east |- split_point) + (0, 0)$);

\draw[arrow, rounded corners=8pt] 
    (loop_start) -- ++(1.8, 0) coordinate (loop_right) 
    -- node[right, align=center] {\textbf{Layer} \\ \textbf{Loop}} (loop_right |- loop_end) 
    -- (loop_end);

\coordinate (exit_point) at ($(backbone.south) + (0,0)$);
\draw[thick] (update.south) -- (update.south |- exit_point);
\coordinate (split) at ($(update.south |- exit_point) + (0,-0.6)$);
\draw[thick] (update.south |- exit_point) -- (split);
\draw[arrow] (split) -| (theta.north);
\draw[arrow] (split) -| (vhead.north);

\draw[arrow] (theta.south) -- (theta_out.north);
\draw[arrow] (vhead.south) -- (v_out.north);

\end{tikzpicture}}
\caption{Proposed heterogeneous \gls{gnn} architecture for \gls{acpf} prediction, highlighting the common components.}
\label{fig:arch_structure}
\end{figure}
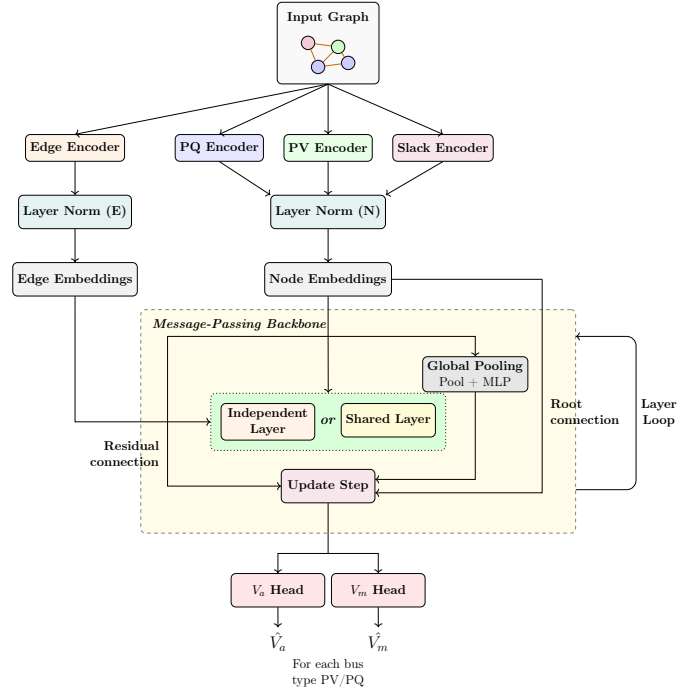

\subsubsection*{Input Encoding}
\label{sec:encoding}
Because the three bus types reside in different physical feature spaces, the encoding stage is type-specific. For each node type \(\tau \in \{\text{PQ}, \text{PV}, \text{slack}\}\), a dedicated linear projection maps the raw input features to a shared hidden dimension \(d_h\), followed by Layer Normalization to stabilize the initial embedding scale. Edge features are encoded analogously: for each directed edge type, a linear layer projects the raw impedance vector (resistance and reactance) to \(\mathbb{R}^{d_h}\), followed by a \gls{silu} activation. These encoded edge representations are computed once and reused across all message-passing layers.

\subsection{Message-Passing Backbone}
\label{sec:backbone}
The backbone performs \(L\) rounds of message-passing over the heterogeneous graph. We evaluate two weight-sharing paradigms:
\begin{itemize}
    \item \textbf{Independent weights (Deep \gls{gnn}):} Each layer \(\ell\) learns distinct parameters.
    \item \textbf{Shared weights (Recurrent \gls{gnn}):} A single layer is applied \(L\) times. A per-type \gls{gru} cell acts as a gating mechanism to modulate the update, effectively unrolling the iterative physical \gls{acpf} solver.
\end{itemize}

\subsubsection*{Residual, Root, and Global Connections}
To mitigate oversmoothing, a widely known problem in \gls{mpnn} deployment where the node representation becomes indistinguishable, the node update at each layer combines the transformed incoming message, a residual connection to the previous layer, and a root connection to the initial encoding \cite{scholkemper2025residual}. Additionally, an optional global context vector—computed via mean-pooling across all node types—can be broadcast to every node to provide a soft global signal \cite{liu2022graph}.

\subsubsection*{Output Decoding}
\label{sec:decoding}
After \(L\) steps, the final node embeddings are passed through type-specific regression heads. For each prediction target (voltage magnitude \(V_m\) and angle \(V_a\)), a separate two-layer \gls{mlp} with \gls{silu} activation produces a scalar output. Independent heads are utilized to accommodate the distinct physical scales and distributional asymmetries of \(V_m\) and \(V_a\). The slack bus requires no decoder, as its states are fixed operator setpoints.

\subsubsection*{Architectural Harmonization}
\label{sec:harmonization}
To isolate the effect of each specific message-passing inductive bias, we apply three uniform modifications across all evaluated \gls{gnn} architectures, ensuring a strictly controlled benchmark:

\begin{itemize}
    \item \textbf{Edge Feature Ingestion:} To prevent performance discrepancies arising simply from missing physical data, all models are harmonized to ingest continuous edge features (line impedances). Architectures that do not natively support edge features are augmented with an additive edge term in their message function. For attention-based models (\gls{gat} and Graph Transformer), edge features are directly incorporated into the attention score computation.
    \item \textbf{Aggregation Function:} We fix the aggregation function to \texttt{sum} pooling across all models as a controlled factor for Phase 1 fair comparison. As established by \cite{xu2018powerful}, sum pooling is the most expressive standard aggregator, ensuring performance differences stem solely from the message generation scheme. Phase 3 explicitly tests whether that theoretical property translates into the best empirical configuration for the winning architecture.
    \item \textbf{Self-Loop Encoding:} As detailed in Section~\ref{sec:self_loops}, models requiring explicit self-loops (\gls{gcn}, \gls{gat}, Graph Transformer) employ a dedicated, topology-neutral root-weight transformation. Architectures that naturally preserve self-representations via their specific message-passing formulations remain unmodified.
\end{itemize}

\section{Experimental Setup}\label{sec:experimental_setup}

The experimental setup is designed to provide a controlled comparison of the evaluated architectures under consistent data, training, and evaluation conditions. To this end, we define the benchmark grid topologies, dataset generation procedure, graph representation, training protocol, and evaluation criteria used throughout the study.

\subsection{Benchmark Grid Topology}
\label{sec:grid_topology}

The IEEE-14 and IEEE-118 systems differ not only in size but also in topological properties relevant to message passing. As summarized in Table~\ref{tab:grid_topology}, both grids have a similar mean degree, but IEEE-118 has a substantially larger diameter and average shortest-path length, together with an algebraic connectivity ($\lambda_1$) almost 20 times smaller. These properties indicate a stronger bottleneck for long-range information propagation, including the slack-bus reference~\cite{giraldo2023trade}, motivating their treatment as distinct topological regimes in the subsequent analysis.

\begin{table}[ht]
\centering
\caption{Topological properties of the IEEE benchmark grids used in this work.}
\label{tab:grid_topology}
\begin{tabular}{lrrrrrrr}
\toprule
\textbf{Grid}
  & $N$
  & $E$
  & $\bar{d}$
  & PV:PQ
  & Diameter
  & ASPL
  & $\lambda_1$ \\
\midrule
IEEE-14  & 14  & 20  & 2.86 & 0.44 & 5  & 2.37 & 0.182 \\
IEEE-118 & 118 & 179 & 3.03 & 0.83 & 14 & 6.31 & 0.009 \\
\bottomrule
\end{tabular}

\smallskip
\footnotesize
\noindent
$N$: total buses; $E$: undirected transmission lines;
$\bar{d}$: mean node degree ($2E/N$);
PV:PQ: ratio of PV to PQ buses;
ASPL: average shortest path length;
$\lambda_1$: algebraic connectivity (smallest nonzero eigenvalue of the admittance-weighted normalized Laplacian, computed once from the fixed grid topology and line impedances).
\end{table}

\subsection{Dataset Generation}
\label{sec:data_gen}

For each IEEE topology, we generate 2,000 feasible grid states using \texttt{pandapower}~\cite{thurner2018pandapower}. Active and reactive power demands are uniformly perturbed by $\pm 80\%$ around their nominal values, while transmission-line impedances are varied by $\pm5\%$. The load range is chosen to cover a broad set of operating conditions~\cite{donon2020neural, rivera2025pf, talebi2025graph} and is wider than those adopted in several previous studies~\cite{ piloto2024canos, arowolo2025towards,owerko2018predicting, dogoulis2025kclnet}. Non-convergent samples are discarded.

\subsubsection*{Graph Representation}
Each grid state is modeled as an undirected heterogeneous graph. Nodes are categorized into \texttt{PV}, \texttt{PQ}, and \texttt{slack} buses to capture their distinct physical roles. Table \ref{tab:bus_types} summarizes the bus types and their characteristics.

\begin{itemize}
    \item \textbf{Node Features ($H$):} The input feature space consists of the known boundary conditions for the \gls{acpf} problem. These are type-specific: \texttt{PQ} nodes receive active and reactive demand, while \texttt{PV} nodes receive active generation and voltage setpoints. Slack buses act as reference buses, with their active and reactive power outputs adjusting to satisfy the network power balance.
    \item \textbf{Edge Features ($E$):} Edges represent transmission lines and are modeled bidirectionally to facilitate symmetric message passing. Each edge carries continuous physical attributes, specifically the resistance ($R$) and reactance ($X$).
    \item \textbf{Target Variables ($Y$):} The ground-truth targets generated by the numerical solver are limited entirely to the unknown voltage components. Specifically, we only compute the voltage magnitudes ($V_m$) at PQ buses and the voltage angles ($V_a$) at both PQ and PV buses. Although other variables remain unknown prior to solving (see Table~\ref{tab:bus_types}), determining $V_m$ and $V_a$ is sufficient to fully define the network state.
\end{itemize}

    \begin{table}[H]
    \centering
    \caption{Classification of buses in a power network according to their known and unknown variables.}
    \label{tab:bus_types}
    \begin{tabular}{lccc}
        \toprule
        \textbf{Bus Type} & \textbf{Known Variables} & \textbf{Unknown Variables} & \textbf{Role} \\
        \midrule
        PQ bus    & \(P\),\, \(Q\)           & \(V_m\),\, \(V_a\) & Load bus \\
        PV bus    & \(P\),\, \(V_m\)         & \(Q\),\, \(V_a\)   & Generator bus \\
        slack bus\footnotemark & \(V_a\),\, \(V_m\) & \(P\),\, \(Q\)       & Reference bus \\
        \bottomrule
    \end{tabular}
\end{table}
\footnotetext{For the slack bus, \(P\) and \(Q\) are calculated after solving the \gls{acpf}, whereas only \(V_m\) and \(V_a\) are specified a priori. Some formulations do not count \(P\) and \(Q\) as unknown state variables.}

The angles are reported in degrees. All electrical quantities are expressed in \gls{pu} on a 100 \gls{mva} base.

\subsection{Training and Evaluation Setup}
\label{sec:training and setup eval}

\subsubsection{Data Splitting and Standardization}

Samples are partitioned into training, validation, and test sets using a 75\%/10\%/15\% split, with four fixed seeds defining reproducible partitions across runs. All input features, edge attributes, and targets are standardized using Z-score normalization:
\begin{equation}
    \tilde{x} = \frac{x-\mu_x}{\sigma_x}
\end{equation}
where $\mu_x$ and $\sigma_x$ are computed exclusively from the training set and subsequently applied to the validation and test sets to prevent information leakage. Standardization is performed independently for each node type and feature.

\subsubsection{Training Configuration}

Models are trained using~\gls{mse}, with independent loss components for voltage magnitude $V_m$ and angle $V_a$. Because both targets are standardized independently, they contribute comparably to the optimization objective despite their different physical scales. Training uses Adam with a learning rate of $10^{-3}$, weight decay of $10^{-5}$, and batch size 128 for up to 1000 epochs. Early stopping is applied after 50 consecutive epochs without validation-loss improvement. Each configuration is trained with four different seeds, and results are reported as a mean and standard deviation across runs.

\subsection{Evaluation Metrics}
\label{sec:metrics}

We evaluate both statistical accuracy and physical feasibility. All metrics are computed after inverse-transforming predictions to physical units and are disaggregated by bus type where applicable to isolate bus-specific failure modes.

\subsubsection*{\gls{trmae}}
Statistical accuracy is measured using the~\gls{trmae}~\cite{piloto2024canos}, defined as:

\begin{equation}
    \text{\gls{trmae}} = \frac{1}{N} \sum_{i=1}^{N} \frac{|\hat{y}_i - y_i|}{\max(|y_i|,\, \tau)}
\end{equation}

where $N$ is the number of evaluated samples and $\tau=0.001$ prevents undefined or artificially inflated relative errors when $y_i$ approaches zero. This retains the interpretability of a relative percentage error while avoiding its instability near zero.~\gls{trmae} is reported separately for voltage magnitude and angle and by bus type.

\subsubsection*{\gls{kclmae}}
Since statistical accuracy does not guarantee satisfaction of fundamental electrical laws, we use~\gls{kclmae} as the decisive metric for model selection. Because real-world viability depends on physically feasible, \gls{kcl}-compliant solutions, this metric assesses whether the learned inductive biases respect physical constraints by measuring the complex power imbalance at each bus:

\begin{equation}
    \text{KCL-MAE} = \frac{1}{N_b} \sum_{k=1}^{N_b} \left| S_k^{\text{inj}} - \sum_{j \in \mathcal{N}(k)} S_{kj}^{\text{pred}} \right|
\end{equation}

where $N_b$ is the number of buses, $S_k^{\text{inj}}$ is the scheduled complex power injection at bus $k$, and $S_{kj}^{\text{pred}}$ is the complex power flow on line $(k,j)$ derived entirely from the predicted voltage magnitudes $V_m$ and angles $V_a$. A low~\gls{kclmae} therefore indicates greater alignment with Kirchhoff's conservation laws rather than only statistical fit.

To identify which power component drives the overall mismatch, we decompose the complex residual into its active and reactive components, $\Delta P$ and $\Delta Q$:

\begin{equation}
    \Delta P_k = P_k^{\text{inj}} - \sum_{j \in \mathcal{N}(k)} P_{kj}^{\text{pred}}, \qquad
    \Delta Q_k = Q_k^{\text{inj}} - \sum_{j \in \mathcal{N}(k)} Q_{kj}^{\text{pred}}
\end{equation}

where $P_k^{\text{inj}} = \Re\{S_k^{\text{inj}}\}$ and $Q_k^{\text{inj}} = \Im\{S_k^{\text{inj}}\}$ are the scheduled active and reactive injections, while $P_{kj}^{\text{pred}}$ and $Q_{kj}^{\text{pred}}$ are the corresponding real and imaginary components of $S_{kj}^{\text{pred}}$. Their mean absolute mismatches are:

\begin{equation}
    \Delta P = \frac{1}{N_b} \sum_{k=1}^{N_b} \left| \Delta P_k \right|, \qquad
    \Delta Q = \frac{1}{N_b} \sum_{k=1}^{N_b} \left| \Delta Q_k \right|
\end{equation}

While~\gls{kclmae} captures the overall complex power imbalance, $\Delta P$ and $\Delta Q$ isolate violations of active and reactive power balance, respectively. This distinction is particularly informative because both flows are governed by different, non-linearly coupled terms in the power-flow equations.

\subsection{Benchmark Structure}
\label{sec:benchmark_structure}

The benchmark comprises three phases: a broad architecture comparison, architectural diagnostics and a targeted ablation study. To ensure a controlled comparison, all configurations are constrained to approximately 100k trainable parameters by adjusting the hidden dimension within $[32, 256]$. This clearly helps differentiate architectural differences and model capacity, while keeping the experiments computationally feasible~\cite{dwivedi2023benchmarking}.

Each architecture is evaluated with $T \in \{1,2,3,4\}$ message-passing iterations under two propagation configurations. In the \textit{independent} configuration, each iteration has its own trainable parameters, whereas the \textit{shared recurrent} configuration reuses a single message-passing layer and combines its output with the previous node state through a \gls{gru}. Since the hidden dimension may decrease as $T$ increases to maintain the common parameter budget, the results characterize the trade-off between propagation depth and representation width under constrained capacity. Complete architecture specifications are reported in \ref{app:appendix_a}.

\subsubsection{Phase 1 -- Broad Architecture Benchmark}

Phase~1 compares seven representative architectures under the common experimental setting: \gls{gcn}, \gls{sage}, \gls{gine}, \gls{gat}, \gls{resgated}, Graph Transformer, and an \gls{mlp} baseline. Performance is evaluated using \gls{trmae} and \gls{kclmae}, so observed differences primarily reflect the message-passing mechanisms rather than differences in model capacity or optimization settings.

\subsubsection{Phase 2 -- Architectural Diagnostics}

Phase 2 analyzes prediction errors as a function of the shortest-path distance between each bus and the slack bus to assess whether model performance degrades as information must propagate over longer distances. This diagnostic characterizes how predictions are affected by the structural properties of the electrical grid and the locality of message passing. A complementary spectral error analysis is provided in Appendix~\ref{sec:spectral_analysis_appendix}.

\subsubsection{Phase 3 -- Ablation Study}
\label{sec:phase3}
To systematically identify the architectural components driving the best-performing model, we conduct a targeted ablation study. By removing or modifying one component at a time while keeping the base experimental protocol fixed, we isolate the causal impact of each design choice. Key evaluations are executed while maintaining the standard benchmark configuration introduced in Section~\ref{sec:benchmark_structure}. This enables a direct mechanistic interpretation of the model's behavior, closing the loop on our initial architectural hypotheses.

Rather than exhaustively disabling every hyperparameter, this phase targets five specific components chosen for their strong theoretical and physical grounding in power systems. The ablated components and their guiding hypotheses are detailed in Table~\ref{tab:ablation_targets}.

\begin{table}[H]
    \centering
    \renewcommand{\arraystretch}{1.3}
    \caption{Ablation Targets and their Physical or Theoretical Rationale}
    \label{tab:ablation_targets}
    \begin{adjustbox}{max width=\linewidth}
        \begin{tabular}{@{}lp{0.25\linewidth}p{0.55\linewidth}@{}}
            \toprule
            \textbf{Component} & \textbf{Ablated Configuration} & \textbf{Hypothesis \& Physical Rationale} \\
            \midrule
            \textbf{Residual Connections} & Disabled & Tests if skip connections are strictly necessary to prevent oversmoothing \cite{rusch2023survey}. \\
            \textbf{Edge Features} & Disabled (Topology-only) & Tests whether line impedances provide information beyond network topology. \\
            \textbf{Aggregation Function} & \texttt{mean} or \texttt{max} (vs. \texttt{sum}) & Tests whether \texttt{sum} provides a better inductive bias for physical energy conservation. \\
            \textbf{Global Pooling} & Disabled & Assesses the necessity of graph-level context injection for estimating global system states, particularly voltage angles \cite{arowolo2025towards}. \\
            \textbf{Root Connection} & Disabled & Evaluates whether anchoring representations to the original boundary conditions improves physical consistency. \\
            \bottomrule
        \end{tabular}
    \end{adjustbox}
\end{table}

\section{Results}\label{sec:results}

\subsection{Phase 1: Broad Architecture Benchmark}
\label{sec:phase1_results}

We focus the analysis on IEEE-118 because it is more challenging in absolute terms and its markedly lower algebraic connectivity (Table \ref{tab:grid_topology}) makes it the setting in which architectural inductive biases are most likely to be consequential. The \gls{mlp} baseline is omitted from the visualizations due to its larger errors, which make the differences among the graph-based methods difficult to observe. Complete numerical results for both grids are reported in \ref{app:appendix_b}.

\subsubsection*{Statistical Accuracy}
\begin{figure}[H]
    \centering
    \begin{subfigure}{0.45\linewidth}
        \includegraphics[width=\linewidth]{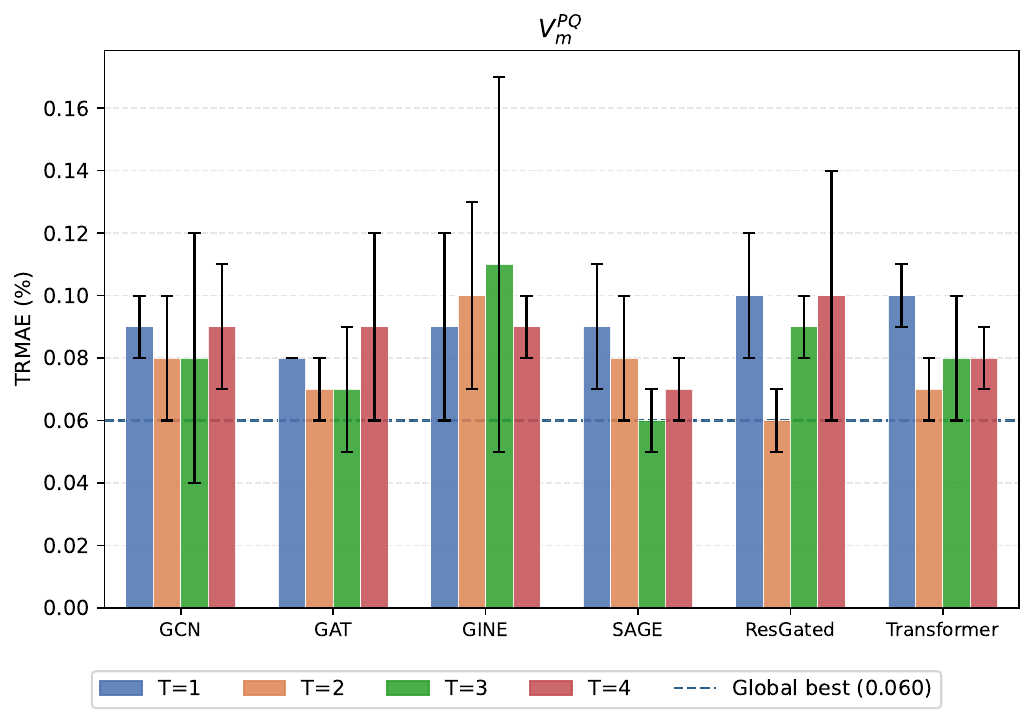}
        \caption{Independent weights}
        \label{fig:magnitude_independent}
    \end{subfigure}
    \hfill
    \begin{subfigure}{0.45\linewidth}
        \includegraphics[width=\linewidth]{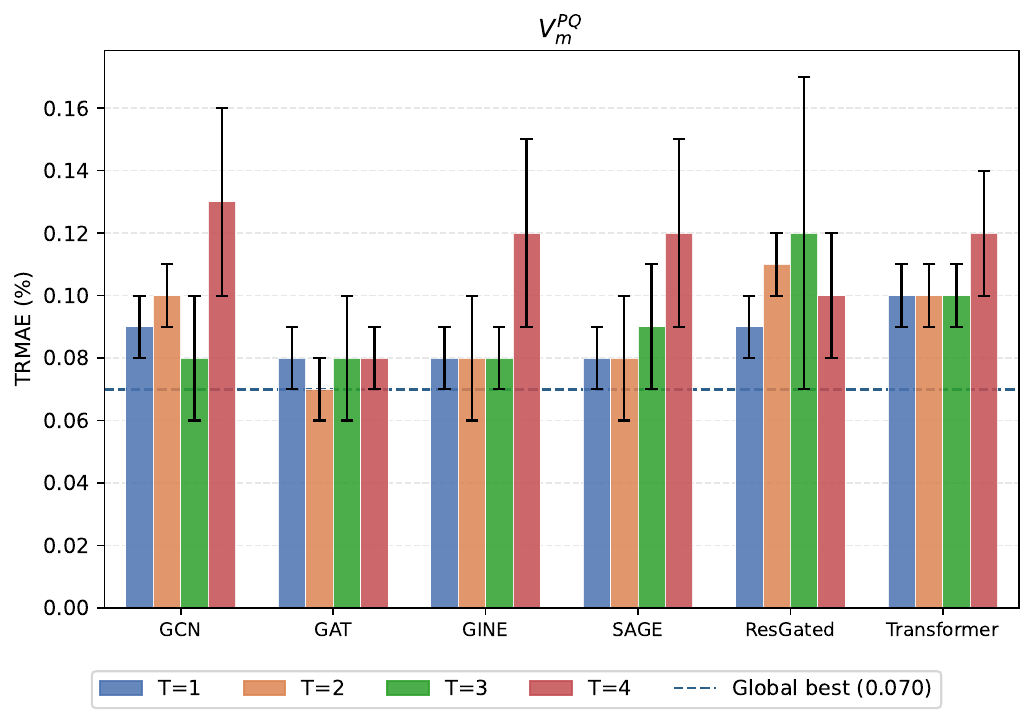}
        \caption{Shared weights}
        \label{fig:magnitude_shared}
    \end{subfigure}

    \caption{IEEE-118 -- Voltage magnitude TRMAE (\%) ($V_m^{PQ}$) vs.\ message-passing
             depth $T$. Dashed line indicates the global best value.
             Error bars denote one standard deviation over four seeds.}
    \label{fig:magnitude_depth}
\end{figure}

\begin{figure}[H]
    \centering
    \begin{subfigure}{0.8\linewidth}
        \includegraphics[width=\linewidth]{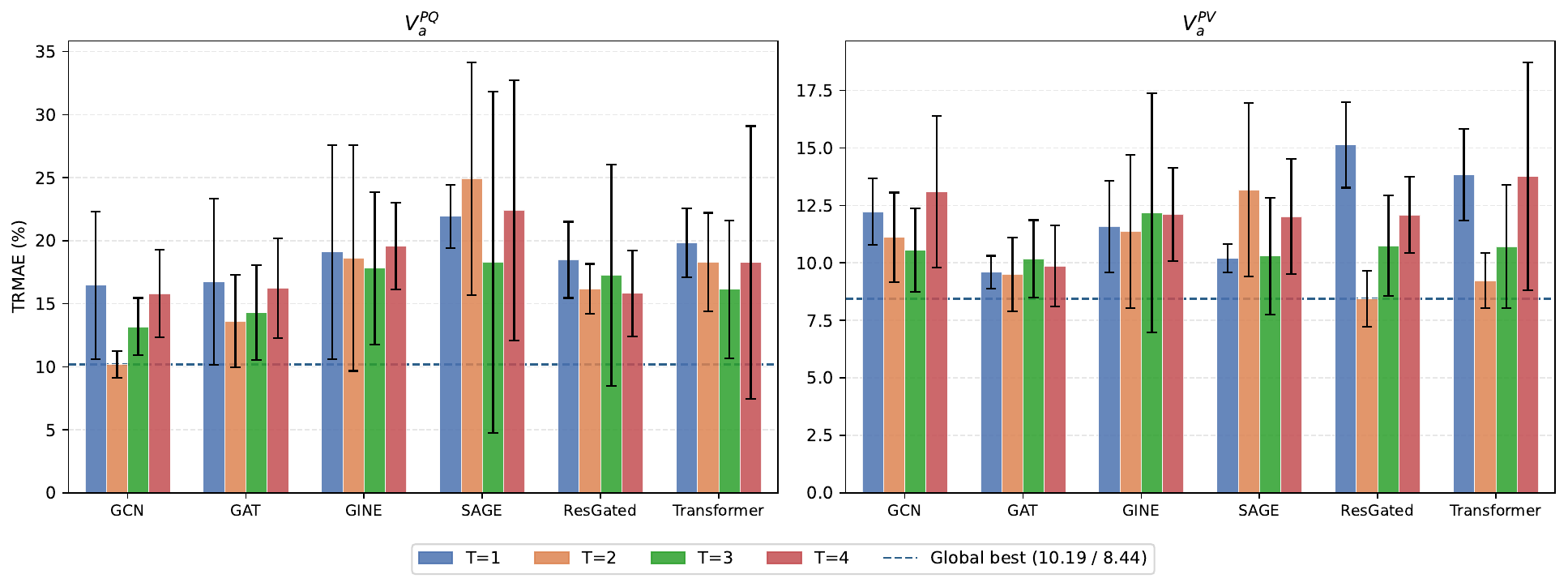}
        \caption{Independent weights}
        \label{fig:angles_independent}
    \end{subfigure}

    \vspace{0.8em}

    \begin{subfigure}{0.8\linewidth}
        \includegraphics[width=\linewidth]{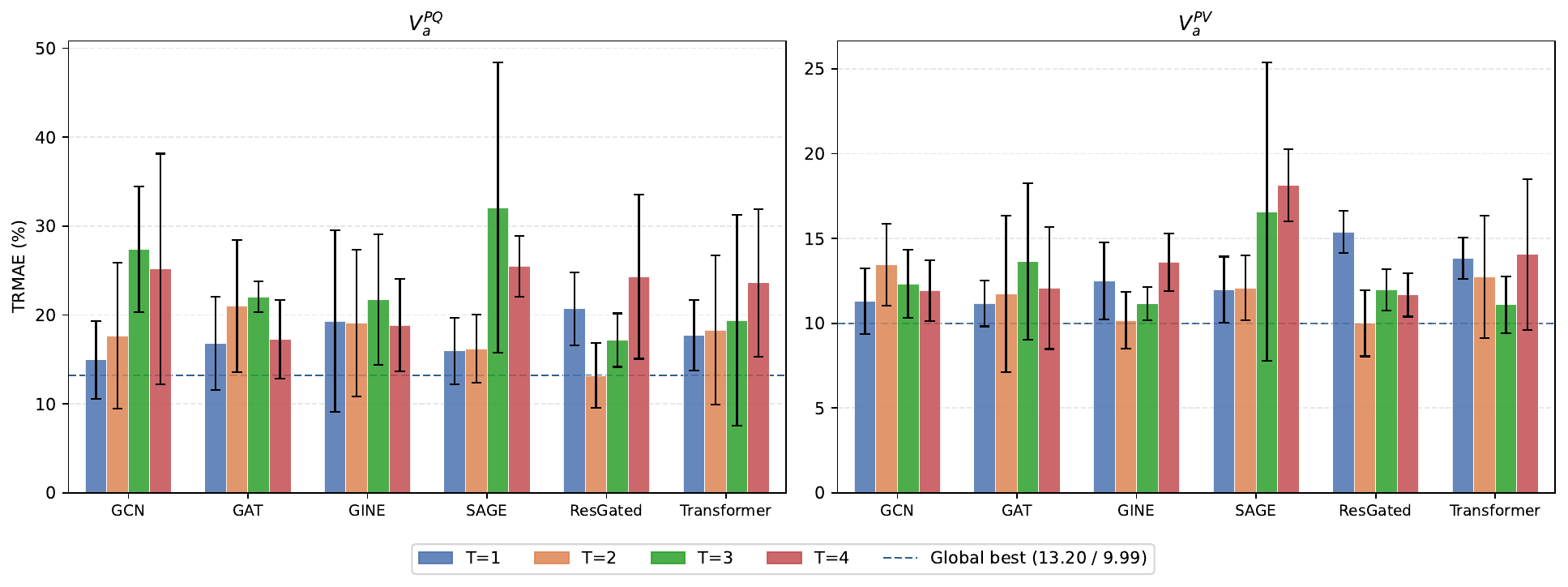}
        \caption{Shared weights}
        \label{fig:angles_shared}
    \end{subfigure}

    \caption{IEEE-118 -- Voltage angle TRMAE (\%) vs.\ message-passing depth $T$
             for $V_a^{PQ}$ (left) and $V_a^{PV}$ (right).
             Dashed line indicates the global best value per metric.
             Error bars denote one standard deviation over four seeds.}
    \label{fig:angles_depth}
\end{figure}

Figures \ref{fig:magnitude_depth} and \ref{fig:angles_depth} report the \gls{trmae} for voltage magnitude and angle predictions across varying message-passing depths and weight-sharing paradigms.

A consistent pattern across all models is that hierarchical feature extraction models (independent weights) outperform the recurrent unrolling ones (shared weights). Overall, using shared weights instead of independent layers degrades predictive performance for both target variables, yielding an average error increase of 16.5\% for voltage angles and 15.9\% for voltage magnitudes across all evaluated configurations. This performance gap indicates that independent layers provide the parameter flexibility required to learn specialized transformations at different receptive fields. Under the fixed $\sim$100k parameter budget — where increasing T necessarily narrows the hidden dimension — configurations with $T = 2$ or $T = 3$ generally provide the most favorable performance and widen the average gap between independent and shared configurations to roughly 20\%, whereas $T = 4$ frequently exhibits higher variability. Because width is traded off against depth in this protocol, this pattern should be read as evidence about capacity allocation (width vs. depth) rather than a claim that deeper propagation is intrinsically harmful.

For voltage magnitudes, the independent \gls{sage} architecture ($T=3$) and \gls{resgated} ($T=2$) achieve the lowest overall errors, both converging to a \gls{trmae} of $0.06\%$. Voltage angles prove to be a fundamentally harder target. Here, the independent \gls{resgated} model ($T=2$) achieves the best performance for \texttt{PV} bus angles, while the standard \gls{gcn} slightly outperforms others on \texttt{PQ} buses with an error of $10.19\%$. Notably, the \gls{sage} model exhibits high variance during angle prediction, struggling to stabilize the harder non-linear trigonometric mapping. From the statistical performance only, it is not possible to determine a clear winner, as the ranking changes between the quantities to predict and the bus types.

Crucially, the \gls{mlp} baseline completely fails to approximate both magnitudes and angles. Because \gls{acpf} equations are structurally defined by the sparse admittance matrix, the \gls{kcl} is a locally defined constraint at each bus. By treating the grid as a flat, fully connected vector, the \gls{mlp} attempts to learn dense correlations between electrically isolated buses, highlighting the need to exploit the graph topology to better approximate the \gls{acpf} problem.

\subsubsection*{Physical Consistency}
\begin{figure}[H]
    \centering
    \begin{subfigure}{0.8\linewidth}
        \includegraphics[width=\linewidth]{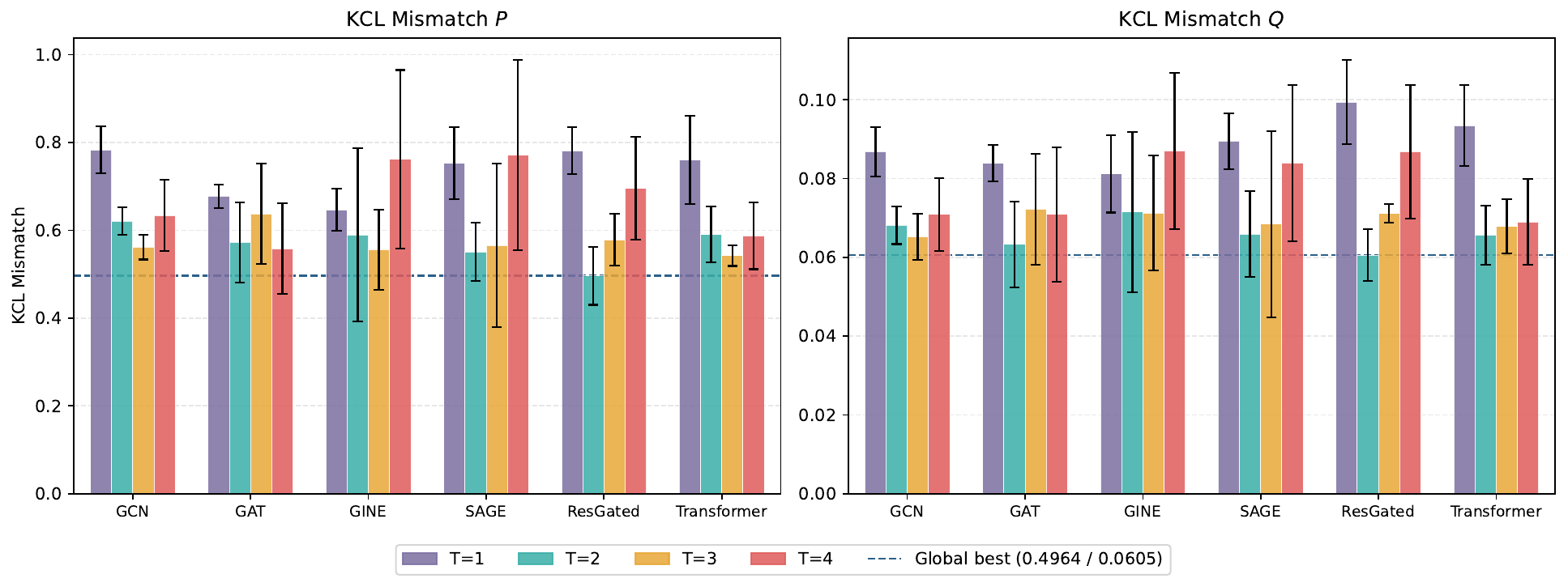}
        \caption{Independent weights}
        \label{fig:kcl_independent}
    \end{subfigure}

    \vspace{0.8em}

    \begin{subfigure}{0.8\linewidth}
        \includegraphics[width=\linewidth]{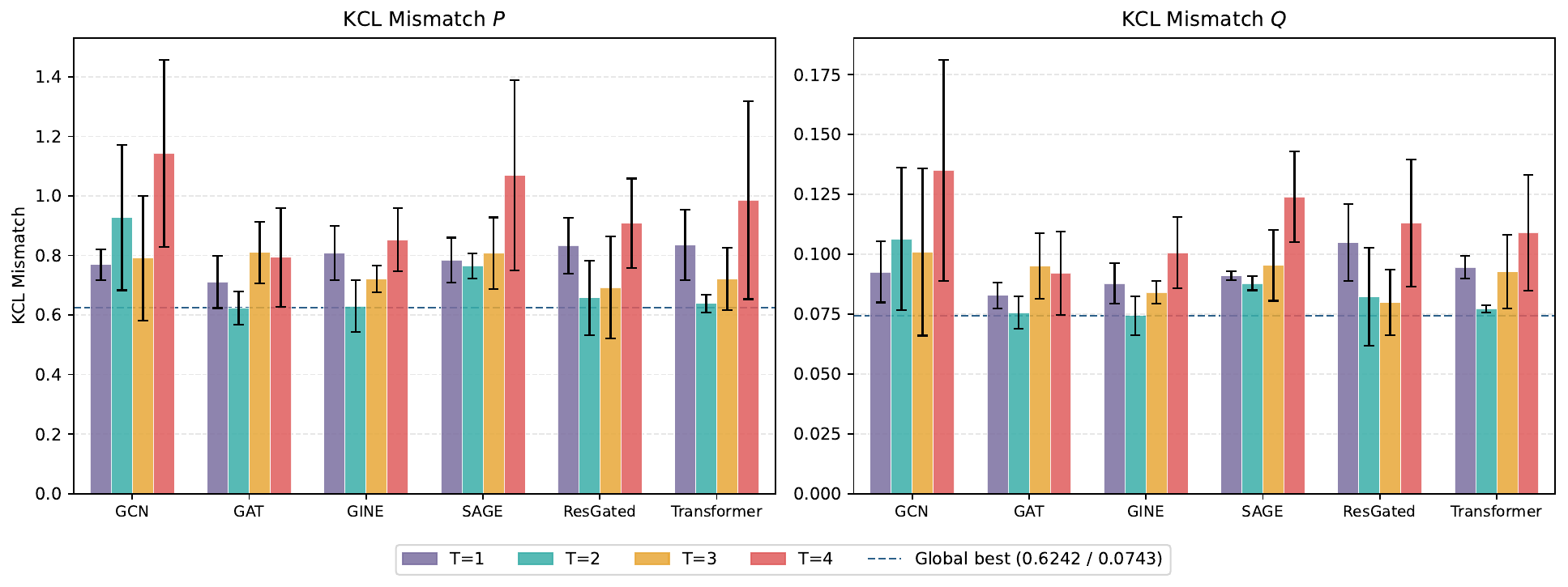}
        \caption{Shared weights}
        \label{fig:kcl_shared}
    \end{subfigure}

    \caption{IEEE-118 -- KCL mismatch vs.\ message-passing depth $T$
             for active power $P$ (left) and reactive power $Q$ (right).
             Dashed line indicates the global best value per metric.
             Error bars denote one standard deviation over four seeds.}
    \label{fig:kcl_depth}
\end{figure}

In Figure \ref{fig:kcl_depth} we report the \gls{kclmae} for both active ($\Delta P$) and reactive ($\Delta Q$) power, as statistical accuracy alone is not sufficient to guarantee physical compliance (see Section \ref{sec:metrics}).

Aligning with the statistical metrics, independent layers yield superior physical compliance compared to their shared counterparts. The independent \gls{resgated} architecture ($T=2$) emerges as the best overall performer, achieving the lowest active power mismatch of $0.4964$ \gls{pu}. This result highlights a fundamental architectural divide: isotropic models (like \gls{gcn} and \gls{sage}) treat all topological connections equally, incorporating line admittances merely as additive features. Conversely, \gls{resgated} is an \textit{anisotropic} model. It natively prioritizes certain paths by applying multiplicative scaling probabilities through a sigmoid gate. This operation mimics Ohm's Law, where current favors high-admittance paths. By explicitly gating message passing based on exact branch parameters, \gls{resgated} inherently produces lower empirical \gls{kcl} residuals far better than its isotropic counterparts.

The robustness of this model selection process is further examined in \ref{sec:trmae_kcl_correlation}, where we assess the differences across seeds. Based on the results, we select \gls{resgated} ($T=2$) for the subsequent diagnostic and ablation phases.

\subsection{Phase 2: Architectural Diagnostics}
\label{sec:phase2_results}

Because Phases 2 and 3 target the mechanisms underlying architectural differences, and because Phase 1 shows these differences are negligible on IEEE-14, we restrict the diagnostic and ablation analysis to IEEE-118 — the grid whose weak algebraic connectivity (Table \ref{tab:grid_topology}) makes propagation-related inductive biases most consequential. We treat the resulting findings as characterizing behavior under topological bottleneck conditions, not as topology-independent architectural universals.

\subsubsection*{Distance from the Slack Bus Analysis}
\label{sec:hop_analysis}

Accurate \gls{acpf} modeling requires capturing both local interactions and global constraints. While voltage magnitudes are largely driven by local reactive power balances, voltage angles are mathematically anchored to a single system-wide reference: the slack bus. To investigate how effectively \glspl{gnn} propagate this reference, we analyze the per-bus \gls{trmae} as a function of the shortest-path distance (in hops) from the slack bus, computed via Breadth-First Search. Results are reported for the IEEE-118 system using \(T=2\) message-passing layers, one of the two best-performing configurations in our benchmark. Since a 2-layer \gls{gnn} has a receptive field limited to two hops, buses farther from the slack bus cannot directly access the global reference, leaving the only global information spreading to the global pooling mechanism.

\begin{figure}[H]
    \centering

    \begin{subfigure}{0.46\textwidth}
        \centering
        \includegraphics[width=\linewidth]{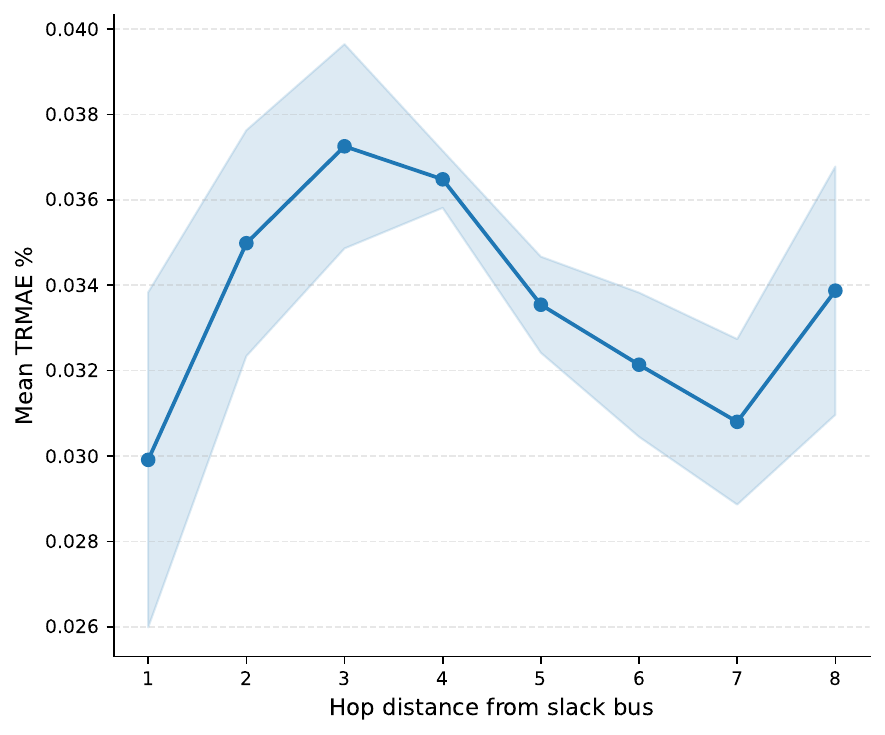}
        \caption{Voltage Magnitude}
        \label{fig:resgated_vm}
    \end{subfigure}
    \hfill
    \begin{subfigure}{0.46\textwidth}
        \centering
        \includegraphics[width=\linewidth]{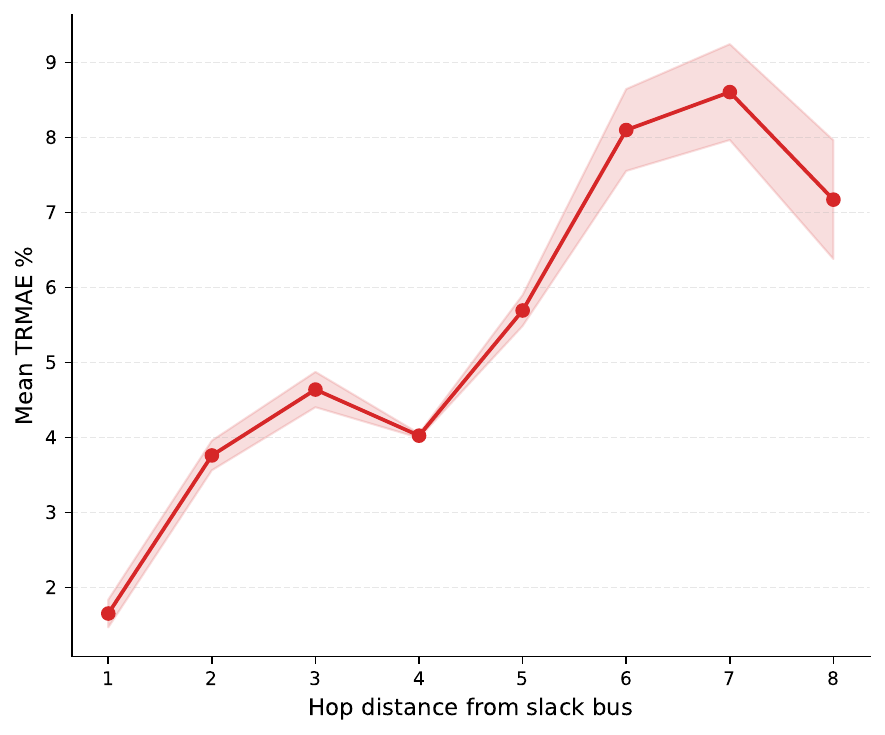}
        \caption{Voltage Angle}
        \label{fig:resgated_va}
    \end{subfigure}

    \caption{Performance of the \gls{resgated} model relative to hop distance from the slack bus on the IEEE-118 grid (\(T=2\), Independent layers). Results reflect the mean and variance across multiple seeds.}
    \label{fig:resgated_hop_error}
\end{figure}

Figure~\ref{fig:resgated_hop_error} shows the hop-wise error for the \gls{resgated} model, which is representative of all local message-passing architectures evaluated (results for the remaining models are reported in~\ref{app:appendix_b}). Voltage magnitude exhibits error variations (remaining below 0.05\%) and no systematic dependence on the distance from the slack bus, consistent with its predominantly local nature. In contrast, voltage angle error increases steadily with hop distance, becoming more pronounced beyond four hops. Given the network's diameter of 14 and average shortest path length of 6.31 (Table~\ref{tab:grid_topology}), this four-hop threshold falls well within the typical, not just the extreme, distance between buses—meaning the degradation affects a substantial share of the network rather than a handful of peripheral nodes. This trend highlights the inability of local message-passing architectures to propagate the global phase reference across distant nodes, leading to progressively larger prediction errors for voltage angle.

\begin{figure}[H]
    \centering

    \begin{subfigure}{0.49\textwidth}
        \centering
        \includegraphics[width=\linewidth]{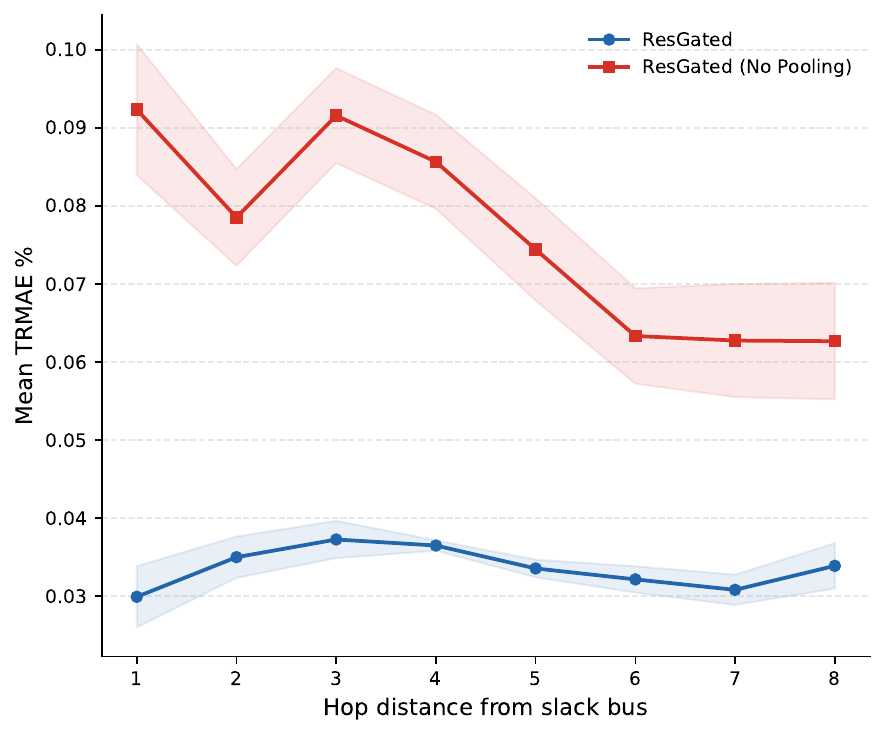}
        \caption{Voltage Magnitude}
        \label{fig:resgated_vm_ablation}
    \end{subfigure}
    \hfill
    \begin{subfigure}{0.49\textwidth}
        \centering
        \includegraphics[width=\linewidth]{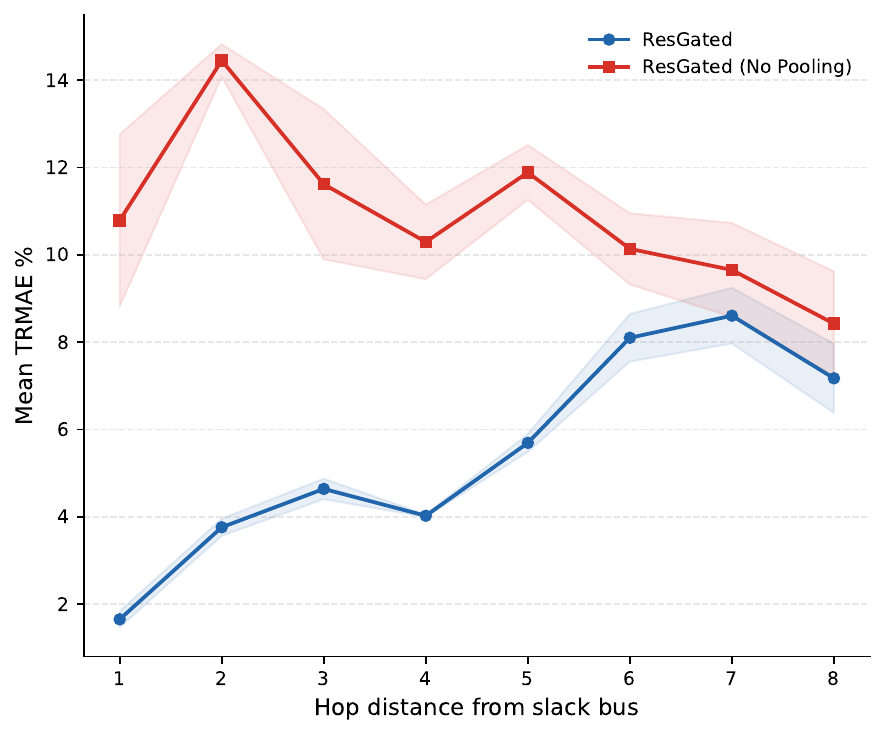}
        \caption{Voltage Angle}
        \label{fig:resgated_va_ablation}
    \end{subfigure}

    \caption{Performance of the \gls{resgated} model (baseline vs ablated without global pooling) relative to hop distance from the slack bus on the IEEE-118 grid (\(T=2\), Independent layers). Results reflect the mean and variance across multiple seeds.}
    \label{fig:resgated_hop_error_ablation}
\end{figure}

To isolate the specific contribution of global pooling to this behavior, we compare the baseline \gls{resgated} model against its no-global-pooling ablation across the same hop-distance axis (Figure~\ref{fig:resgated_hop_error_ablation}). Notably, the degradation is not confined to distant buses: for voltage angle, the ablated model's error rises sharply already at hop~1 (\(\approx\)11\% vs.\ \(\approx\)1.5\% for the baseline) and remains high, with substantially larger seed-to-seed variance, across all hop distances, while the relative gap to the baseline narrows rather than widens with increasing distance. This indicates that global pooling does not merely act as a long-range shortcut for buses beyond the local receptive field; it supplies a macroscopic reference signal that even topologically nearby buses rely on to resolve the globally coupled power flow equations. This is consistent with IEEE-118's markedly weak algebraic connectivity ($\lambda_1=0.009$, roughly 20$\times$ smaller than IEEE-14's; Table~\ref{tab:grid_topology}), which suggests a structural bottleneck in global information propagation.

\subsection{Phase 3: Ablation Study}

To quantify the individual contributions of key architectural choices, we dissect the best-performing \gls{resgated} model ($T=2$) on the IEEE-118 grid. As shown in Table~\ref{tab:ablation_results_ieee118}, evaluating performance across both statistical prediction errors (\gls{trmae}) and physical power flow mismatches ($\Delta P, \Delta Q$) reveals distinct physical and structural trade-offs. We report standardized effect sizes (Cohen's $d_z$) and bootstrap 95\% confidence intervals related to the ablated models in Table \ref{tab:effect_size_results_ieee118}.

The empirical results demonstrate that edge features and global pooling are fundamental to the network's architectural integrity. Omitting edge features deprives the model of explicit line impedance parameters needed to align with Ohm's law, causing voltage angle prediction errors ($V_a^{PQ}$) to jump to 24.17\%. This degradation is statistically robust, exhibiting large negative effect sizes on both active ($d_z = -1.99$) and reactive ($d_z = -2.38$) power mismatches with confidence intervals excluding zero. Even more critically, removing global pooling severely destabilizes physical power balance, driving the active power mismatch ($\Delta P$) up by nearly 87\% (from 0.4964 to 0.9292\,\gls{pu}). This represents the most massive, unambiguous degradation in the study ($d_z = -5.30$ for $\Delta P$ and $d_z = -5.67$ for $\Delta Q$). This aligns with the findings described in Section \ref{sec:phase2_results}, where the global pooling bridges the long-range dependencies of the network.

Conversely, removing residual connections yields a notable and highly robust improvement in physical constraint adherence, achieving the lowest active ($\Delta P = 0.3148$\,\gls{pu}) and reactive ($\Delta Q = 0.0381$\,\gls{pu}) power mismatches, as well as top voltage magnitude accuracy together with the ablated root connection ($V_m^{PQ} = 0.05\%$). This counterintuitive improvement is supported by enormous effect sizes ($d_z = 4.13$ for $\Delta P$, $d_z = 5.84$ for $\Delta Q$) and consistent signs across all random seeds. Similarly, removing the root connection also significantly improves \gls{kclmae} ($d_z = 2.39$ for $\Delta P$, $d_z = 3.39$ for $\Delta Q$). While residual shortcuts and root connections are standard in deep architectures to prevent representation collapse, they appear counterproductive in shallow power flow \glspl{gnn}. In this context, these connections impose an identity bias that forcibly adds unpropagated initial node state noise into the updated representations, slightly hindering the model's ability to satisfy non-linear physical power constraints by aggregating information from the neighboring nodes.

\begin{table}[ht]
\centering
\caption{Ablation study results on best performing model (\gls{resgated}, 2 Independent layers) for IEEE-118. \textbf{Bold} indicates the best result per metric.}
\label{tab:ablation_results_ieee118}
\scriptsize
\setlength{\tabcolsep}{4pt}
\begin{adjustbox}{max width=\linewidth}
\begin{tabular}{lrrrrr}
\toprule
Ablation & $V_a^{PQ}$ (\gls{trmae} \%) & $V_m^{PQ}$ (\gls{trmae} \%) & $V_a^{PV}$ (\gls{trmae} \%) & $\Delta P$ (\gls{pu}) & $\Delta Q$ (\gls{pu}) \\
\midrule
Baseline (\gls{resgated}) & 16.19 $\pm$ 1.97 & 0.06 $\pm$ 0.01 & 8.44 $\pm$ 1.22 & 0.4964 $\pm$ 0.0662 & 0.0605 $\pm$ 0.0066 \\
Aggregation: Max & 10.57 $\pm$ 3.19 & 0.07 $\pm$ 0.02 & \textbf{7.41 $\pm$ 0.32} & 0.3594 $\pm$ 0.0506 & 0.0435 $\pm$ 0.0070 \\
Aggregation: Mean & \textbf{9.78 $\pm$ 1.90} & 0.08 $\pm$ 0.02 & 8.73 $\pm$ 1.19 & 0.4848 $\pm$ 0.0489 & 0.0593 $\pm$ 0.0063 \\
No Edge Features & 24.17 $\pm$ 3.79 & 0.10 $\pm$ 0.01 & 17.36 $\pm$ 1.16 & 0.7442 $\pm$ 0.0807 & 0.0899 $\pm$ 0.0086 \\
No Global Pooling & 18.59 $\pm$ 6.23 & 0.12 $\pm$ 0.02 & 18.71 $\pm$ 1.59 & 0.9292 $\pm$ 0.0787 & 0.0976 $\pm$ 0.0048 \\
No Residuals & 10.26 $\pm$ 2.03 & \textbf{0.05 $\pm$ 0.01} & 8.66 $\pm$ 2.22 & \textbf{0.3148 $\pm$ 0.0234} & \textbf{0.0381 $\pm$ 0.0030} \\
No Root & 10.08 $\pm$ 2.76 & \textbf{0.05 $\pm$ 0.00} & 8.72 $\pm$ 1.47 & 0.3685 $\pm$ 0.0146 & 0.0437 $\pm$ 0.0018 \\
\bottomrule
\end{tabular}
\end{adjustbox}
\end{table}

Finally, evaluating alternative aggregation operators reveals a subtle trade-off between topological summation and local feature smoothing. Replacing the baseline \texttt{sum} aggregator with \texttt{mean} or \texttt{max} reduces voltage angle prediction errors---lowering $V_a^{PQ}$ to $9.78\%$ (\texttt{mean}) and $V_a^{PV}$ to $7.41\%$ (\texttt{max}). However, the effect size analysis reveals a divergence in physical consistency: while the \texttt{max} aggregator significantly improves \gls{kclmae} over the baseline ($d_z = 1.19$ for $\Delta P$), the \texttt{mean} aggregator produces only a small, non-robust effect ($d_z = 0.50$, with bootstrap intervals including zero). This suggests that while mean-based aggregation provides a stable representation for continuous nodal quantities, max-pooling may be uniquely suited for emphasizing dominant local power flows to satisfy \gls{kcl}.

\noindent\textbf{Synthesis \& Architectural Recommendations.} Taken together, these ablation findings suggest a set of design rules for graph-based \gls{acpf} modeling. Edge attributes and global pooling represent non-negotiable architectural anchors, essential for capturing local line admittance and propagating system-wide slack references, respectively. Conversely, for shallow graph models ($T=2$), stripping residual shortcuts and root connections reliably eliminates identity-bias noise, while adopting a \texttt{max} aggregator stabilizes continuous node states and physical consistency. This refined configuration optimizes both predictive accuracy ($V_a, V_m$) and physical constraint adherence ($\Delta P, \Delta Q$).

\section{Discussion}\label{sec:discussion}

Taken together, the three experimental phases do not identify a universally dominant architecture. Instead, they point to a set of design principles that appear well aligned with the structural characteristics of the~\gls{acpf} problem. Performance depends not only on the message-passing mechanism itself, but on how model capacity is allocated across representation width, propagation depth, local edge processing, and global context.

A first observation concerns the interaction between depth and width. Configurations with two or three layers tend to provide the most favorable balance, while deeper configurations show more variability. Since the hidden dimension shrinks as depth increases to preserve the fixed parameter budget, this does not isolate the causal effect of depth alone: it instead suggests that, when capacity is limited, investing it in wider node representations can be at least as valuable as extending the receptive field. This is plausible for power grids, where many interactions are local, and the remaining system-wide dependencies seem to be captured more efficiently through explicit global aggregation than through additional message-passing rounds.

Independent-layer configurations generally outperform their shared recurrent counterparts, though the magnitude of this gap fluctuates depending on propagation depth. The performance penalty of using shared weights is most pronounced in the best-performing 2- and 3-layer configurations, confirming that independent layers successfully leverage their flexibility to specialize transformations at different receptive fields, whereas a shared layer must apply the same rule regardless of how far information has already traveled. Since the shared configuration also introduces a~\gls{gru}-based update, however, this gap cannot be attributed to weight sharing alone.

The diagnostic analyses draw a clear qualitative line between the two predicted quantities. Voltage magnitude is small in error and largely insensitive to distance from the slack bus, consistent with it being governed mainly by local reactive-power interactions. Voltage angle, in contrast, degrades with distance and depends much more strongly on the presence of global pooling, since it is defined relative to a single, distant reference point that local, fixed-depth message passing cannot easily reach on its own. Because IEEE-118 has a substantially weaker algebraic connectivity than IEEE-14 (roughly 20$\times$ smaller), it is tempting to read this as a general rule: grids that are topologically bottlenecked relative to their size, rather than simply larger, may be the ones that benefit most from global context and moderate depth. We see this as a hypothesis worth testing on a wider range of topologies rather than a conclusion this two-grid comparison can support.

Edge features prove similarly important: removing them degrades both statistical accuracy and physical consistency together, reinforcing that topology alone under-specifies the electrical coupling between buses, and that line impedances are needed to approximate it. More broadly, our results suggest that statistical accuracy and physical consistency do not always move together -- several configurations reach comparable \gls{trmae} while differing noticeably in \gls{kclmae} -- which is why we treat \gls{kclmae}, and not \gls{trmae}, as the decisive criterion throughout this benchmark. This points out that standard regression metrics alone may not be a sufficient basis for selecting \gls{acpf} surrogates, and physics-aware metrics deserve comparable weight in model selection, suggesting the use of physics-aware models to obtain physically compliant results.

The ablation study adds a note of caution to otherwise intuitive design choices. Replacing sum aggregation with \texttt{mean} or \texttt{max}, and removing residual or root connections, improves several metrics rather than harming them. A plausible reading is that in a shallow, two-layer model, repeatedly re-injecting previous or initial node states can constrain the learned transformation more than it helps stabilize it. We would not extend this beyond the shallow regime tested here, since deeper or differently structured models may benefit from these components in ways our ablation was not designed to probe.

Overall, our results favor shallow-to-moderately-deep, edge-conditioned architectures with an explicit source of global context. At the same time, it should be interpreted within the scope of this study: two fixed IEEE topologies, in-distribution load perturbations, a fixed parameter budget that entangles depth and width, physical constraints assessed through post-hoc residuals rather than enforced during training, and results averaged over four seeds, for which we report effect sizes and bootstrap intervals as a robustness check but which remain too few to support fully powered, formal hypothesis testing. Within these bounds, we see the recurring alignment between architectural choices and physical structure as a meaningful signal rather than an artifact of any single phase.

\section{Conclusion}
\label{sec:conclusion}

This work presented a controlled benchmark of~\gls{gnn} design choices for~\gls{acpf} prediction. Rather than introducing a new architecture, the objective was to determine which message-passing and information-flow mechanisms are better aligned with the structural requirements of the problem. The evaluation combined a broad architectural comparison on the IEEE 14-bus and 118-bus systems, a structural diagnosis of the best-performing models, and a targeted ablation study.

Enforcing a common parameter budget across architectures was central to this design, ensuring that observed differences reflect the underlying message-passing mechanism rather than unequal representation capacity. As discussed in Section \ref{sec:discussion}, this choice also means that propagation depth and representation width cannot be fully disentangled, and the results are best read as a comparison of capacity-allocation strategies rather than an unconstrained search for each architecture's maximum performance.

Within this controlled setting, a fairly consistent picture emerged across all three phases. Anisotropic, edge-conditioned message passing -- exemplified by the ResGated architecture with two independent propagation layers -- offered the most favorable trade-off between statistical accuracy and physical consistency, particularly on the larger and more weakly-connected IEEE-118 grid. Edge features and global context aggregation proved to be the components the model could least afford to lose, while residual and root connections, though theoretically motivated, were not clearly beneficial at this shallow depth. We find this qualitative picture reasonably robust, since the same pattern emerged independently across the broad benchmark, the diagnostic analyses, and the ablation study, rather than resting on any single result.

Taken together, these findings translate into a small set of practical starting points for designing~\gls{gnn}-based~\gls{acpf} surrogates:
\begin{itemize}
    \item For grids exhibiting weak algebraic connectivity relative to size (as in IEEE-118), favor anisotropic, edge-conditioned message passing over isotropic, topology-only schemes when physical consistency is a priority according to our experimental protocol.
    \item Treat global context aggregation as necessary rather than optional, especially for voltage-angle prediction and for topologically weakly-connected grids.
    \item Evaluate architectural choices, including residual and root connections, against physical-consistency metrics rather than assuming their benefit from other domains.
    \item Report physics-aware metrics such as~\gls{kclmae} alongside standard statistical error, since the two do not always agree.
\end{itemize}
These points are intended as a structured starting point for narrowing the design space, rather than as fixed rules to be applied without validation on a given system.

As detailed in Section \ref{sec:discussion}, these guidelines were derived under a controlled setting: two fixed IEEE topologies, in-distribution load perturbations, a shared parameter budget, and physical constraints assessed through post-hoc residuals. Future work should test whether they extend to conditions explicitly outside this scope, including topology changes and $N-1$ contingencies, larger and differently structured grids, and physics-enforcing approaches that impose constraints directly during training rather than evaluating them post hoc and relying on pure supervised training. Increasing the number of random seeds would also improve the statistical robustness of the comparison, enabling formal hypothesis testing in addition to reporting confidence intervals and effect sizes. Additionally, complementing parameter-controlled comparisons with runtime and memory measurements would further support the transition from these architectural guidelines to deployment-oriented model selection. Finally, further investigating the effects of oversmoothing and oversquashing would provide a more complete understanding of the interaction between \glspl{gnn} and power grids.

\bibliographystyle{elsarticle-num}
\bibliography{references}

\appendix
\section{Extended Experimental Setup and Implementation}
\label{app:appendix_a}

\subsection{Model Architectures and Hyperparameters}

\begin{table}[H]
\centering
\caption{Unified Parameter Configurations Across Evaluated Architectures. Detailed specifications for all evaluated \glspl{gnn} and the \gls{mlp} baseline, showing the message-passing depth ($T$), hidden representation dimensions, and total parameter counts. Hidden dimensions were specifically scaled for each configuration to maintain a tightly controlled capacity budget of approximately 100,000 parameters across both independent and shared weight modes.}
\label{tab:model_architectures}
\scriptsize
\setlength{\tabcolsep}{4pt}

\begin{minipage}[t]{0.48\linewidth}
\centering
\begin{tabular}{llrrr}
\toprule
Model & Mode & $T$ & Hidden & Params \\
\midrule
GCN             & Indep.  &  1 &  78 & 100,698 \\
                &         &  2 &  60 &  99,780 \\
                &         &  3 &  51 & 101,031 \\
                &         &  4 &  45 & 101,160 \\
\cmidrule(lr){2-5}
                & Shared  & -- &  53 &  98,739 \\
\midrule
GAT             & Indep.  &  1 &  60 &  98,460 \\
                &         &  2 &  44 &  95,524 \\
                &         &  3 &  36 &  92,484 \\
                &         &  4 &  32 &  95,584 \\
\cmidrule(lr){2-5}
                & Shared  & -- &  48 & 105,840 \\
\midrule
GINE            & Indep.  &  1 &  93 &  99,141 \\
                &         &  2 &  75 &  98,856 \\
                &         &  3 &  65 &  99,979 \\
                &         &  4 &  58 & 100,062 \\
\cmidrule(lr){2-5}
                & Shared  & -- &  58 & 101,097 \\
\midrule
GraphSAGE            & Indep.  &  1 &  78 & 100,698 \\
                &         &  2 &  60 &  99,780 \\
                &         &  3 &  51 & 101,031 \\
                &         &  4 &  45 & 101,160 \\
\cmidrule(lr){2-5}
                & Shared  & -- &  53 &  98,739 \\
\bottomrule
\end{tabular}
\end{minipage}%
\hfill
\begin{minipage}[t]{0.48\linewidth}
\centering
\begin{tabular}{llrrr}
\toprule
Model & Mode & $T$ & Hidden & Params \\
\midrule
MLP             & Indep.  &  1 & 127 & 100,587 \\
                &         &  2 & 104 & 100,467 \\
                &         &  3 &  90 &  99,903 \\
                &         &  4 &  81 & 100,848 \\
\midrule
\gls{resgated}  & Indep.  &  1 &  64 & 101,056 \\
                &         &  2 &  48 & 101,136 \\
                &         &  3 &  40 & 100,920 \\
                &         &  4 &  35 & 100,730 \\
\cmidrule(lr){2-5}
                & Shared  & -- &  48 &  99,696 \\
\midrule
Transformer     & Indep.  &  1 &  52 &  93,236 \\
                &         &  2 &  40 & 101,560 \\
                &         &  3 &  32 &  94,816 \\
                &         &  4 &  32 & 124,256 \\
\cmidrule(lr){2-5}
                & Shared  & -- &  44 & 102,740 \\
\bottomrule
\end{tabular}
\end{minipage}
\end{table}

\subsection{Additional Architecture Implementation Specifications}
\label{app:phase1_components}

This subsection outlines model-specific implementation decisions and structural modifications optimized for electrical grid deployment.

\begin{itemize}[leftmargin=*]
    \item \textbf{GCN} \label{app:gcn_choices}: Preserves node self-identity via a dedicated per-node-type root transform $\mathbf{W}_{\text{root},\tau}$ instead of artificial self-loops (Section~\ref{sec:self_loops}). Edge features $\mathbf{e}_{ij}$ are scaled alongside normalized node messages: 
    
    $\mathbf{m}_{ij} = \tilde{d}_i^{-1/2}\tilde{d}_j^{-1/2}(\mathbf{W}_r \mathbf{h}_i + \mathbf{e}_{ij})$.

    \item \textbf{GraphSAGE} \label{app:sage_choices}: Applies the projection $\mathbf{W}_{\text{neigh},r}$ prior to edge slicing, reducing memory overhead from $\mathcal{O}(|\mathcal{E}|)$ to $\mathcal{O}(N)$. Edge features are injected directly ($\mathbf{m}_{ij} = \mathbf{W}_{\text{neigh},r}\mathbf{h}_i + \mathbf{e}_{ij}$), valid by mean-operator linearity. Native \texttt{mean} aggregation is overridden to \texttt{sum} in Phase~1; neighbor sampling is omitted due to small grid diameters.

    \item \textbf{GAT} \label{app:gat_choices}: Uses GATv2 \cite{brody2021attentive}\\ ($
    s_{vu} =
    \mathbf{a}^\top
    \operatorname{LeakyReLU}(
        \mathbf{W}_{\mathrm{src}}^{(l)} \mathbf{h}_v^{(l)}
        +
        \mathbf{W}_{\mathrm{dst}}^{(l)} \mathbf{h}_u^{(l)}
        +
        \mathbf{W}_{e}^{(l)} \mathbf{e}_{vu})$) with edge features \cite{wang2021egat} and negative slope $0.2$ \cite{velivckovic2017graph} to prevent query-independent static attention. Employs a single attention head, stores an attention score cache for post-hoc analysis, and uses a root transform $\mathbf{W}_{\text{root},\tau}$ outside the attention mechanism to preserve self-identity.

    \item \textbf{GINE} \label{app:gine_choices}: Learns an independent parameter $\varepsilon_\tau$ per bus type $\tau$ (PV, PQ, slack; initialized at $0$). Self-identity is intrinsically encoded via the $(1+\varepsilon_\tau)\mathbf{h}_j$ term without requiring root-weights or self-loops. Features use 
    
    $\operatorname{ReLU}(\mathbf{h}_i + \mathbf{e}_{ij})$ injection, with SiLU activations in downstream MLPs.

    \item \textbf{ResGated} \label{app:resgated_choices}: Computes dimension-wise edge gates based on joint endpoint states and line attributes: $\eta_{ij} = \sigma(\mathbf{W}_{\text{src},r}\mathbf{h}_i + \mathbf{W}_{\text{dst},r}\mathbf{h}_j + \mathbf{e}_{ij})$. Transforms node ego-states via a dedicated $\mathbf{U}_\tau$ matrix added as a residual connection.

    \item \textbf{Graph Transformer} \label{app:transformer_choices}: Restricts attention to physical grid topology ($\mathcal{O}(N)$ complexity) with a single head. Injects line features into both Key and Value projections ($\mathbf{k}_{ij} = \mathbf{W}_K \mathbf{h}_i + \mathbf{e}_{ij}$, $\mathbf{v}_{ij} = \mathbf{W}_V \mathbf{h}_i + \mathbf{e}_{ij}$). Relies on an external root transform $\mathbf{W}_{\text{root},\tau}$ and omits internal gated residuals ($\beta$), which are managed by the shared backbone.

    \item \textbf{MLP} \label{app:mlp_choices}: Acts as a topology-ablation baseline by implementing an identical API signature (\texttt{data}, \texttt{edge\_attr\_dict}, etc.) while ignoring edge structures. Applies $L$ independent linear layers per node type to match the message-passing depth of GNN variants. Finally, the number of parameters is tuned to match the 100k budget setting.
\end{itemize}

\section{Supplementary Experimental Results}
\label{app:appendix_b}

\subsection{Phase 1: Complete Results}

This subsection reports the complete results about statistical and physical performance from Phase 1 of the benchmark. We include the complete aggregated tables and the plots for a visual comparison.

\begin{landscape}
\begin{table}[ht]
\centering
\caption{PF TRMAE (\%) per bus type. \textbf{Bold} = best GNN per metric, dataset, and mode. \underline{\textbf{Underline}} = best overall per metric and dataset.}
\label{tab:pf_trmae}
\scriptsize
\setlength{\tabcolsep}{4pt}
\begin{adjustbox}{max width=\linewidth}
\begin{tabular}{llrrrrrrrrrrrr}
\toprule
 & & \multicolumn{4}{c}{$V_a^{PQ}$} & \multicolumn{4}{c}{$V_m^{PQ}$} & \multicolumn{4}{c}{$V_a^{PV}$} \\
\cmidrule(lr){3-6}
\cmidrule(lr){7-10}
\cmidrule(lr){11-14}
 & & \multicolumn{2}{c}{IEEE-14} & \multicolumn{2}{c}{IEEE-118} & \multicolumn{2}{c}{IEEE-14} & \multicolumn{2}{c}{IEEE-118} & \multicolumn{2}{c}{IEEE-14} & \multicolumn{2}{c}{IEEE-118} \\
\cmidrule(lr){3-4}
\cmidrule(lr){5-6}
\cmidrule(lr){7-8}
\cmidrule(lr){9-10}
\cmidrule(lr){11-12}
\cmidrule(lr){13-14}
Model & $T$ & Independent & Shared & Independent & Shared & Independent & Shared & Independent & Shared & Independent & Shared & Independent & Shared \\
\midrule
\multirow{4}{*}{MLP}
& T1 & 16.14 $\pm$ 0.46 & -- & 292.65 $\pm$ 24.92 & -- & 2.56 $\pm$ 0.00 & -- & 2.95 $\pm$ 0.00 & -- & 30.05 $\pm$ 1.84 & -- & 173.24 $\pm$ 19.11 & -- \\
& T2 & 16.56 $\pm$ 0.60 & -- & 217.04 $\pm$ 18.99 & -- & 2.56 $\pm$ 0.00 & -- & 2.94 $\pm$ 0.00 & -- & 27.98 $\pm$ 0.82 & -- & 152.05 $\pm$ 5.26 & -- \\
& T3 & 16.86 $\pm$ 0.58 & -- & 218.86 $\pm$ 28.54 & -- & 2.56 $\pm$ 0.00 & -- & 2.94 $\pm$ 0.00 & -- & 28.38 $\pm$ 1.51 & -- & 143.93 $\pm$ 4.45 & -- \\
& T4 & 16.57 $\pm$ 1.06 & -- & 218.88 $\pm$ 36.29 & -- & 2.56 $\pm$ 0.00 & -- & 2.94 $\pm$ 0.01 & -- & 32.35 $\pm$ 1.68 & -- & 165.11 $\pm$ 15.16 & -- \\
\midrule
\multirow{4}{*}{GCN}
& T1 & 1.59 $\pm$ 0.30 & 1.82 $\pm$ 0.67 & 16.46 $\pm$ 5.87 & 14.96 $\pm$ 4.35 & 0.05 $\pm$ 0.01 & 0.05 $\pm$ 0.02 & 0.09 $\pm$ 0.01 & 0.09 $\pm$ 0.01 & 2.85 $\pm$ 0.90 & 2.91 $\pm$ 0.65 & 12.22 $\pm$ 1.45 & 11.30 $\pm$ 1.92 \\
& T2 & 1.02 $\pm$ 0.15 & 1.96 $\pm$ 1.26 & \underline{\textbf{10.19 $\pm$ 1.07}} & 17.68 $\pm$ 8.17 & 0.04 $\pm$ 0.01 & 0.05 $\pm$ 0.01 & 0.08 $\pm$ 0.02 & 0.10 $\pm$ 0.01 & 2.15 $\pm$ 0.76 & 2.86 $\pm$ 1.75 & 11.11 $\pm$ 1.95 & 13.45 $\pm$ 2.41 \\
& T3 & 0.93 $\pm$ 0.27 & 1.33 $\pm$ 0.57 & 13.18 $\pm$ 2.28 & 27.38 $\pm$ 7.07 & \underline{\textbf{0.03 $\pm$ 0.02}} & 0.05 $\pm$ 0.02 & 0.08 $\pm$ 0.04 & 0.08 $\pm$ 0.02 & 2.30 $\pm$ 0.69 & 3.17 $\pm$ 1.25 & 10.55 $\pm$ 1.82 & 12.32 $\pm$ 2.01 \\
& T4 & 1.31 $\pm$ 0.17 & 1.58 $\pm$ 0.58 & 15.79 $\pm$ 3.47 & 25.18 $\pm$ 12.98 & 0.04 $\pm$ 0.01 & 0.05 $\pm$ 0.00 & 0.09 $\pm$ 0.02 & 0.13 $\pm$ 0.03 & 2.84 $\pm$ 0.76 & 3.35 $\pm$ 0.81 & 13.09 $\pm$ 3.29 & 11.93 $\pm$ 1.79 \\
\midrule
\multirow{4}{*}{GAT}
& T1 & 1.31 $\pm$ 0.24 & 1.19 $\pm$ 0.16 & 16.74 $\pm$ 6.59 & 16.79 $\pm$ 5.23 & 0.04 $\pm$ 0.01 & \underline{\textbf{0.03 $\pm$ 0.01}} & 0.08 $\pm$ 0.00 & 0.08 $\pm$ 0.01 & 2.69 $\pm$ 0.51 & 2.26 $\pm$ 0.21 & 9.59 $\pm$ 0.72 & 11.17 $\pm$ 1.35 \\
& T2 & 1.13 $\pm$ 0.16 & \textbf{0.89 $\pm$ 0.13} & 13.63 $\pm$ 3.68 & 21.02 $\pm$ 7.43 & 0.06 $\pm$ 0.02 & \underline{\textbf{0.03 $\pm$ 0.00}} & 0.07 $\pm$ 0.01 & \textbf{0.07 $\pm$ 0.01} & 2.16 $\pm$ 0.63 & \underline{\textbf{1.49 $\pm$ 0.27}} & 9.50 $\pm$ 1.59 & 11.73 $\pm$ 4.63 \\
& T3 & 1.26 $\pm$ 0.19 & 1.50 $\pm$ 0.80 & 14.30 $\pm$ 3.77 & 22.06 $\pm$ 1.70 & 0.04 $\pm$ 0.01 & 0.05 $\pm$ 0.02 & 0.07 $\pm$ 0.02 & 0.08 $\pm$ 0.02 & 2.21 $\pm$ 1.05 & 3.27 $\pm$ 1.80 & 10.17 $\pm$ 1.69 & 13.64 $\pm$ 4.60 \\
& T4 & 1.71 $\pm$ 0.41 & 1.59 $\pm$ 0.71 & 16.23 $\pm$ 3.95 & 17.25 $\pm$ 4.42 & 0.05 $\pm$ 0.01 & 0.07 $\pm$ 0.04 & 0.09 $\pm$ 0.03 & 0.08 $\pm$ 0.01 & 2.35 $\pm$ 0.87 & 3.57 $\pm$ 0.89 & 9.87 $\pm$ 1.75 & 12.07 $\pm$ 3.59 \\
\midrule
\multirow{4}{*}{GINE}
& T1 & 1.63 $\pm$ 0.36 & 1.58 $\pm$ 0.28 & 19.11 $\pm$ 8.49 & 19.33 $\pm$ 10.18 & 0.06 $\pm$ 0.02 & 0.04 $\pm$ 0.01 & 0.09 $\pm$ 0.03 & 0.08 $\pm$ 0.01 & 4.66 $\pm$ 2.03 & 2.47 $\pm$ 0.35 & 11.58 $\pm$ 1.98 & 12.49 $\pm$ 2.27 \\
& T2 & 1.02 $\pm$ 0.27 & 1.17 $\pm$ 0.28 & 18.62 $\pm$ 8.95 & 19.11 $\pm$ 8.23 & 0.04 $\pm$ 0.02 & 0.05 $\pm$ 0.01 & 0.10 $\pm$ 0.03 & 0.08 $\pm$ 0.02 & 3.49 $\pm$ 0.73 & 2.09 $\pm$ 0.54 & 11.37 $\pm$ 3.33 & 10.16 $\pm$ 1.67 \\
& T3 & 1.05 $\pm$ 0.16 & 1.89 $\pm$ 0.67 & 17.82 $\pm$ 6.04 & 21.72 $\pm$ 7.34 & 0.04 $\pm$ 0.02 & 0.06 $\pm$ 0.02 & 0.11 $\pm$ 0.06 & 0.08 $\pm$ 0.01 & 2.19 $\pm$ 0.66 & 3.13 $\pm$ 1.62 & 12.17 $\pm$ 5.20 & 11.14 $\pm$ 0.98 \\
& T4 & 1.09 $\pm$ 0.27 & 1.66 $\pm$ 0.73 & 19.57 $\pm$ 3.44 & 18.86 $\pm$ 5.22 & 0.05 $\pm$ 0.02 & 0.07 $\pm$ 0.02 & 0.09 $\pm$ 0.01 & 0.12 $\pm$ 0.03 & 2.52 $\pm$ 0.91 & 3.24 $\pm$ 0.89 & 12.12 $\pm$ 2.03 & 13.60 $\pm$ 1.70 \\
\midrule
\multirow{4}{*}{ResGated}
& T1 & 1.88 $\pm$ 0.52 & 1.51 $\pm$ 0.08 & 18.48 $\pm$ 3.02 & 20.72 $\pm$ 4.11 & 0.04 $\pm$ 0.01 & 0.05 $\pm$ 0.01 & 0.10 $\pm$ 0.02 & 0.09 $\pm$ 0.01 & 3.80 $\pm$ 2.52 & 2.33 $\pm$ 0.46 & 15.13 $\pm$ 1.86 & 15.39 $\pm$ 1.24 \\
& T2 & 0.89 $\pm$ 0.19 & 1.49 $\pm$ 0.70 & 16.19 $\pm$ 1.97 & \textbf{13.20 $\pm$ 3.66} & 0.04 $\pm$ 0.01 & 0.04 $\pm$ 0.01 & \underline{\textbf{0.06 $\pm$ 0.01}} & 0.11 $\pm$ 0.01 & 1.89 $\pm$ 0.93 & 2.85 $\pm$ 0.56 & \underline{\textbf{8.44 $\pm$ 1.22}} & \textbf{9.99 $\pm$ 1.94} \\
& T3 & 0.86 $\pm$ 0.24 & 1.45 $\pm$ 0.30 & 17.27 $\pm$ 8.78 & 17.18 $\pm$ 3.01 & \underline{\textbf{0.03 $\pm$ 0.00}} & 0.06 $\pm$ 0.01 & 0.09 $\pm$ 0.01 & 0.12 $\pm$ 0.05 & 1.68 $\pm$ 0.66 & 2.96 $\pm$ 0.19 & 10.75 $\pm$ 2.19 & 11.98 $\pm$ 1.21 \\
& T4 & 0.87 $\pm$ 0.28 & 1.52 $\pm$ 0.50 & 15.83 $\pm$ 3.41 & 24.31 $\pm$ 9.23 & 0.04 $\pm$ 0.03 & 0.08 $\pm$ 0.03 & 0.10 $\pm$ 0.04 & 0.10 $\pm$ 0.02 & 2.54 $\pm$ 1.30 & 4.93 $\pm$ 1.62 & 12.09 $\pm$ 1.67 & 11.68 $\pm$ 1.29 \\
\midrule
\multirow{4}{*}{GraphSAGE}
& T1 & 1.38 $\pm$ 0.19 & 1.49 $\pm$ 0.08 & 21.93 $\pm$ 2.52 & 15.96 $\pm$ 3.71 & 0.05 $\pm$ 0.02 & 0.06 $\pm$ 0.01 & 0.09 $\pm$ 0.02 & 0.08 $\pm$ 0.01 & 2.90 $\pm$ 1.01 & 2.61 $\pm$ 0.36 & 10.20 $\pm$ 0.62 & 11.98 $\pm$ 1.95 \\
& T2 & 1.16 $\pm$ 0.29 & 0.91 $\pm$ 0.10 & 24.92 $\pm$ 9.22 & 16.22 $\pm$ 3.85 & 0.05 $\pm$ 0.03 & 0.05 $\pm$ 0.02 & 0.08 $\pm$ 0.02 & 0.08 $\pm$ 0.02 & 2.43 $\pm$ 1.31 & 2.36 $\pm$ 0.93 & 13.17 $\pm$ 3.77 & 12.07 $\pm$ 1.91 \\
& T3 & 1.47 $\pm$ 0.76 & 1.07 $\pm$ 0.28 & 18.30 $\pm$ 13.54 & 32.08 $\pm$ 16.34 & 0.04 $\pm$ 0.01 & 0.05 $\pm$ 0.03 & \underline{\textbf{0.06 $\pm$ 0.01}} & 0.09 $\pm$ 0.02 & 2.89 $\pm$ 0.44 & 2.27 $\pm$ 0.49 & 10.30 $\pm$ 2.54 & 16.58 $\pm$ 8.80 \\
& T4 & 1.87 $\pm$ 0.62 & 1.40 $\pm$ 0.26 & 22.43 $\pm$ 10.32 & 25.48 $\pm$ 3.39 & 0.08 $\pm$ 0.04 & 0.08 $\pm$ 0.02 & 0.07 $\pm$ 0.01 & 0.12 $\pm$ 0.03 & 5.31 $\pm$ 1.24 & 3.15 $\pm$ 1.48 & 12.02 $\pm$ 2.49 & 18.12 $\pm$ 2.13 \\
\midrule
\multirow{4}{*}{Transformer}
& T1 & 1.72 $\pm$ 0.28 & 2.45 $\pm$ 0.27 & 19.83 $\pm$ 2.72 & 17.71 $\pm$ 3.98 & 0.05 $\pm$ 0.01 & 0.07 $\pm$ 0.00 & 0.10 $\pm$ 0.01 & 0.10 $\pm$ 0.01 & 4.99 $\pm$ 1.74 & 4.25 $\pm$ 1.36 & 13.83 $\pm$ 2.00 & 13.83 $\pm$ 1.23 \\
& T2 & 1.08 $\pm$ 0.34 & 2.12 $\pm$ 0.88 & 18.31 $\pm$ 3.90 & 18.29 $\pm$ 8.40 & \underline{\textbf{0.03 $\pm$ 0.00}} & 0.06 $\pm$ 0.02 & 0.07 $\pm$ 0.01 & 0.10 $\pm$ 0.01 & \textbf{1.51 $\pm$ 0.24} & 4.27 $\pm$ 1.40 & 9.23 $\pm$ 1.21 & 12.73 $\pm$ 3.62 \\
& T3 & 0.86 $\pm$ 0.12 & 1.79 $\pm$ 0.70 & 16.14 $\pm$ 5.45 & 19.41 $\pm$ 11.87 & 0.04 $\pm$ 0.02 & 0.05 $\pm$ 0.03 & 0.08 $\pm$ 0.02 & 0.10 $\pm$ 0.01 & 2.14 $\pm$ 0.62 & 2.88 $\pm$ 0.78 & 10.71 $\pm$ 2.68 & 11.09 $\pm$ 1.66 \\
& T4 & \underline{\textbf{0.85 $\pm$ 0.12}} & 1.26 $\pm$ 0.30 & 18.28 $\pm$ 10.81 & 23.63 $\pm$ 8.28 & \underline{\textbf{0.03 $\pm$ 0.01}} & 0.04 $\pm$ 0.01 & 0.08 $\pm$ 0.01 & 0.12 $\pm$ 0.02 & 2.06 $\pm$ 0.32 & 3.27 $\pm$ 0.85 & 13.76 $\pm$ 4.96 & 14.06 $\pm$ 4.45 \\
\bottomrule
\end{tabular}
\end{adjustbox}
\end{table}
\end{landscape}

\begin{landscape}
\begin{table}[ht]
\centering
\caption{PF MAE KCL Mismatch ($\Delta P$, $\Delta Q$) across datasets, modes, and $T$. \textbf{Bold} = best GNN (model, $T$, mode) per metric and dataset. \underline{\textbf{Underline}} = best overall per metric and dataset.}
\label{tab:pf_kcl}
\scriptsize
\setlength{\tabcolsep}{4pt}
\begin{adjustbox}{max width=\linewidth}
\begin{tabular}{llrrrrrrrr}
\toprule
 & & \multicolumn{4}{c}{$\Delta P$} & \multicolumn{4}{c}{$\Delta Q$} \\
\cmidrule(lr){3-6}\cmidrule(lr){7-10}
 & & \multicolumn{2}{c}{IEEE-14} & \multicolumn{2}{c}{IEEE-118} & \multicolumn{2}{c}{IEEE-14} & \multicolumn{2}{c}{IEEE-118} \\
\cmidrule(lr){3-4}
\cmidrule(lr){5-6}
\cmidrule(lr){7-8}
\cmidrule(lr){9-10}
Model & $T$ & Independent & Shared & Independent & Shared & Independent & Shared & Independent & Shared \\
\midrule
\multirow{4}{*}{MLP}
& T1 & 0.3273 $\pm$ 0.0205 & -- & 14.8493 $\pm$ 0.1968 & -- & 0.1040 $\pm$ 0.0042 & -- & 3.3410 $\pm$ 0.0339 & -- \\
& T2 & 0.3243 $\pm$ 0.0163 & -- & 14.3881 $\pm$ 0.1904 & -- & 0.1022 $\pm$ 0.0047 & -- & 3.3894 $\pm$ 0.1626 & -- \\
& T3 & 0.3446 $\pm$ 0.0215 & -- & 14.4344 $\pm$ 0.1540 & -- & 0.0998 $\pm$ 0.0022 & -- & 3.4607 $\pm$ 0.0635 & -- \\
& T4 & 0.3203 $\pm$ 0.0184 & -- & 14.5430 $\pm$ 0.2336 & -- & 0.0968 $\pm$ 0.0007 & -- & 3.3823 $\pm$ 0.0561 & -- \\
\midrule
\multirow{4}{*}{GCN}
& T1 & 0.0267 $\pm$ 0.0010 & 0.0319 $\pm$ 0.0065 & 0.7834 $\pm$ 0.0529 & 0.7693 $\pm$ 0.0513 & 0.0074 $\pm$ 0.0011 & 0.0068 $\pm$ 0.0021 & 0.0868 $\pm$ 0.0063 & 0.0926 $\pm$ 0.0127 \\
& T2 & 0.0215 $\pm$ 0.0015 & 0.0318 $\pm$ 0.0201 & 0.6206 $\pm$ 0.0311 & 0.9277 $\pm$ 0.2445 & 0.0060 $\pm$ 0.0008 & 0.0082 $\pm$ 0.0007 & 0.0681 $\pm$ 0.0048 & 0.1064 $\pm$ 0.0298 \\
& T3 & 0.0169 $\pm$ 0.0038 & 0.0224 $\pm$ 0.0028 & 0.5618 $\pm$ 0.0285 & 0.7908 $\pm$ 0.2101 & 0.0060 $\pm$ 0.0008 & 0.0071 $\pm$ 0.0013 & 0.0652 $\pm$ 0.0059 & 0.1009 $\pm$ 0.0349 \\
& T4 & 0.0239 $\pm$ 0.0053 & 0.0369 $\pm$ 0.0150 & 0.6339 $\pm$ 0.0806 & 1.1429 $\pm$ 0.3138 & 0.0072 $\pm$ 0.0015 & 0.0092 $\pm$ 0.0027 & 0.0709 $\pm$ 0.0092 & 0.1350 $\pm$ 0.0461 \\
\midrule
\multirow{4}{*}{GAT}
& T1 & 0.0272 $\pm$ 0.0029 & 0.0293 $\pm$ 0.0056 & 0.6775 $\pm$ 0.0264 & 0.7112 $\pm$ 0.0879 & 0.0064 $\pm$ 0.0020 & 0.0055 $\pm$ 0.0005 & 0.0839 $\pm$ 0.0046 & 0.0828 $\pm$ 0.0054 \\
& T2 & 0.0181 $\pm$ 0.0028 & \textbf{0.0146 $\pm$ 0.0020} & 0.5726 $\pm$ 0.0915 & \textbf{0.6242 $\pm$ 0.0560} & 0.0070 $\pm$ 0.0036 & \textbf{0.0042 $\pm$ 0.0005} & 0.0633 $\pm$ 0.0109 & 0.0755 $\pm$ 0.0067 \\
& T3 & 0.0171 $\pm$ 0.0037 & 0.0240 $\pm$ 0.0089 & 0.6375 $\pm$ 0.1145 & 0.8104 $\pm$ 0.1037 & 0.0046 $\pm$ 0.0006 & 0.0083 $\pm$ 0.0025 & 0.0722 $\pm$ 0.0141 & 0.0950 $\pm$ 0.0136 \\
& T4 & 0.0235 $\pm$ 0.0045 & 0.0291 $\pm$ 0.0046 & 0.5584 $\pm$ 0.1032 & 0.7941 $\pm$ 0.1655 & 0.0064 $\pm$ 0.0028 & 0.0081 $\pm$ 0.0013 & 0.0709 $\pm$ 0.0171 & 0.0921 $\pm$ 0.0174 \\
\midrule
\multirow{4}{*}{GINE}
& T1 & 0.0285 $\pm$ 0.0045 & 0.0281 $\pm$ 0.0012 & 0.6466 $\pm$ 0.0474 & 0.8079 $\pm$ 0.0904 & 0.0070 $\pm$ 0.0015 & 0.0065 $\pm$ 0.0008 & 0.0812 $\pm$ 0.0098 & 0.0878 $\pm$ 0.0086 \\
& T2 & 0.0193 $\pm$ 0.0039 & 0.0233 $\pm$ 0.0083 & 0.5896 $\pm$ 0.1971 & 0.6298 $\pm$ 0.0869 & 0.0054 $\pm$ 0.0019 & 0.0085 $\pm$ 0.0038 & 0.0715 $\pm$ 0.0203 & \textbf{0.0743 $\pm$ 0.0080} \\
& T3 & 0.0180 $\pm$ 0.0065 & 0.0271 $\pm$ 0.0116 & 0.5557 $\pm$ 0.0919 & 0.7206 $\pm$ 0.0454 & 0.0053 $\pm$ 0.0015 & 0.0088 $\pm$ 0.0038 & 0.0713 $\pm$ 0.0145 & 0.0841 $\pm$ 0.0047 \\
& T4 & 0.0208 $\pm$ 0.0047 & 0.0326 $\pm$ 0.0088 & 0.7620 $\pm$ 0.2029 & 0.8529 $\pm$ 0.1054 & 0.0070 $\pm$ 0.0015 & 0.0091 $\pm$ 0.0026 & 0.0870 $\pm$ 0.0199 & 0.1006 $\pm$ 0.0150 \\
\midrule
\multirow{4}{*}{ResGated}
& T1 & 0.0349 $\pm$ 0.0061 & 0.0366 $\pm$ 0.0045 & 0.7812 $\pm$ 0.0533 & 0.8328 $\pm$ 0.0928 & 0.0080 $\pm$ 0.0011 & 0.0080 $\pm$ 0.0007 & 0.0994 $\pm$ 0.0107 & 0.1049 $\pm$ 0.0162 \\
& T2 & 0.0157 $\pm$ 0.0028 & 0.0235 $\pm$ 0.0070 & \underline{\textbf{0.4964 $\pm$ 0.0662}} & 0.6576 $\pm$ 0.1242 & 0.0050 $\pm$ 0.0013 & 0.0057 $\pm$ 0.0015 & \underline{\textbf{0.0605 $\pm$ 0.0066}} & 0.0822 $\pm$ 0.0205 \\
& T3 & 0.0145 $\pm$ 0.0063 & 0.0221 $\pm$ 0.0031 & 0.5778 $\pm$ 0.0591 & 0.6926 $\pm$ 0.1706 & 0.0047 $\pm$ 0.0018 & 0.0063 $\pm$ 0.0020 & 0.0712 $\pm$ 0.0023 & 0.0798 $\pm$ 0.0137 \\
& T4 & 0.0165 $\pm$ 0.0040 & 0.0309 $\pm$ 0.0023 & 0.6951 $\pm$ 0.1169 & 0.9079 $\pm$ 0.1506 & 0.0059 $\pm$ 0.0017 & 0.0092 $\pm$ 0.0011 & 0.0868 $\pm$ 0.0169 & 0.1130 $\pm$ 0.0266 \\
\midrule
\multirow{4}{*}{GraphSAGE}
& T1 & 0.0291 $\pm$ 0.0026 & 0.0315 $\pm$ 0.0056 & 0.7528 $\pm$ 0.0820 & 0.7845 $\pm$ 0.0756 & 0.0062 $\pm$ 0.0007 & 0.0069 $\pm$ 0.0016 & 0.0895 $\pm$ 0.0071 & 0.0910 $\pm$ 0.0020 \\
& T2 & 0.0174 $\pm$ 0.0025 & 0.0189 $\pm$ 0.0018 & 0.5506 $\pm$ 0.0666 & 0.7651 $\pm$ 0.0426 & 0.0056 $\pm$ 0.0027 & 0.0068 $\pm$ 0.0033 & 0.0659 $\pm$ 0.0109 & 0.0878 $\pm$ 0.0029 \\
& T3 & 0.0219 $\pm$ 0.0071 & 0.0263 $\pm$ 0.0109 & 0.5658 $\pm$ 0.1856 & 0.8079 $\pm$ 0.1201 & 0.0067 $\pm$ 0.0030 & 0.0078 $\pm$ 0.0027 & 0.0684 $\pm$ 0.0237 & 0.0954 $\pm$ 0.0149 \\
& T4 & 0.0260 $\pm$ 0.0044 & 0.0329 $\pm$ 0.0073 & 0.7711 $\pm$ 0.2167 & 1.0691 $\pm$ 0.3188 & 0.0092 $\pm$ 0.0029 & 0.0092 $\pm$ 0.0013 & 0.0839 $\pm$ 0.0199 & 0.1239 $\pm$ 0.0189 \\
\midrule
\multirow{4}{*}{Transformer}
& T1 & 0.0462 $\pm$ 0.0130 & 0.0327 $\pm$ 0.0045 & 0.7597 $\pm$ 0.1008 & 0.8344 $\pm$ 0.1184 & 0.0110 $\pm$ 0.0029 & 0.0079 $\pm$ 0.0009 & 0.0934 $\pm$ 0.0103 & 0.0944 $\pm$ 0.0047 \\
& T2 & 0.0133 $\pm$ 0.0033 & 0.0230 $\pm$ 0.0067 & 0.5905 $\pm$ 0.0635 & 0.6388 $\pm$ 0.0292 & \underline{\textbf{0.0039 $\pm$ 0.0008}} & 0.0065 $\pm$ 0.0023 & 0.0656 $\pm$ 0.0076 & 0.0772 $\pm$ 0.0014 \\
& T3 & \underline{\textbf{0.0132 $\pm$ 0.0021}} & 0.0245 $\pm$ 0.0040 & 0.5419 $\pm$ 0.0233 & 0.7210 $\pm$ 0.1041 & 0.0060 $\pm$ 0.0031 & 0.0065 $\pm$ 0.0017 & 0.0678 $\pm$ 0.0068 & 0.0927 $\pm$ 0.0153 \\
& T4 & 0.0139 $\pm$ 0.0042 & 0.0237 $\pm$ 0.0033 & 0.5869 $\pm$ 0.0757 & 0.9856 $\pm$ 0.3325 & 0.0043 $\pm$ 0.0009 & 0.0061 $\pm$ 0.0008 & 0.0690 $\pm$ 0.0110 & 0.1090 $\pm$ 0.0241 \\
\bottomrule
\end{tabular}
\end{adjustbox}
\end{table}
\end{landscape}

\begin{figure}[h]
    \centering
    \includegraphics[width=\linewidth]{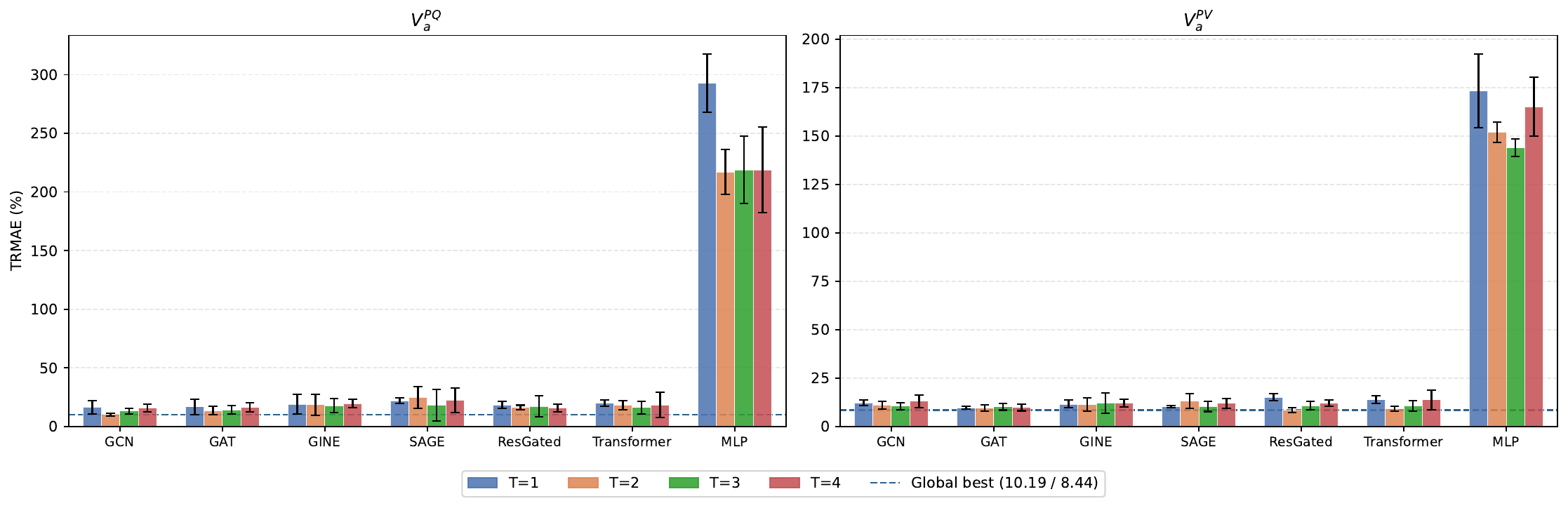}
    \caption{IEEE-118 -- Voltage angle TRMAE (\%) for $V_a^{PQ}$ (left) and
             $V_a^{PV}$ (right) vs.\ independent message-passing depth $T$, including MLP
             baseline.}
    \label{fig:ieee118_angles_mlp}
\end{figure}

\begin{figure}[h]
    \centering
    \includegraphics[width=0.5\linewidth]{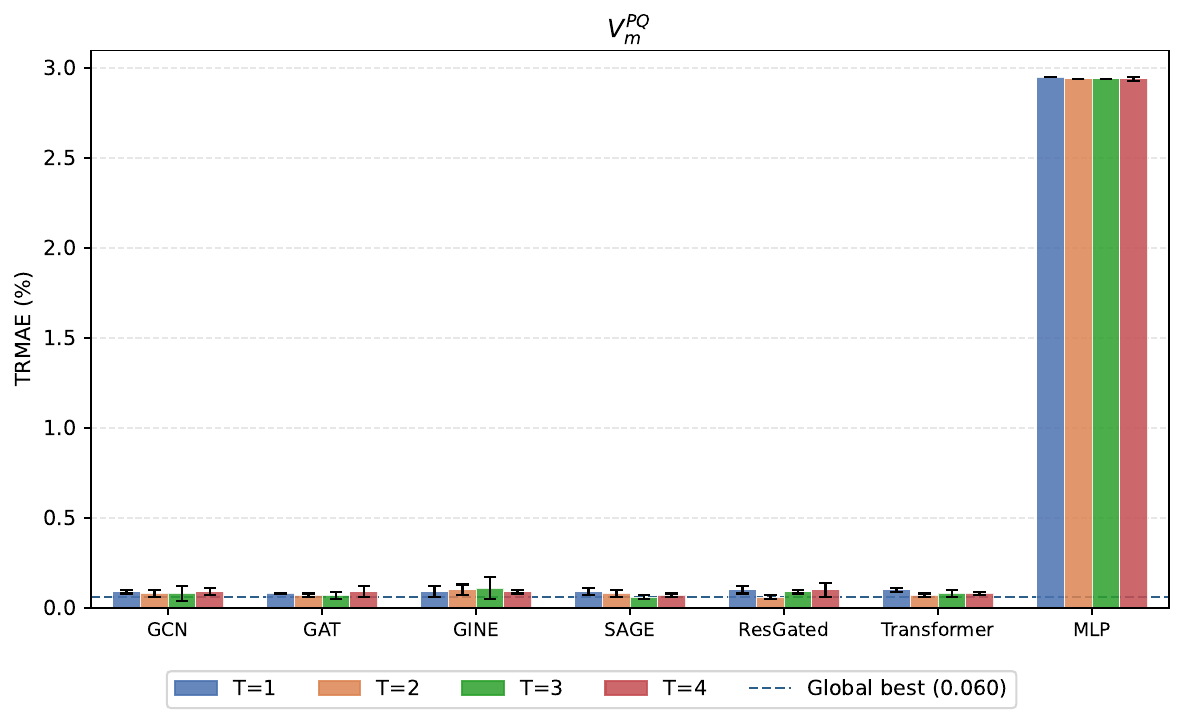}
    \caption{IEEE-118 -- Voltage magnitude TRMAE (\%) ($V_m^{PQ}$) vs.\
             independent message-passing depth $T$, including MLP baseline.}
    \label{fig:ieee118_magnitude_mlp}
\end{figure}

\begin{figure}[H]
    \centering
    \includegraphics[width=\linewidth]{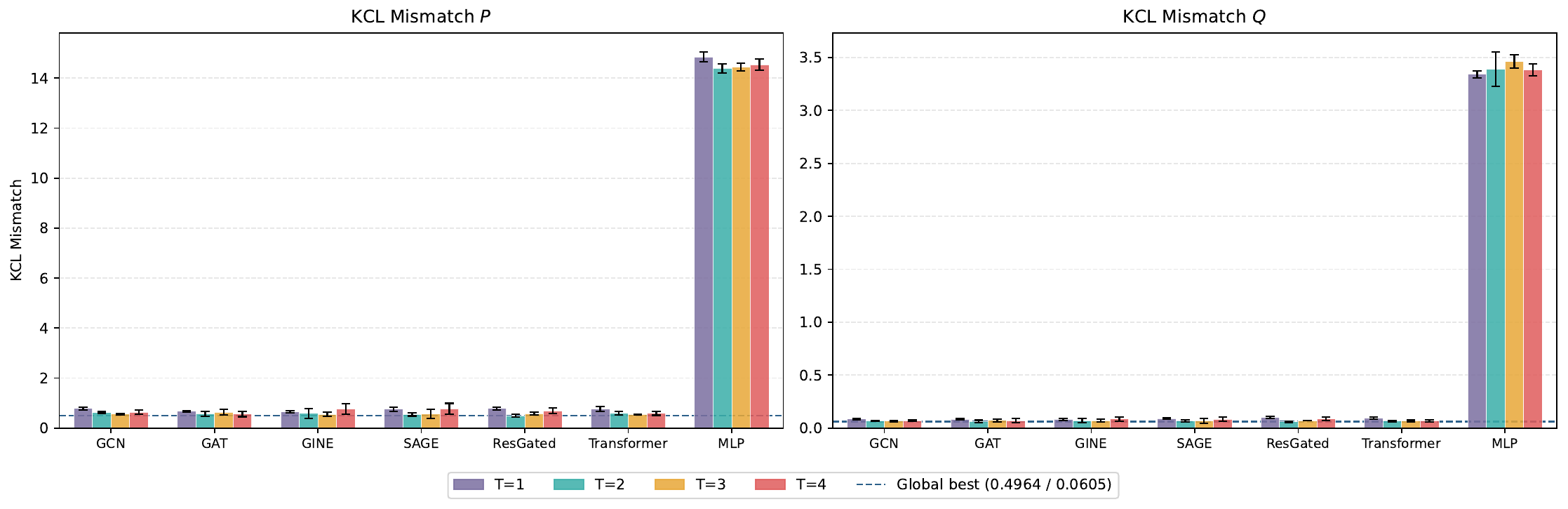}
    \caption{IEEE-118 -- KCL mismatch vs.\ independent message-passing depth $T$
             for active power $P$ (left) and reactive power $Q$ (right),
             including MLP baseline.}
    \label{fig:ieee118_kcl_mlp}
\end{figure}

\begin{figure}[H]
    \centering
    \begin{subfigure}{\linewidth}
        \includegraphics[width=\linewidth]{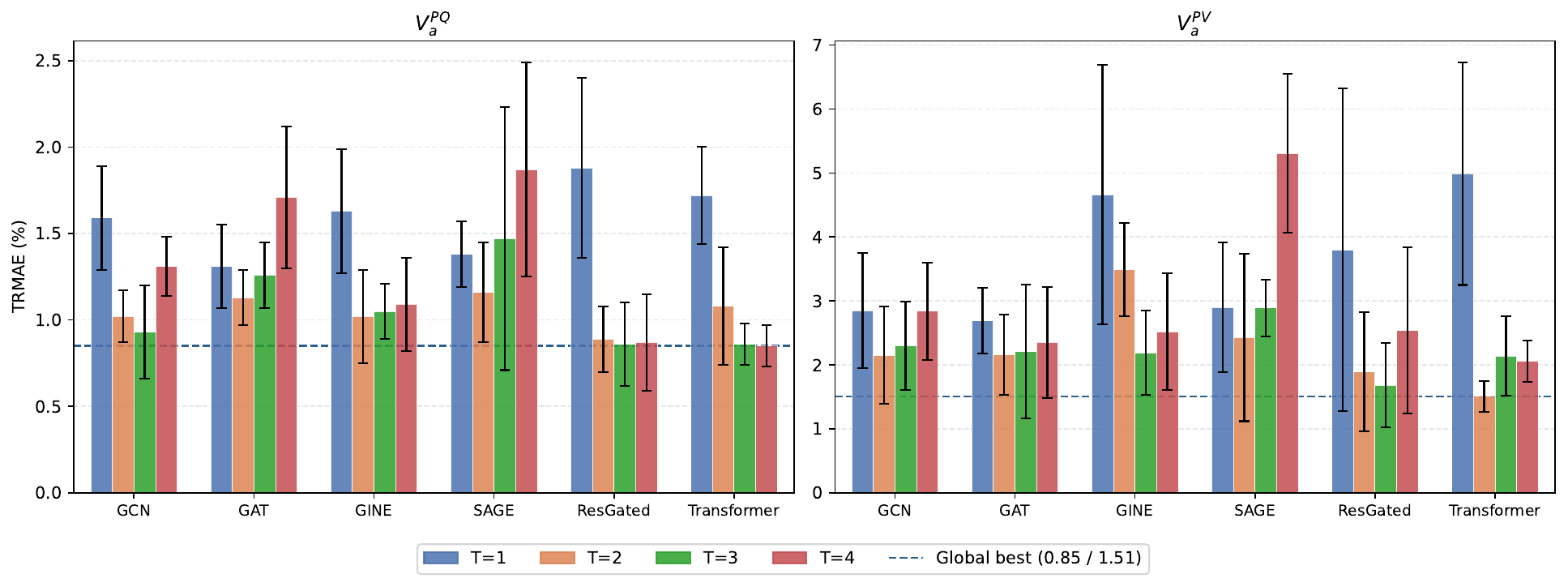}
        \caption{Independent weights}
        \label{fig:ieee14_angles_independent}
    \end{subfigure}
    \vspace{0.5em}
    \begin{subfigure}{\linewidth}
        \includegraphics[width=\linewidth]{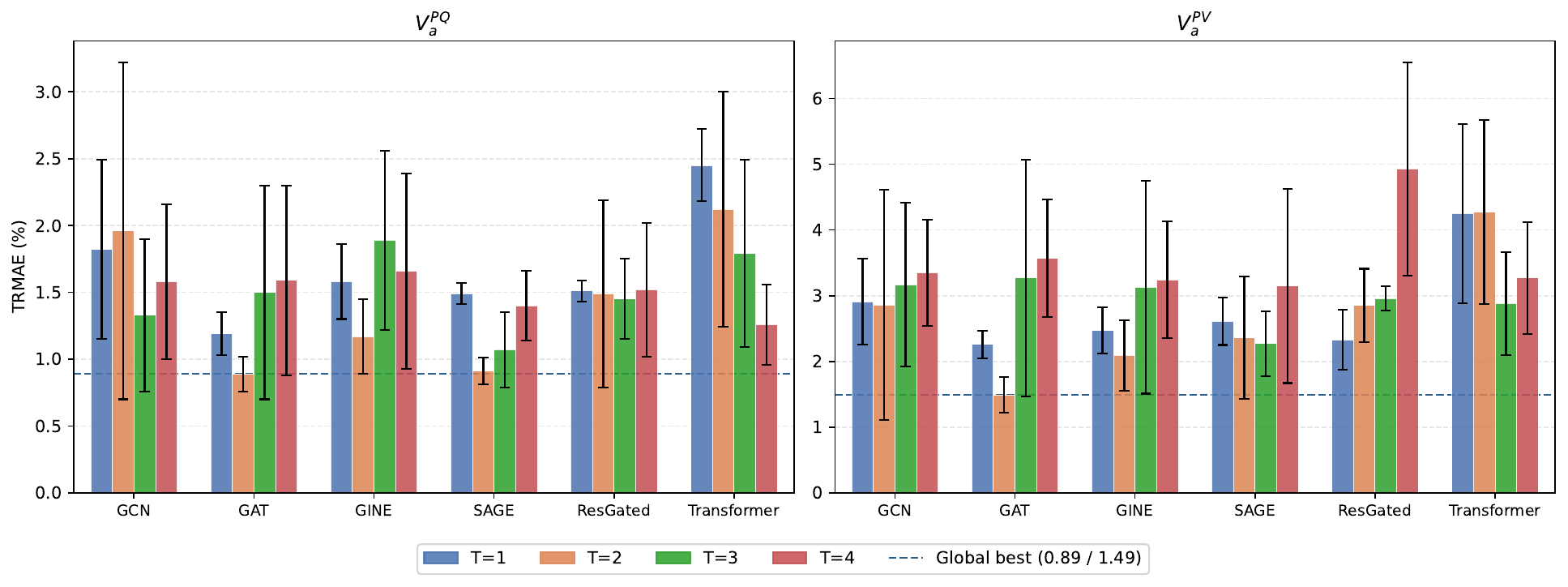}
        \caption{Shared weights}
        \label{fig:ieee14_angles_shared}
    \end{subfigure}
    \caption{IEEE-14 -- Voltage angle TRMAE (\%) for $V_a^{PQ}$ (left) and
             $V_a^{PV}$ (right) vs.\ message-passing depth $T$.}
    \label{fig:ieee14_angles_depth}
\end{figure}

\begin{figure}[H]
    \centering
    \includegraphics[width=\linewidth]{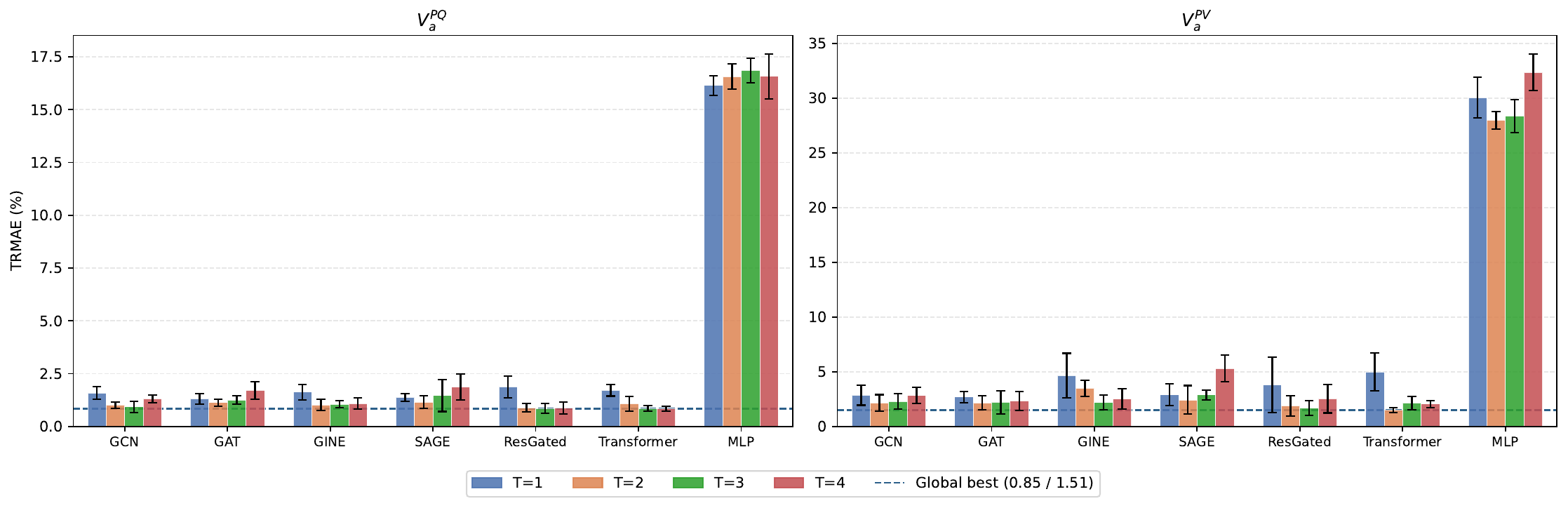}
    \caption{IEEE-14 -- Voltage angle TRMAE (\%) for $V_a^{PQ}$ (left) and
             $V_a^{PV}$ (right) vs.\ independent message-passing depth $T$, including
             MLP baseline.}
    \label{fig:ieee14_angles_mlp}
\end{figure}

\begin{figure}[H]
    \centering
    \begin{subfigure}{0.45\linewidth}
        \includegraphics[width=\linewidth]{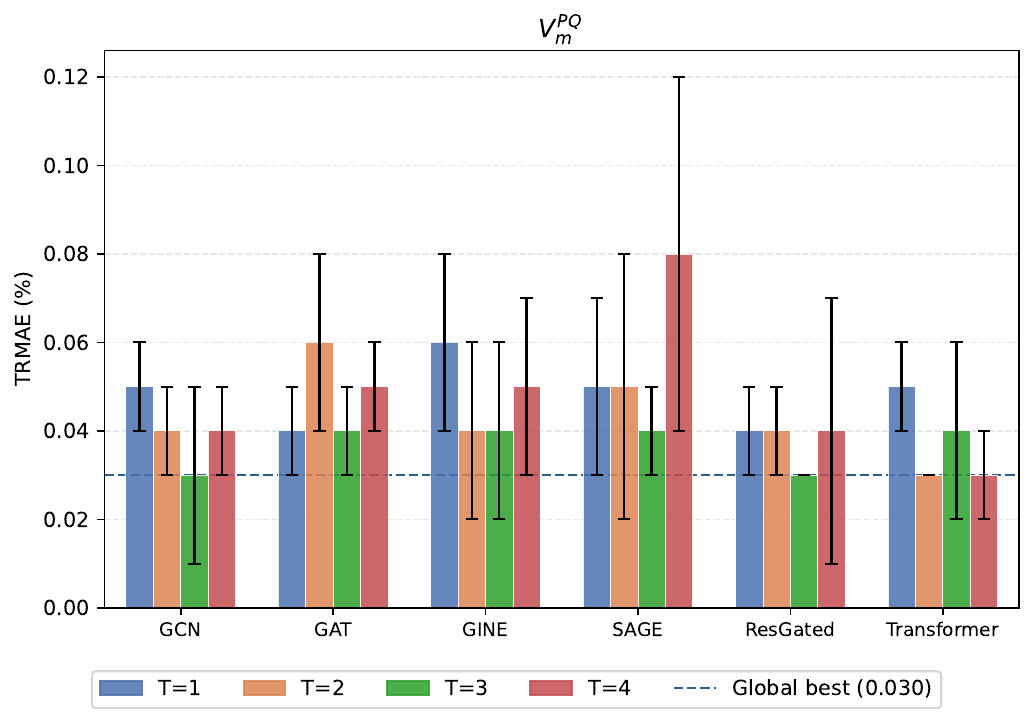}
        \caption{Independent weights}
        \label{fig:ieee14_magnitude_independent}
    \end{subfigure}
    \hfill
    \begin{subfigure}{0.45\linewidth}
        \includegraphics[width=\linewidth]{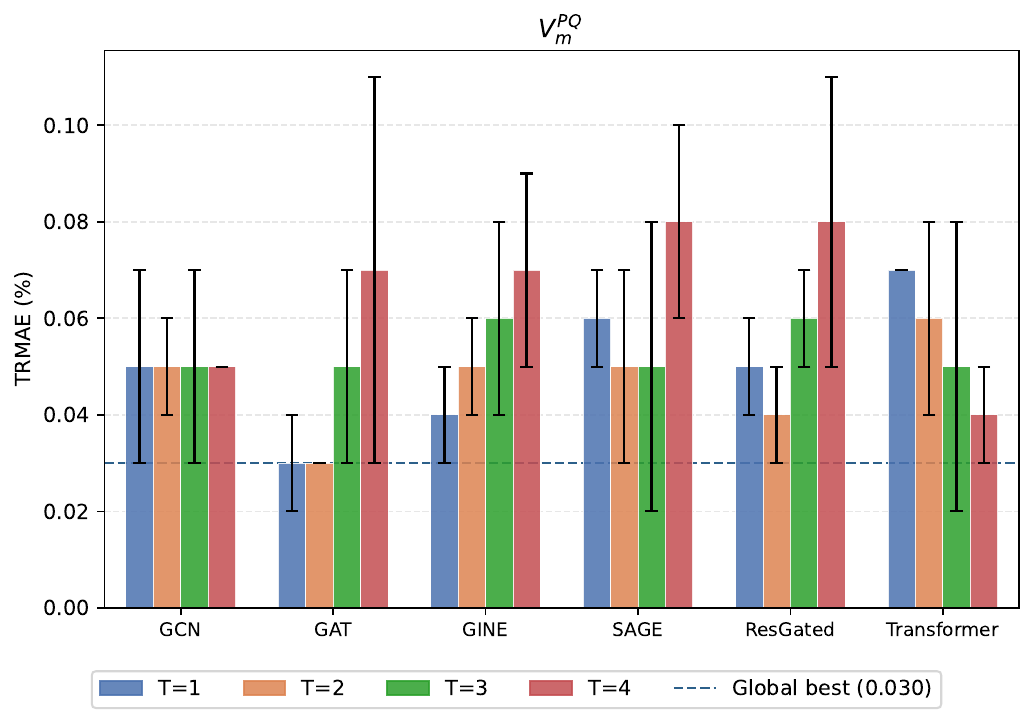}
        \caption{Shared weights}
        \label{fig:ieee14_magnitude_shared}
    \end{subfigure}
    \caption{IEEE-14 -- Voltage magnitude TRMAE (\%) ($V_m^{PQ}$) vs.\
             message-passing depth $T$.}
    \label{fig:ieee14_magnitude_depth}
\end{figure}

\begin{figure}[H]
    \centering
    \includegraphics[width=0.5\linewidth]{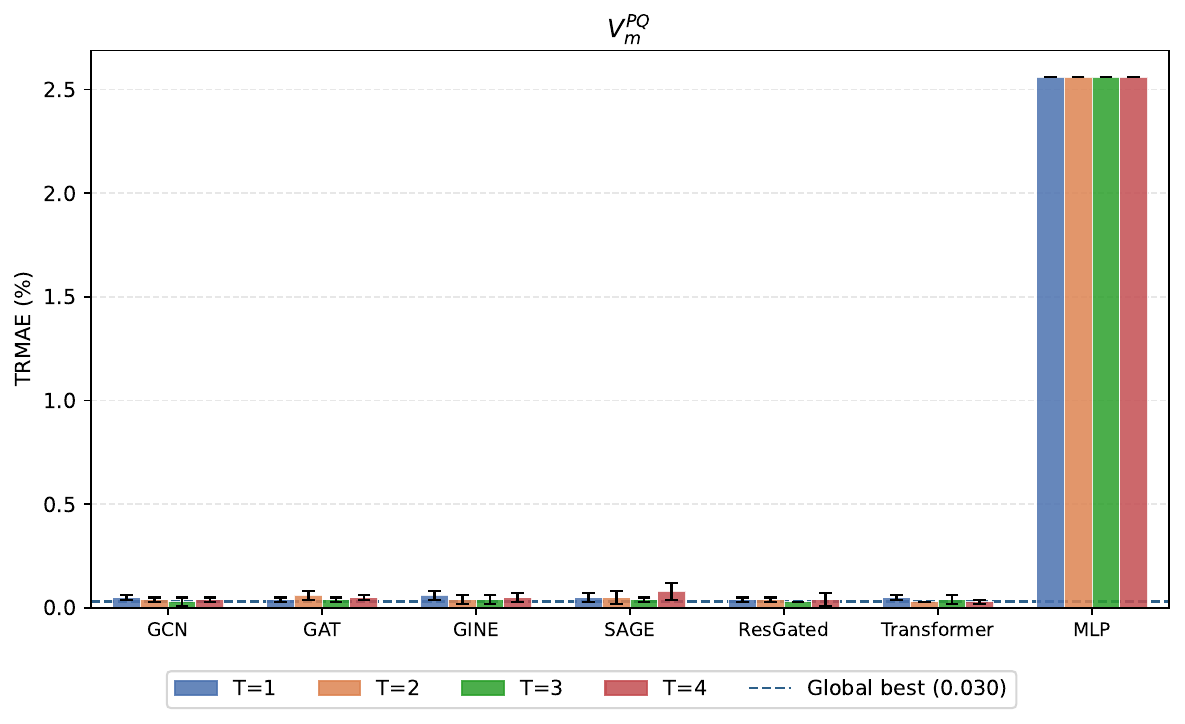}
    \caption{IEEE-14 -- Voltage magnitude TRMAE (\%) ($V_m^{PQ}$) vs.\
             independent message-passing depth $T$, including MLP baseline.}
    \label{fig:ieee14_magnitude_mlp}
\end{figure}

\begin{figure}[H]
    \centering
    \begin{subfigure}{\linewidth}
        \includegraphics[width=\linewidth]{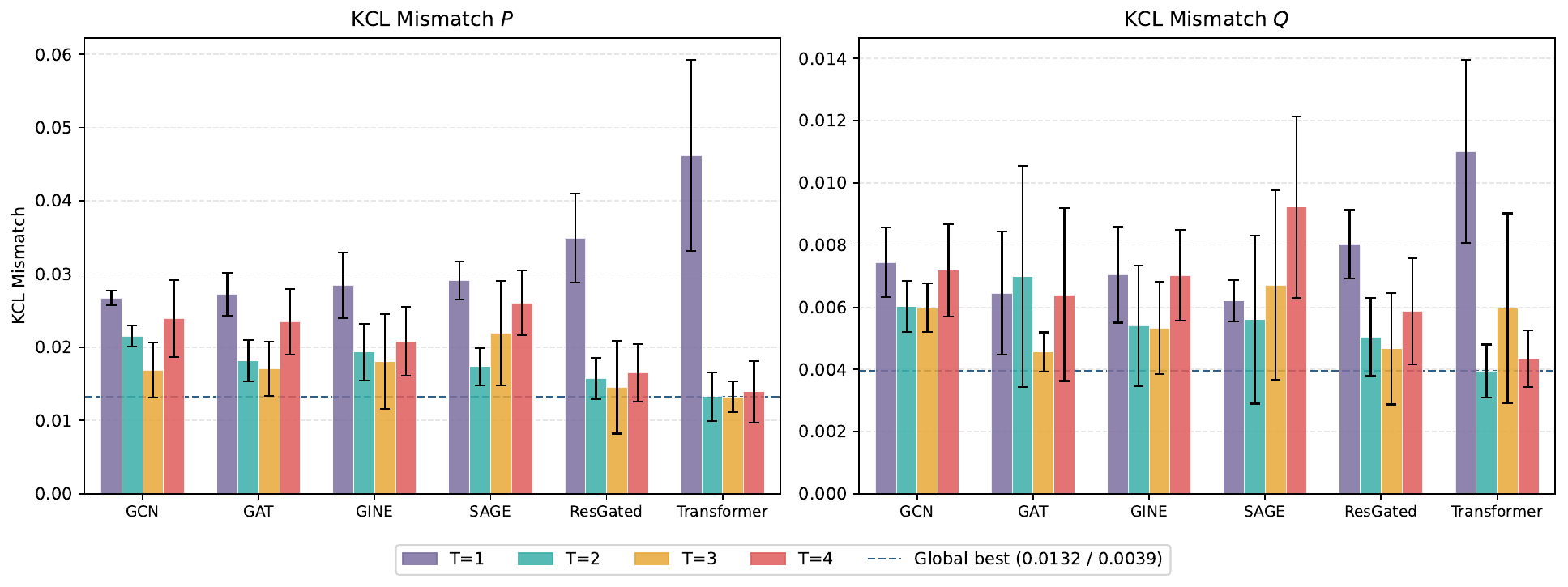}
        \caption{Independent weights}
        \label{fig:ieee14_kcl_independent}
    \end{subfigure}
    \vspace{0.5em}
    \begin{subfigure}{\linewidth}
        \includegraphics[width=\linewidth]{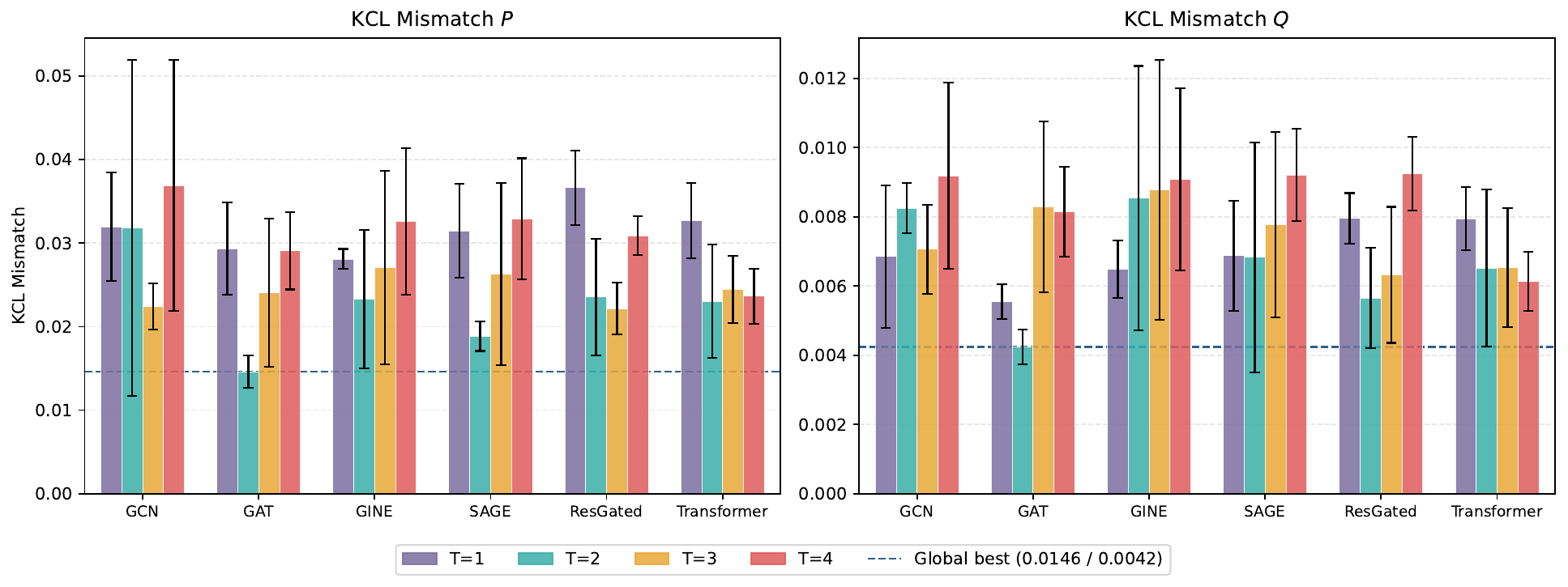}
        \caption{Shared weights}
        \label{fig:ieee14_kcl_shared}
    \end{subfigure}
    \caption{IEEE-14 -- KCL mismatch vs.\ message-passing depth $T$
             for active power $P$ (left) and reactive power $Q$ (right).}
    \label{fig:ieee14_kcl_depth}
\end{figure}

\begin{figure}[H]
    \centering
    \includegraphics[width=\linewidth]{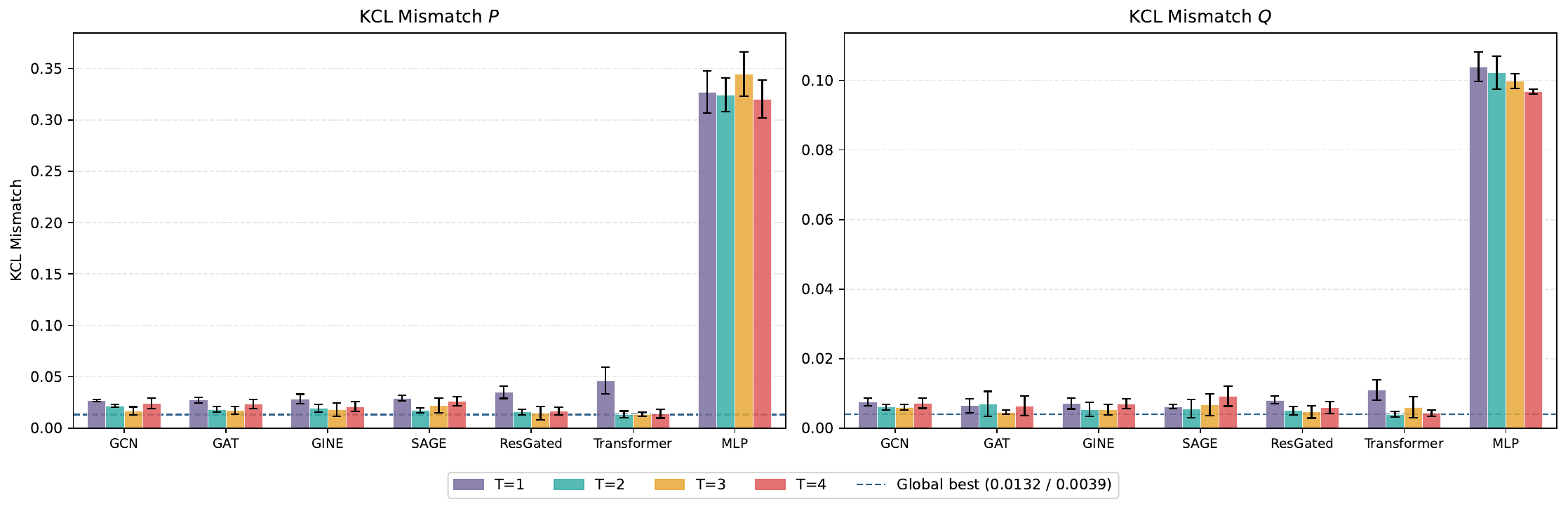}
    \caption{IEEE-14 -- KCL mismatch vs.\ independent message-passing depth $T$
             for active power $P$ (left) and reactive power $Q$ (right),
             including MLP baseline.}
    \label{fig:ieee14_kcl_mlp}
\end{figure}

\subsection{Phase 1: Correlation Between Error (\gls{trmae}) and \gls{kcl} Mismatch}
\label{sec:trmae_kcl_correlation}

This subsection complements the main model-selection analysis by examining the relationship between statistical prediction error and physical consistency. We also assess whether the observed performance differences are stable across the evaluated seeds.

\begin{figure}[H]
    \centering
    \includegraphics[width=0.8\linewidth]{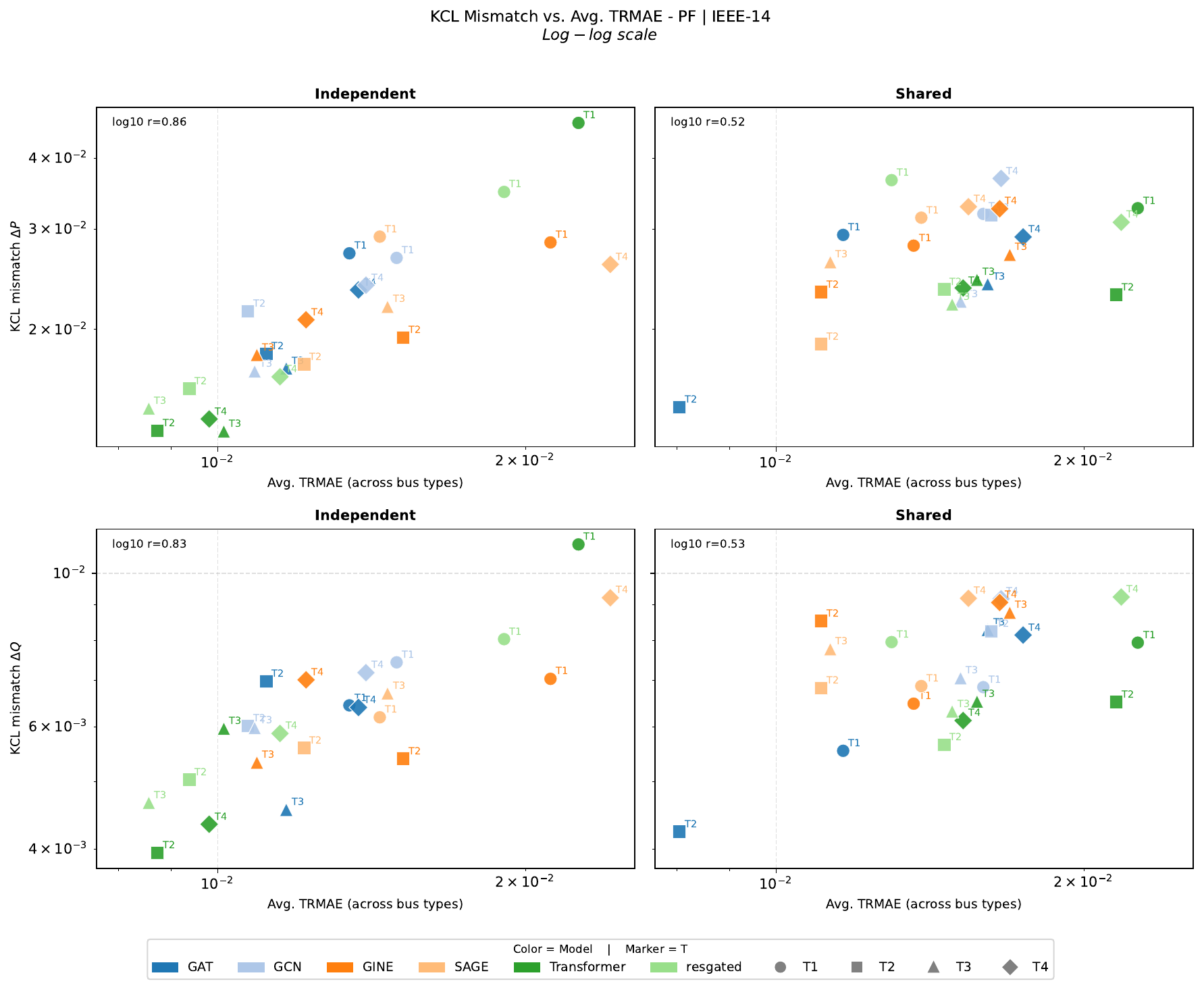}
    \caption{Physical consistency versus prediction accuracy on the IEEE-14 grid. The scatter plots compare active power ($\Delta P$) and reactive power ($\Delta Q$) mismatch against the average \gls{trmae} across bus types. Results show the evaluated \gls{gnn} architectures across transformation depths $T_1$ through $T_4$ under Independent and Shared parameter configurations.}
    \label{fig:trmae_vs_kcl_ieee14}
\end{figure}

\begin{figure}[H]
    \centering
    \includegraphics[width=0.8\linewidth]{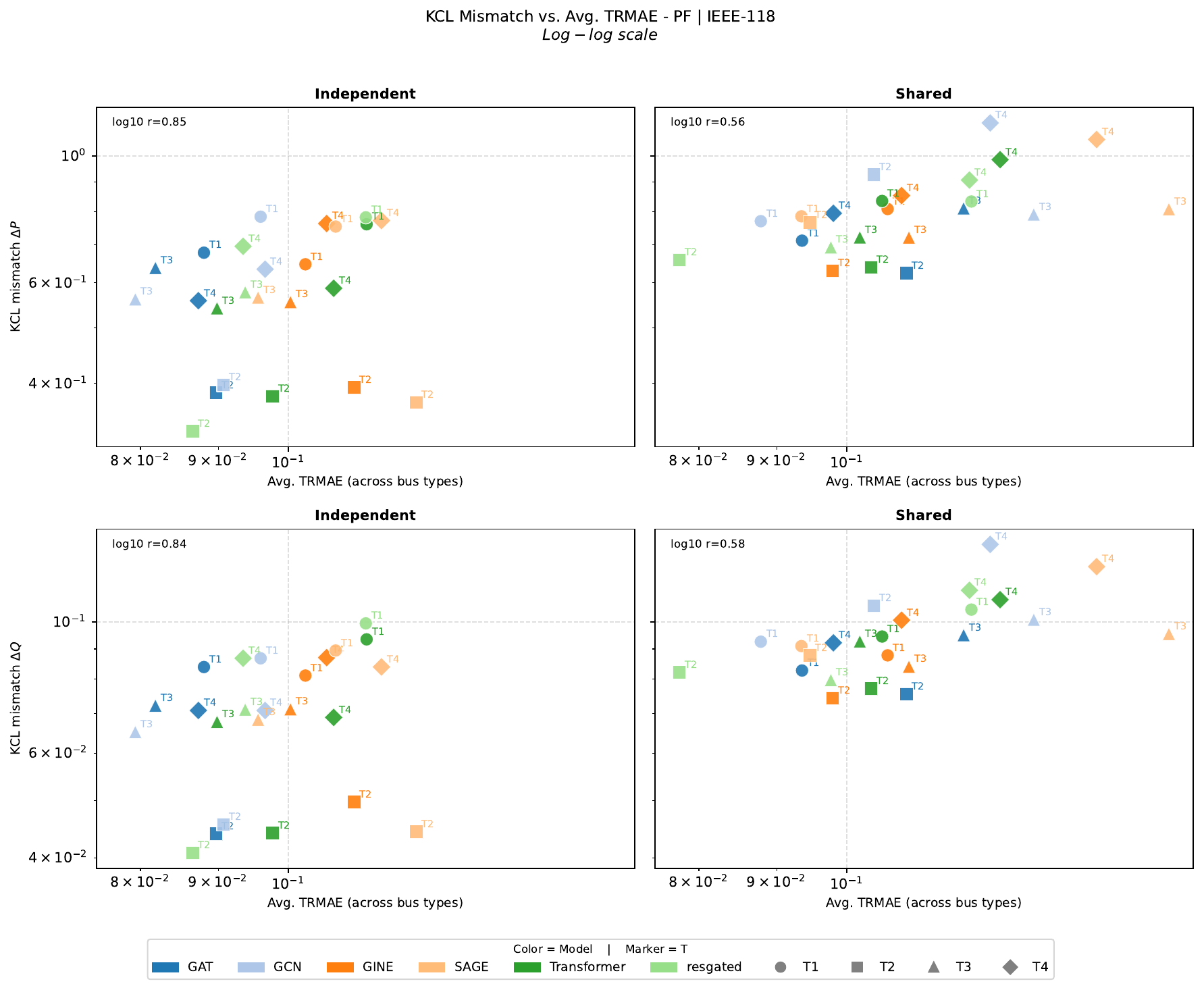}
    \caption{Physical consistency versus prediction accuracy on the IEEE-118 grid. The scatter plots compare active power ($\Delta P$) and reactive power ($\Delta Q$) mismatch against the average \gls{trmae} across bus types. Results show the evaluated \gls{gnn} architectures across transformation depths $T_1$ through $T_4$ under Independent and Shared parameter configurations.}
    \label{fig:trmae_vs_kcl_ieee118}
\end{figure}

As shown in Figures~\ref{fig:trmae_vs_kcl_ieee14} and \ref{fig:trmae_vs_kcl_ieee118}, a lower \gls{trmae} does not perfectly correlate with lower physical mismatches, particularly for the larger IEEE-118 grid. Because several configurations achieve comparable statistical accuracy while exhibiting markedly different levels of \gls{kcl} violations, we designate \gls{kclmae} as the decisive metric for model selection. In terms of raw point estimates, the \gls{resgated} architecture ($T=2$) achieves the lowest mean mismatch for both active and reactive power. However, several top-performing models report overlapping error bars; in this setting, the limited number of independent initializations ($n=4$ seeds) renders standard hypothesis testing severely underpowered and reliant on unverifiable normality assumptions.

To rigorously assess the stability of these performance gaps, we rely on standardized effect sizes (Cohen's $d_z$) paired with bootstrap 95\% confidence intervals (Table~\ref{tab:effect_size_results_ieee118}) to compare \gls{resgated} against its closest competitors (\gls{gat}, \gls{sage}, and Graph Transformer). The analysis reveals medium-to-large effect sizes favoring \gls{resgated} over \gls{gat} ($d_z = -0.94$) and Transformer ($d_z = -0.75$) on active power mismatch ($\Delta P$), with confidence intervals strictly excluding zero. For reactive power ($\Delta Q$) and the comparison against \gls{sage}, the effect sizes are small-to-medium with intervals containing zero, indicating statistically comparable performance. Ultimately, because it yields the best empirical point estimates and no competitor demonstrates a robust advantage over it, we select \gls{resgated} as the representative best-performing architecture for the subsequent diagnostic and ablation phases.

\begin{table}[H]
\centering
\caption{Statistical Robustness of the Top-Performing Architecture. Effect size (Cohen's $d_z$), mean difference, and bootstrap 95\% confidence intervals comparing the baseline \gls{resgated} model ($T=2$, independent layers) against its ablated variants and leading alternative GNNs on the IEEE-118 grid. These metrics assess the statistical significance of the performance gaps for active ($\Delta P$) and reactive ($\Delta Q$) power mismatches, demonstrating whether architectural changes yield robust improvements or degradations.}
\label{tab:effect_size_results_ieee118}
\scriptsize
\setlength{\tabcolsep}{4pt}
\renewcommand{\arraystretch}{1.4}
\begin{adjustbox}{max width=\linewidth}
\begin{tabular}{lrrrcrrrc}
\toprule
& \multicolumn{4}{c}{$\Delta P$} & \multicolumn{4}{c}{$\Delta Q$} \\
\cmidrule(lr){2-5} \cmidrule(lr){6-9}
Comparison & Mean diff (\gls{pu}) & $d_z$ & 95\% CI (\gls{pu}) & Excl.\ 0 & Mean diff (\gls{pu}) & $d_z$ & 95\% CI (\gls{pu}) & Excl.\ 0 \\
\midrule
ResGated Baseline vs AggrMax & 0.1370 & 1.19 & [0.039, 0.224] & Yes & 0.0170 & 1.35 & [0.004, 0.025] & Yes \\
ResGated Baseline vs AggrMean & 0.0115 & 0.50 & [-0.010, 0.029] & No & 0.0013 & 0.21 & [-0.003, 0.007] & No \\
ResGated Baseline vs NoEdgeFeatures & -0.2478 & -1.99 & [-0.327, -0.127] & Yes & -0.0293 & -2.38 & [-0.039, -0.019] & Yes \\
ResGated Baseline vs NoGlobalPooling & -0.4329 & -5.30 & [-0.496, -0.369] & Yes & -0.0371 & -5.67 & [-0.041, -0.031] & Yes \\
ResGated Baseline vs NoResiduals & 0.1816 & 4.13 & [0.145, 0.215] & Yes & 0.0224 & 5.84 & [0.019, 0.026] & Yes \\
ResGated Baseline vs NoRoot & 0.1279 & 2.39 & [0.083, 0.172] & Yes & 0.0169 & 3.39 & [0.013, 0.021] & Yes \\
ResGated vs GAT & -0.0763 & -0.94 & [-0.146, -0.006] & Yes & -0.0028 & -0.23 & [-0.013, 0.007] & No \\
ResGated vs GraphSAGE & -0.0542 & -0.48 & [-0.169, 0.019] & No & -0.0054 & -0.36 & [-0.018, 0.006] & No \\
ResGated vs Graph Transformer ($T=3$) & -0.0456 & -0.75 & [-0.105, -0.009] & Yes & -0.0073 & -0.59 & [-0.020, 0.003] & No \\
\bottomrule
\end{tabular}
\end{adjustbox}
\end{table}

\newpage
\subsection{Phase 2 diagnostic: Distance from the Slack Bus Analysis -- Other Models}
\label{app:distance_slack}

\begin{figure}[H]
    \centering

    \begin{subfigure} {0.49\textwidth}
        \centering
        \includegraphics[width=\linewidth]{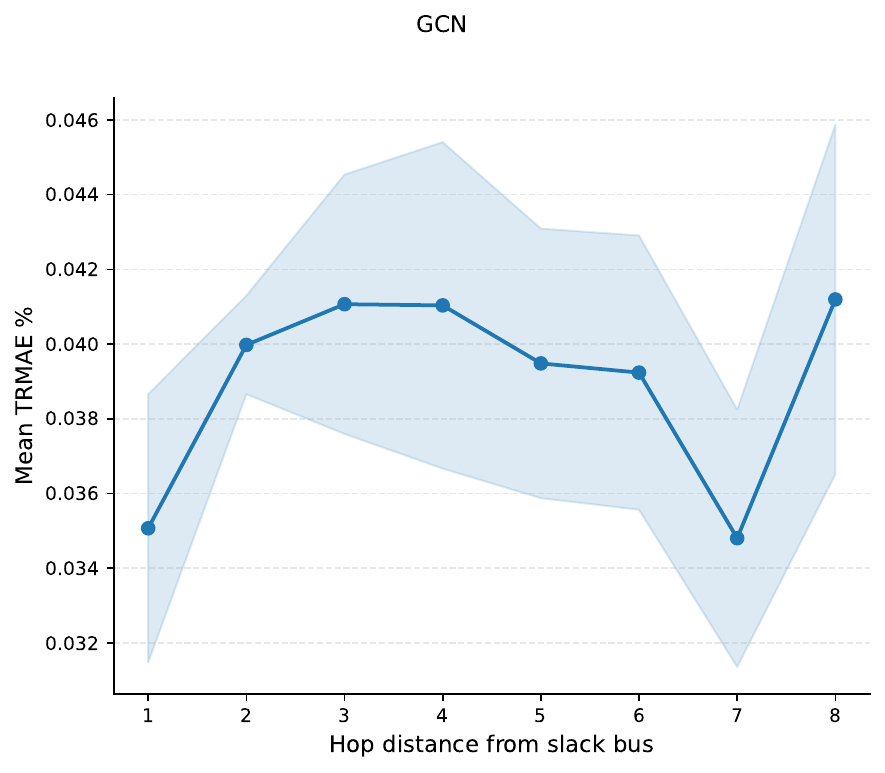}
        \caption{Voltage magnitude}
    \end{subfigure}
    \hfill
    \begin{subfigure}{0.49\textwidth}
        \centering
        \includegraphics[width=\linewidth]{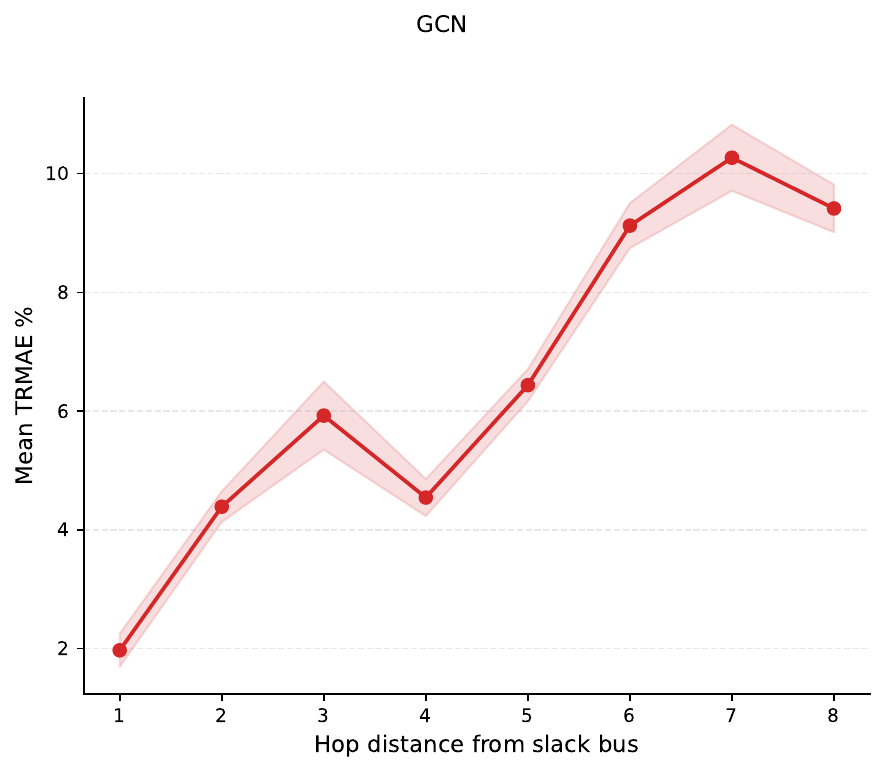}
        \caption{Voltage angle}
    \end{subfigure}

    \caption{Performance of the GCN model relative to hop distance from the slack bus on the IEEE-118 grid (\(T=2\), Independent layers). Results reflect the mean and variance across multiple seeds.}
    \label{fig:gcn_hop_error}
\end{figure}

\begin{figure}[H]
    \centering

    \begin{subfigure}{0.49\textwidth}
        \centering
        \includegraphics[width=\linewidth]{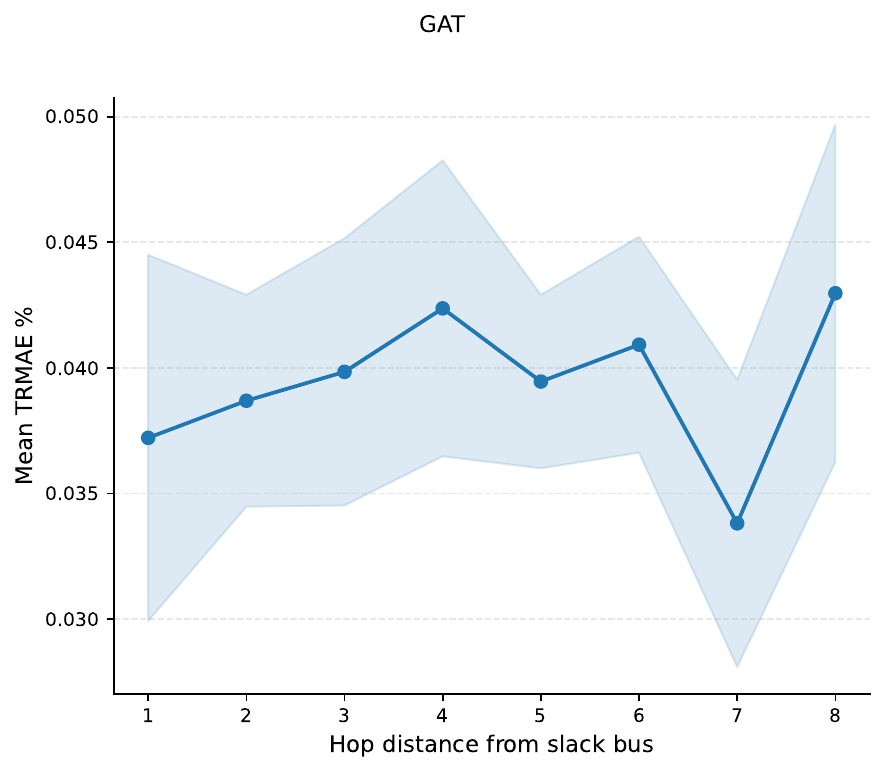}
        \caption{Voltage magnitude}
    \end{subfigure}
    \hfill
    \begin{subfigure}{0.49\textwidth}
        \centering
        \includegraphics[width=\linewidth]{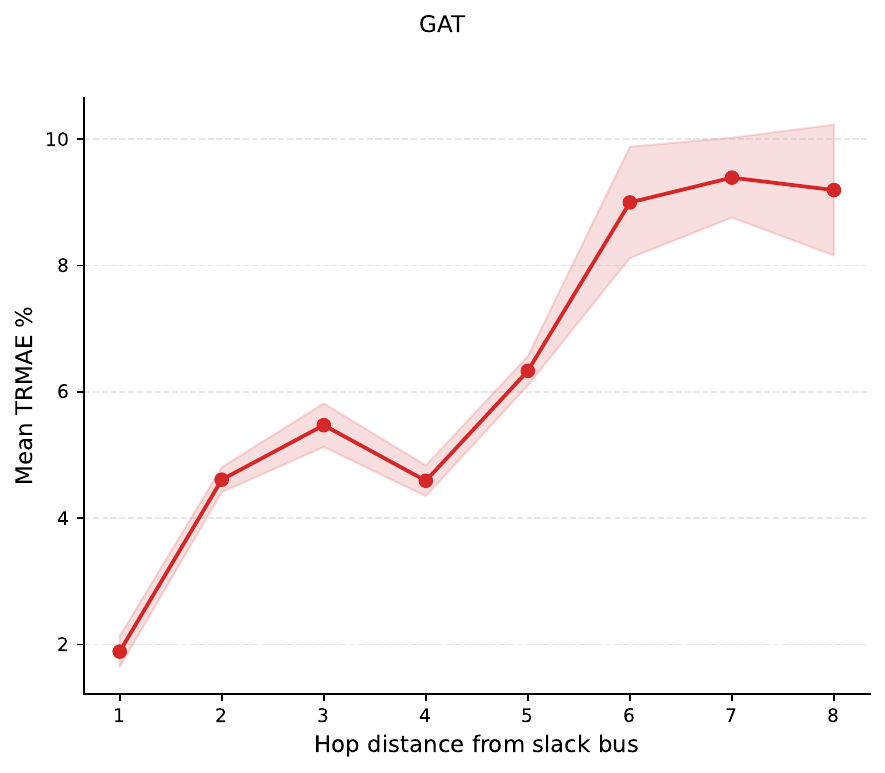}
        \caption{Voltage angle}
    \end{subfigure}

    \caption{Performance of the GAT model relative to hop distance from the slack bus on the IEEE-118 grid (\(T=2\), Independent layers). Results reflect the mean and variance across multiple seeds.}
    \label{fig:gat_hop_error}
\end{figure}

\begin{figure}[H]
    \centering

    \begin{subfigure}{0.49\textwidth}
        \centering
        \includegraphics[width=\linewidth]{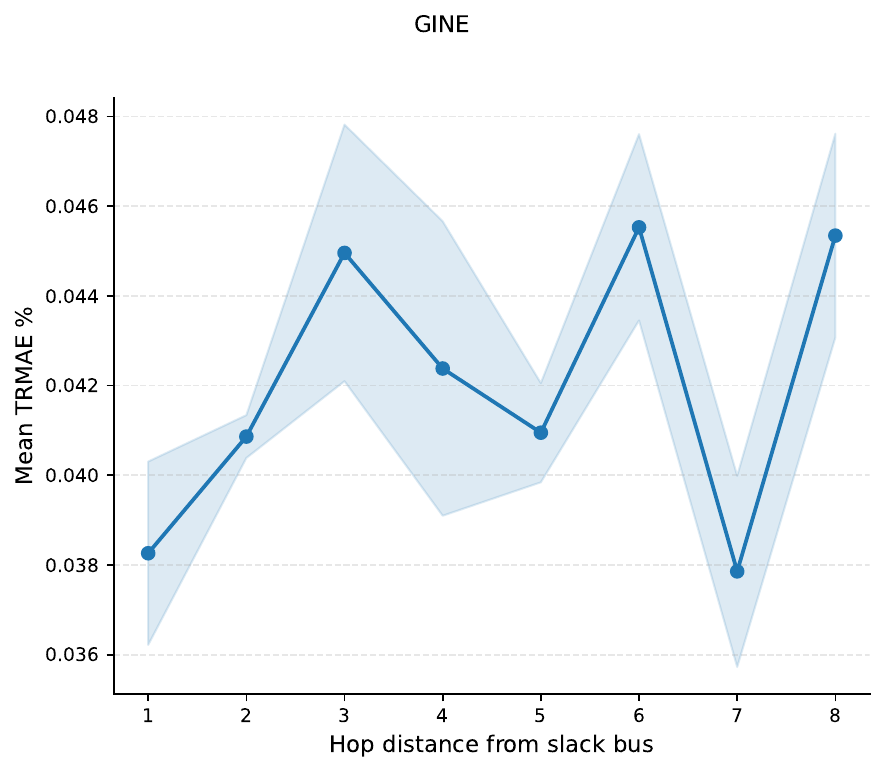}
        \caption{Voltage magnitude}
    \end{subfigure}
    \hfill
    \begin{subfigure}{0.49\textwidth}
        \centering
        \includegraphics[width=\linewidth]{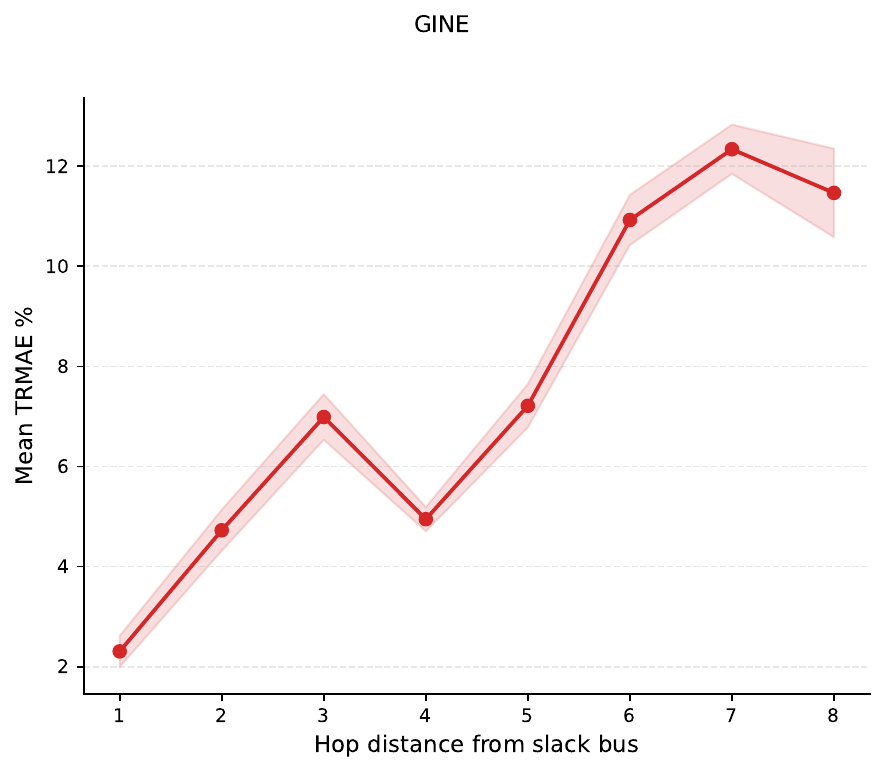}
        \caption{Voltage angle}
    \end{subfigure}

    \caption{Performance of the GINE model relative to hop distance from the slack bus on the IEEE-118 grid (\(T=2\), Independent layers). Results reflect the mean and variance across multiple seeds.}
    \label{fig:gine_hop_error}
\end{figure}

\begin{figure}[H]
    \centering

    \begin{subfigure}{0.49\textwidth}
        \centering
        \includegraphics[width=\linewidth]{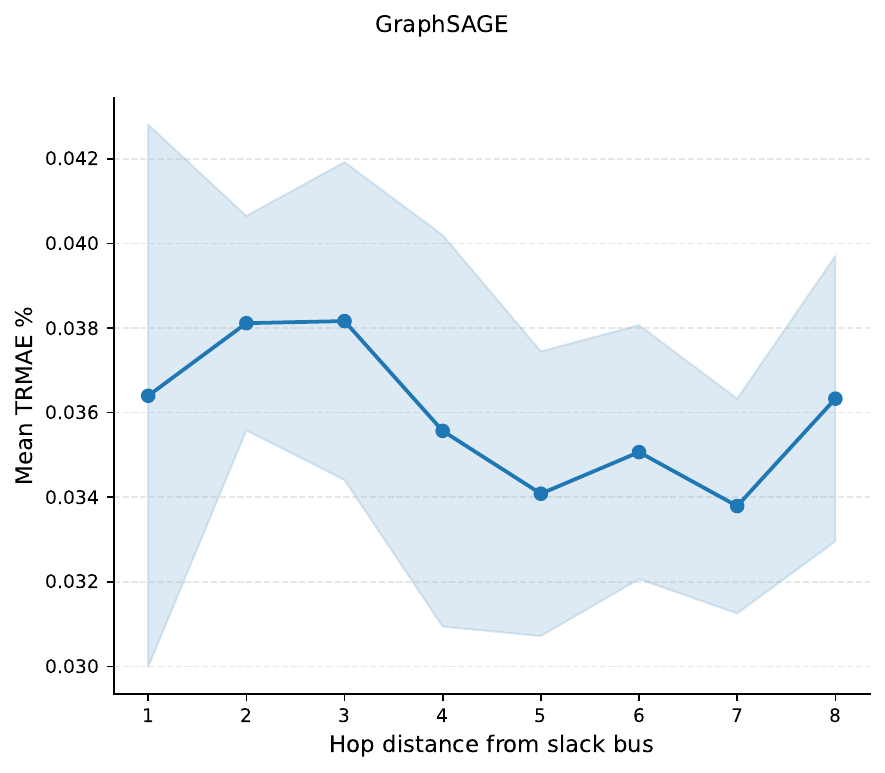}
        \caption{Voltage magnitude}
    \end{subfigure}
    \hfill
    \begin{subfigure}{0.49\textwidth}
        \centering
        \includegraphics[width=\linewidth]{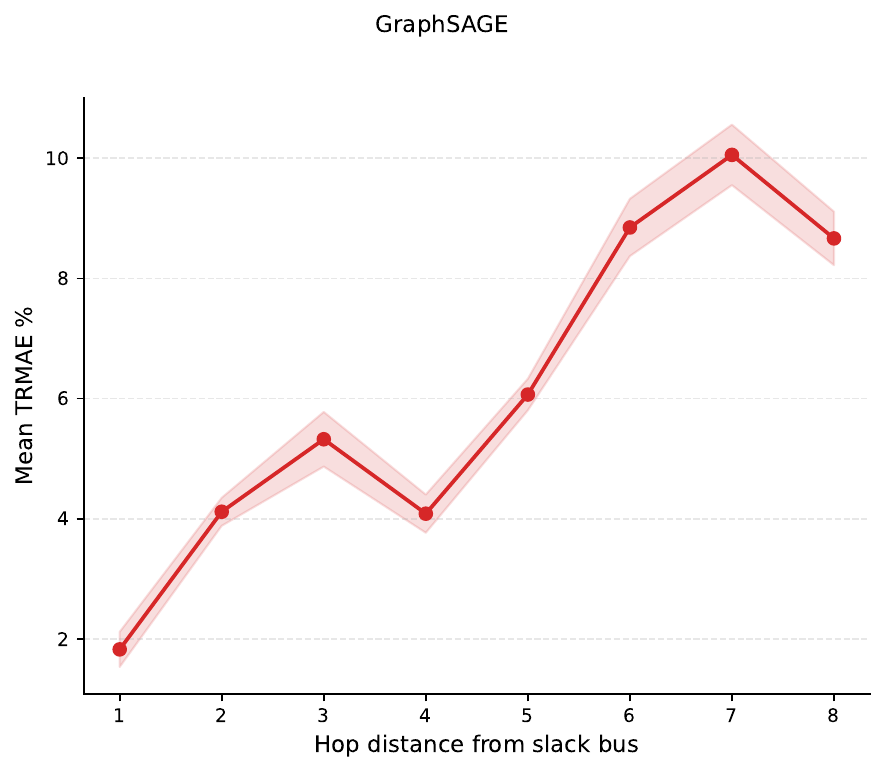}
        \caption{Voltage angle}
    \end{subfigure}

    \caption{Performance of the GraphSAGE model relative to hop distance from the slack bus on the IEEE-118 grid (\(T=2\), Independent layers). Results reflect the mean and variance across multiple seeds.}
    \label{fig:sage_hop_error}
\end{figure}

\begin{figure}[H]
    \centering

    \begin{subfigure}{0.49\textwidth}
        \centering
        \includegraphics[width=\linewidth]{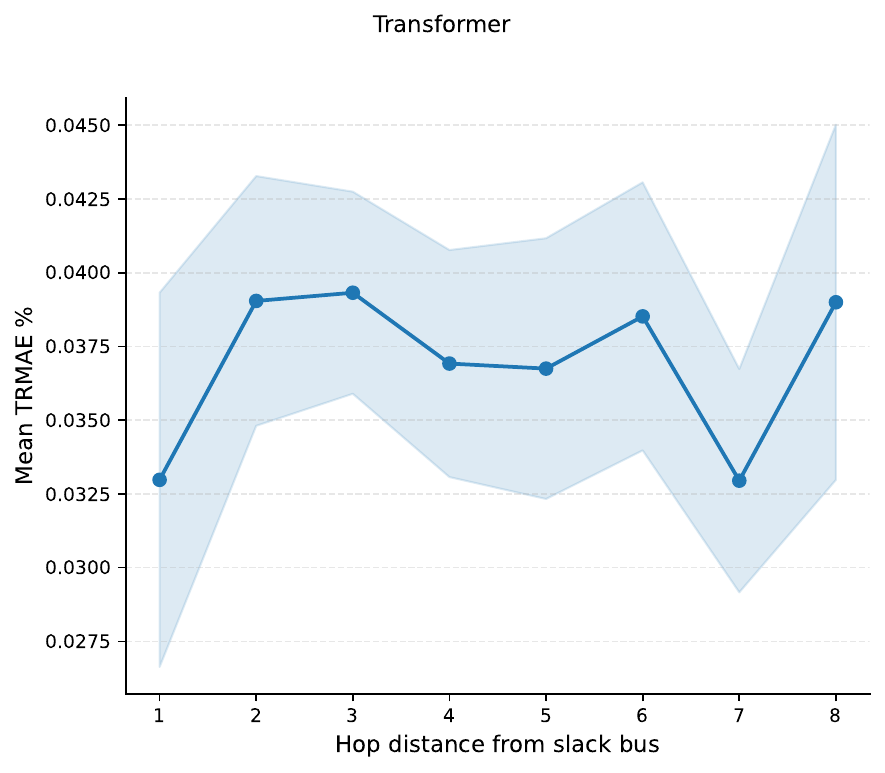}
        \caption{Voltage magnitude}
    \end{subfigure}
    \hfill
    \begin{subfigure}{0.49\textwidth}
        \centering
        \includegraphics[width=\linewidth]{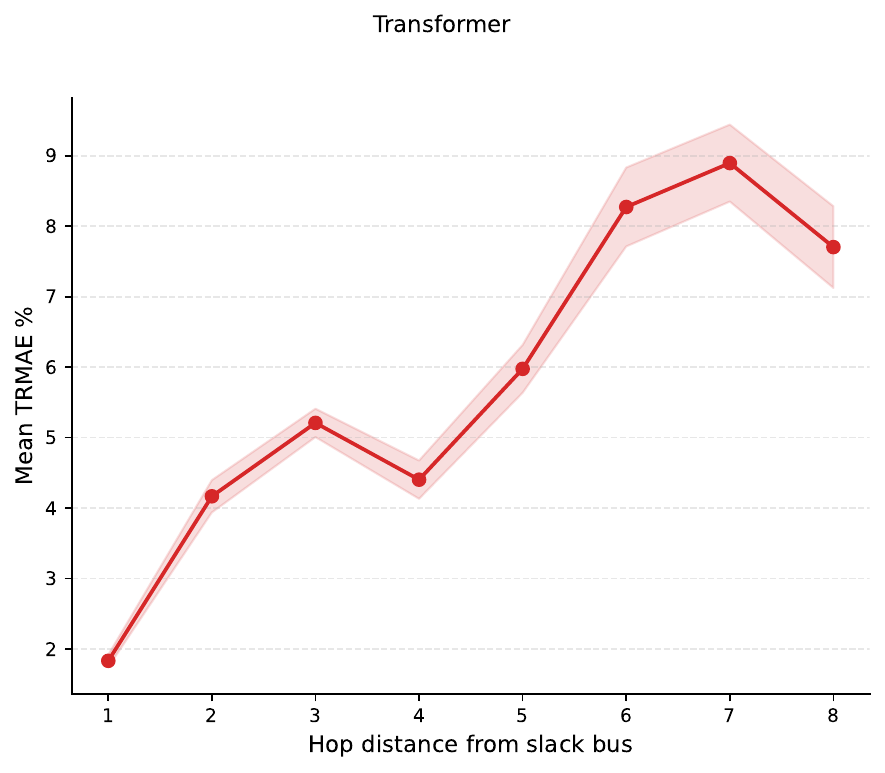}
        \caption{Voltage angle}
    \end{subfigure}

    \caption{Performance of the Transformer model relative to hop distance from the slack bus on the IEEE-118 grid (\(T=2\), Independent layers). Results reflect the mean and variance across multiple seeds.}
    \label{fig:transformer_hop_error}
\end{figure}

\subsection{Phase 2 Diagnostic: Graph Spectral Evaluation of Model Errors}
\label{sec:spectral_analysis_appendix}

This subsection provides a complementary spectral analysis of the Phase~2 diagnostic results presented in Section~\ref{sec:phase2_results}.

\subsubsection{Theoretical Framework: Spectral Error Decomposition}
\label{sec:spectral_analysis}

While standard error metrics evaluate point-wise accuracy, they do not reveal the behavior of the models in learning localized bus interactions or global grid dynamics. To systematically analyze this characteristic in Phase 2, we decompose the nodewise prediction residuals in a graph spectral basis using \gls{gsp} \cite{shuman2013emerging, ramakrishna2021grid}. This spectral view is also motivated by a theoretical characterization of message passing itself. Nt et al. \cite{nt2019revisiting} show that most \gls{gnn} aggregation schemes act as low-pass graph filters, attenuating high-frequency signal content while preserving smooth, low-frequency components. Balcilar et al. \cite{balcilar2021analyzing} refine this picture from a spectral perspective, showing that isotropic, sum-aggregating architectures (e.g., \gls{gcn}, \gls{sage}, \gls{gine} in our taxonomy) are structurally restricted to low-pass responses, whereas anisotropic, gated or attention-based architectures (\gls{resgated}, \gls{gat}, Graph Transformer) can realize richer band-pass or high-pass filter shapes through their learned, feature-dependent edge weights. This distinction offers an a priori expectation for our benchmark: because the isotropic/anisotropic axis of our taxonomy (Section~\ref{sec:mp_tax}) coincides closely with this low-pass/expressive-filter distinction, we expect the spectral error profiles of the two model classes to differ systematically.

\subsubsection*{Defining the Admittance-Weighted Laplacian Proxy}
In conventional graph \gls{ml}, edge weights typically rely on unweighted binary connectivity. For power grids, however, the true coupling between nodes is determined by complex branch admittances $y_{ij} = g_{ij} + \mathrm{j}b_{ij}$. Handling this complexity requires a dedicated framework \cite{ramakrishna2021grid} that lies beyond the scope of this diagnostic analysis. Because standard undirected \gls{gsp} requires a real orthonormal eigenbasis \cite{shuman2013emerging}, we construct a real-valued structural proxy by defining edge weights as the magnitude of the series admittance: $A_{ij} = |y_{ij}| = 1 / \sqrt{R_{ij}^{2}+X_{ij}^{2}}$. While this formulation intentionally discards the complex phase and the distinct roles of conductance and susceptance, it provides a sufficient proxy for this context. By directly encoding the strength of the physical electrical coupling, this admittance-weighted proxy aligns the spectral basis closer to the actual physical reality of the grid than a standard topological adjacency matrix.

Let $\mathbf{D}$ denote the weighted degree matrix. We utilize the symmetric normalized Laplacian $\mathcal{L} = \mathbf{I} - \mathbf{D}^{-1/2} \mathbf{A} \mathbf{D}^{-1/2}$, which admits the eigendecomposition $\mathcal{L} = \mathbf{U} \mathbf{\Lambda} \mathbf{U}^\top$. The eigenvalues $\mathbf{\Lambda} = \mathrm{diag}(\lambda_0, \dots, \lambda_{N-1})$ act as "graph frequencies." Lower frequencies (eigenvalues near 0) correspond to eigenvectors $\mathbf{u}_k$ that vary smoothly across strongly coupled buses, representing globally determined states. Higher frequencies correspond to eigenvectors that oscillate rapidly across edges, representing locally determined interactions.

\subsubsection*{Residual Projection and Cumulative Energy}
For a test set of size $S$, let $\mathbf{e}^{(s)} = \hat{\mathbf{y}}^{(s)} - \mathbf{y}^{(s)}$ denote the nodewise prediction residual. We apply the Graph Fourier Transform (GFT) to project these errors into the spectral domain: $\hat{\mathbf{e}}^{(s)} = \mathbf{U}^\top \mathbf{e}^{(s)}$. 

To quantify how error is distributed across spatial frequencies, we compute the mean spectral residual energy at mode $k$:
\begin{equation}
    \bar{E}[k] = \frac{1}{S}\sum_{s=1}^{S} \left(\mathbf{u}_k^\top\mathbf{e}^{(s)}\right)^2
\end{equation}
Furthermore, we adapt the spectral energy ratio of \cite{tang2022rethinking} to obtain the cumulative spectral-energy proportion $C(k) = \sum_{i=0}^{k}\bar{E}[i] / \sum_{i=0}^{N-1}\bar{E}[i]$. A rapidly increasing $C(k)$ curve indicates that the model's error is more related to smooth, low-frequency graph modes, whereas a slowly increasing curve indicates errors tend to be distributed over highly localized, high-frequency mismatches. 

\subsubsection*{Assumptions and Limitations}
This analysis serves as an empirical diagnostic of residual spatial structure rather than strict formal \gls{gsp} theory. By using a real-valued, normalized Laplacian proxy, we discard the complex phase of electrical coupling and mathematically distort the strict nodal \gls{kcl} \cite{dorfler2018electrical}. Consequently, the resulting eigenvectors are not exact physical wave propagation modes. Instead, they serve as a mathematically convenient basis to map spatial variations—from smooth grid-wide averages to sharp, neighboring-bus mismatches—assuming a fixed network topology over the evaluated steady-state benchmark.

\subsubsection{Diagnostic Findings: Network Anchoring and Topological Bottlenecks}

To better understand how prediction errors relate to the grid's topology, we analyze their spectral distribution. Voltage angles ($V_a$) are typically driven by global network states relative to the slack bus, while voltage magnitudes ($V_m$) are governed by localized reactive power balances. Recent work highlights the necessity of globally aware tools for accurate grid predictions \cite{arowolo2025towards}. To investigate this, we ablate the global pooling mechanism from our baseline architecture (\gls{resgated}, $T=2$ Independent layers) to isolate its impact on both global and local variables.

\begin{figure}[H]
    \centering
    \begin{subfigure}{0.49\linewidth}
        \centering
        \includegraphics[width=\linewidth]{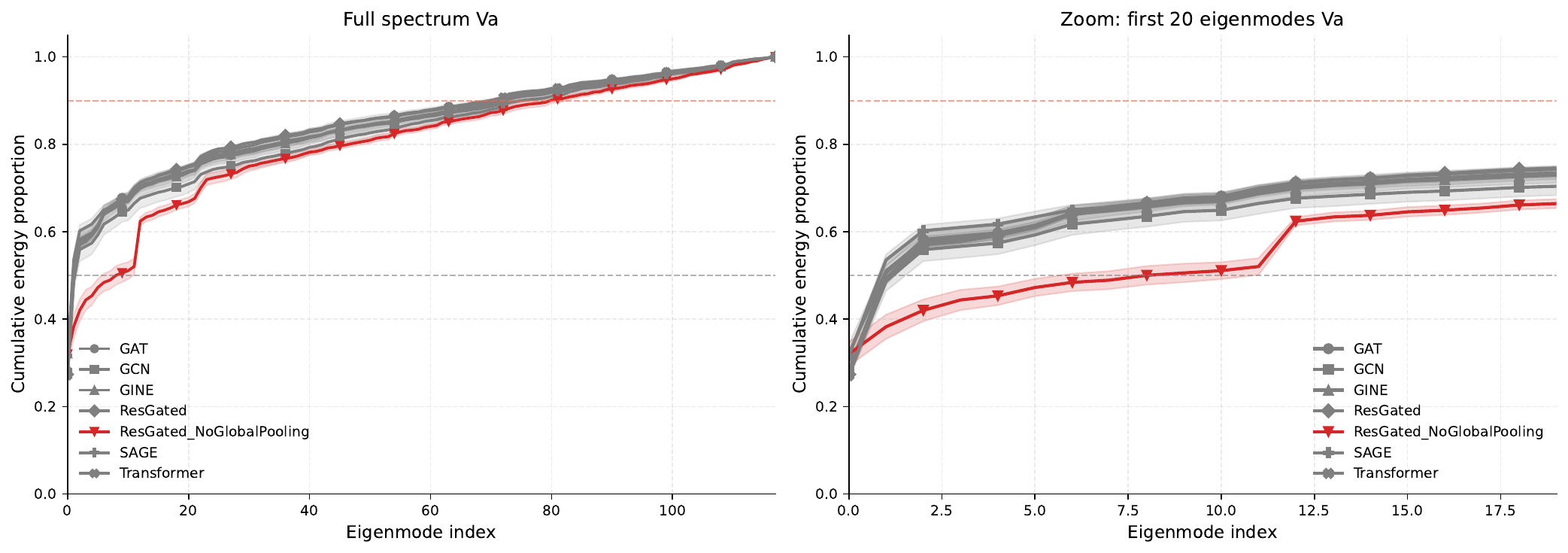}
        \caption{Voltage Angle ($V_a$)}
        \label{fig:cum_spectral_error_va}
    \end{subfigure}
    \hfill
    \begin{subfigure}{0.49\linewidth}
        \centering
        \includegraphics[width=\linewidth]{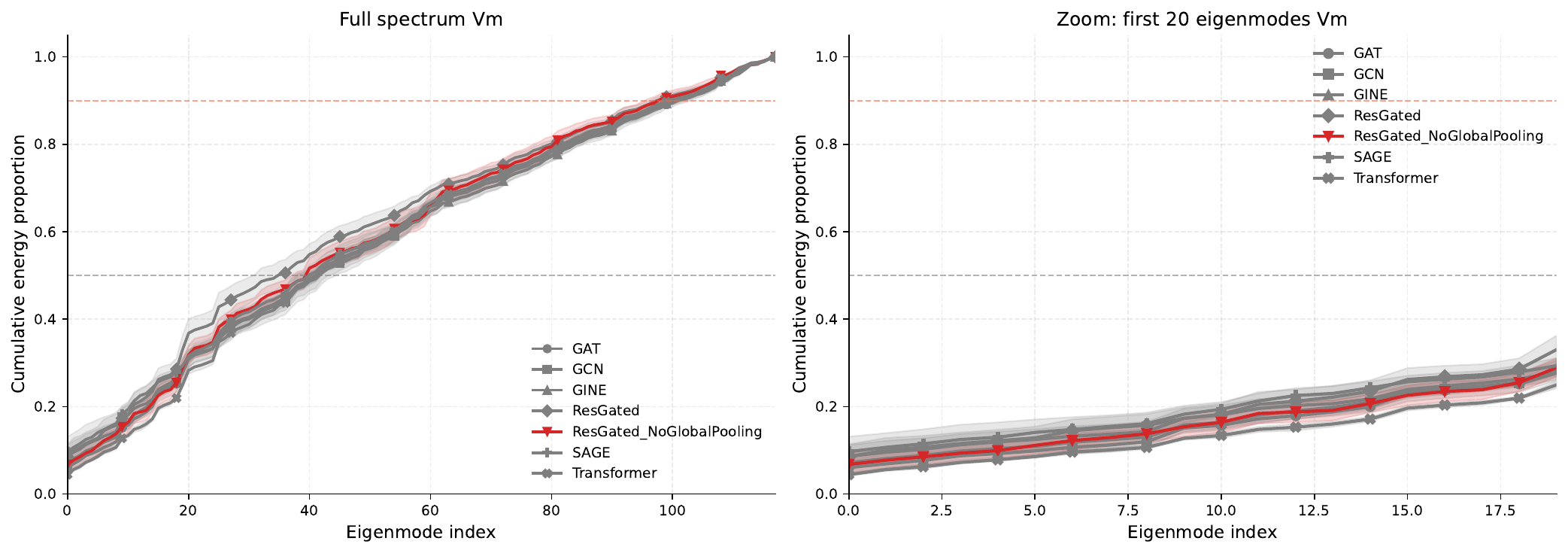}
        \caption{Voltage Magnitude ($V_m$)}
        \label{fig:cum_spectral_error_vm}
    \end{subfigure}
    \caption{Cumulative spectral energy proportion for $V_m$ and $V_a$ averaged across buses for the models on the IEEE-118 grid (\(T=2\), Independent layers).}
    \label{fig:cumulative_spectral_error}
\end{figure}

Figure~\ref{fig:cumulative_spectral_error} illustrates the cumulative spectral error energy for $V_a$ and $V_m$. For the baseline models, $V_a$ errors accumulate heavily in the lowest eigenmodes. This indicates that angle errors manifest primarily as smooth, global network shifts, meaning architectures are differentiated largely by their low-frequency behavior. In contrast, baseline $V_m$ errors are distributed more gradually across higher-frequency modes, reflecting the localized nature of voltage regulation.

At first glance, this concentration of $V_a$ error in the lowest eigenmodes may appear to contradict the low-pass characterization of message-passing GNNs discussed in Section~\ref{sec:spectral_analysis} \citep{nt2019revisiting}: a low-pass filter should, in principle, be best suited to reconstructing exactly the smooth, low-frequency signal content that dominates $V_a$. This apparent tension resolves once we distinguish between a model's ability to learn a signal's \emph{relative shape} versus its \emph{absolute anchor}. A low-pass filter can faithfully preserve the smooth, relative differences between connected buses while still predicting the entire network with a constant, global offset. In graph spectral terms, this network-wide shift corresponds to the zero-frequency (DC) component---and physically, it represents a failure to anchor the network's angles strictly to the global slack-bus reference. Since nearly all of the true $V_a$ signal energy is concentrated in this exact component, even a small uniform anchoring error is severely amplified in absolute residual energy. Under this reading, our results do not contradict the low-pass characterization of \glspl{gnn} so much as expose its limitation: reproducing the low-frequency \emph{shape} of a physical signal is not sufficient to guarantee accurate reconstruction of its absolute baseline level, which for voltage angles corresponds directly to anchoring against the slack bus. This also motivates the isotropic/anisotropic distinction introduced in Section~\ref{sec:mp_tax}: because isotropic aggregators (\gls{gcn}, \gls{sage}, \gls{gine}) are structurally confined to low-pass responses \citep{balcilar2021analyzing}, they have comparatively less flexibility to correct this DC-level miscalibration than the anisotropic, gated ResGated architecture, consistent with the latter's superior performance on $V_a$ observed in Phase~1 (Section~\ref{sec:phase1_results}) and its more moderate error growth under the global-pooling ablation shown below.

Crucially, ablating the global pooling mechanism (e.g., in the \gls{resgated} model) significantly alters the error distribution for $V_a$. The ablated model exhibits a broader, step-like error distribution, consistent with two compounding effects: an initial step plausibly reflecting the loss of slack bus information due to the model's limited local receptive field, and a further step consistent with information bottlenecks caused by topological clusters within the network \cite{sanchez2014hierarchical}. We note that this two-step interpretation is suggestive rather than formally identified, as we do not isolate each mechanism independently. For $V_m$, by contrast, the shape of the cumulative error distribution remains largely unchanged by the ablation, suggesting that the structural benefits of global pooling are primarily associated with angle prediction.

\begin{figure}[H]
    \centering
    \includegraphics[width=0.4\linewidth]{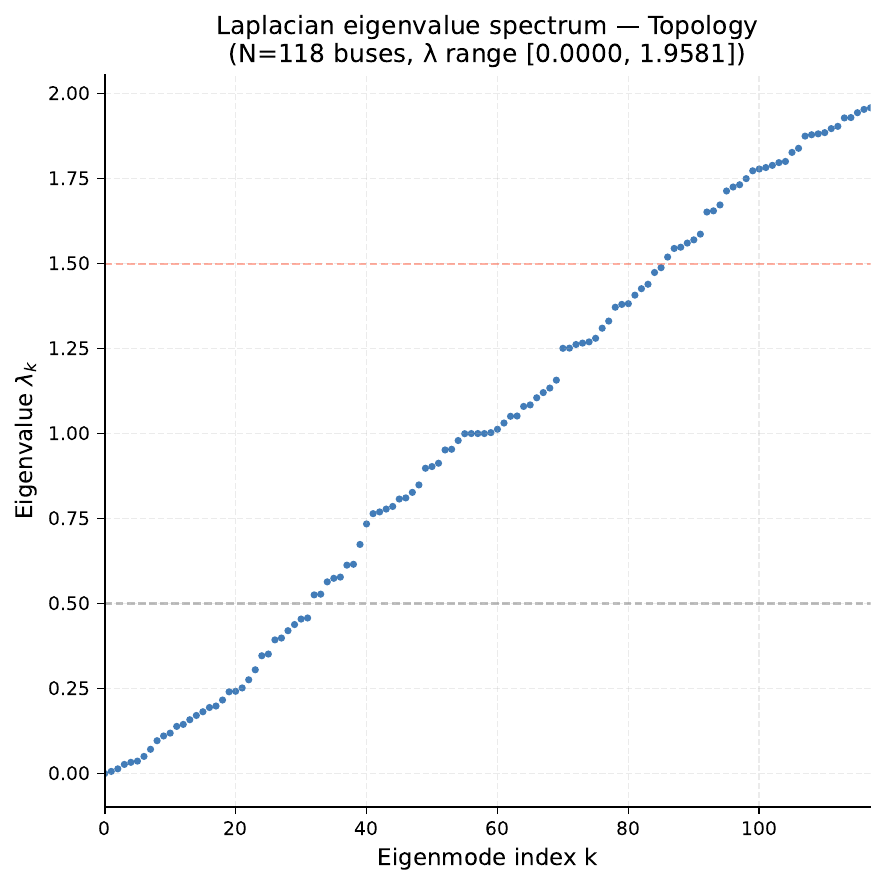}
    \caption{Eigenvalue spectrum ($\lambda_k$) of the admittance-weighted normalized graph Laplacian for the IEEE-118 network.}
    \label{fig:eig_mag}
\end{figure}

While the cumulative distribution highlights structural behavior, the absolute magnitude of these errors is equally important. We adopt fixed boundaries at $\lambda = 0.5$ and $\lambda = 1.5$, partitioning the admissible range $[0,2]$ into three spectral bands: low-frequency ($\lambda < 0.5$), mid-frequency ($0.5 \leq \lambda < 1.5$), and high-frequency ($\lambda \geq 1.5$). While these boundaries are not derived from a data-driven threshold, we verified that our qualitative conclusions are insensitive to small perturbations of the selected cutoffs.

\begin{figure}[H]
    \centering
    \begin{subfigure}[t]{0.49\linewidth}
        \centering
        \includegraphics[width=\linewidth]{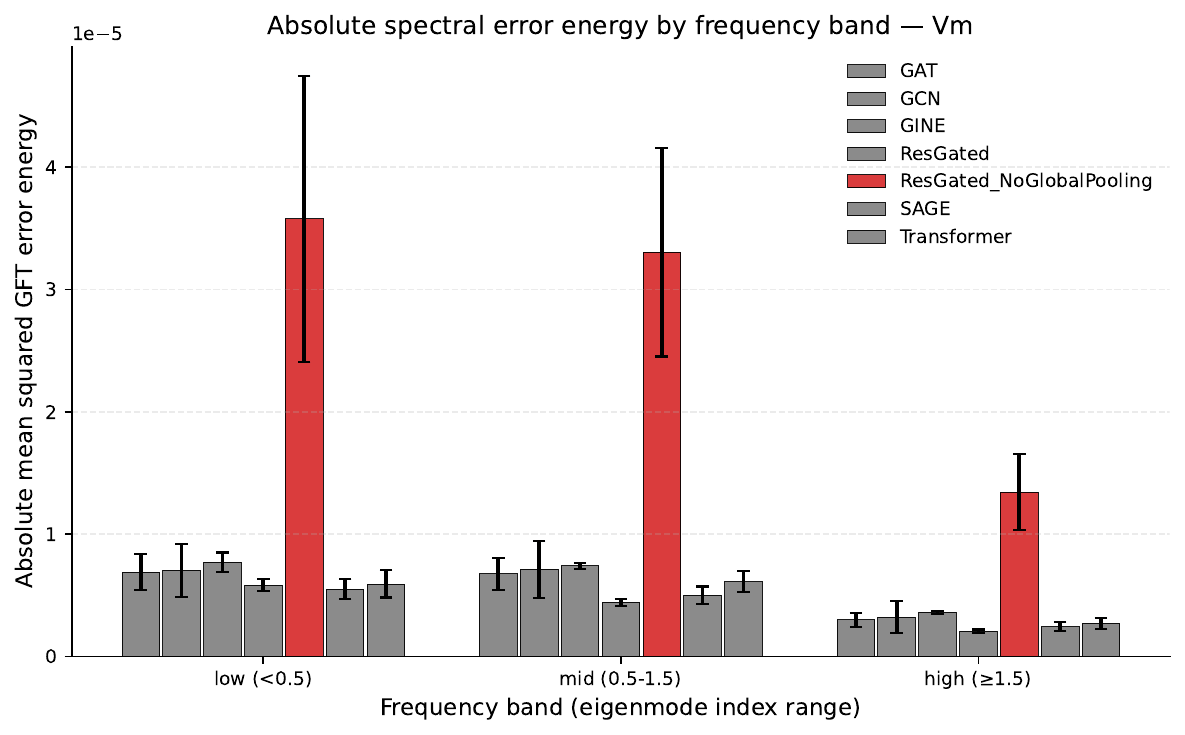}
        \caption{Voltage Magnitude ($V_m$)}
        \label{fig:abs_spectral_error_vm}
    \end{subfigure}
    \hfill
    \begin{subfigure}[t]{0.49\linewidth}
        \centering
        \includegraphics[width=\linewidth]{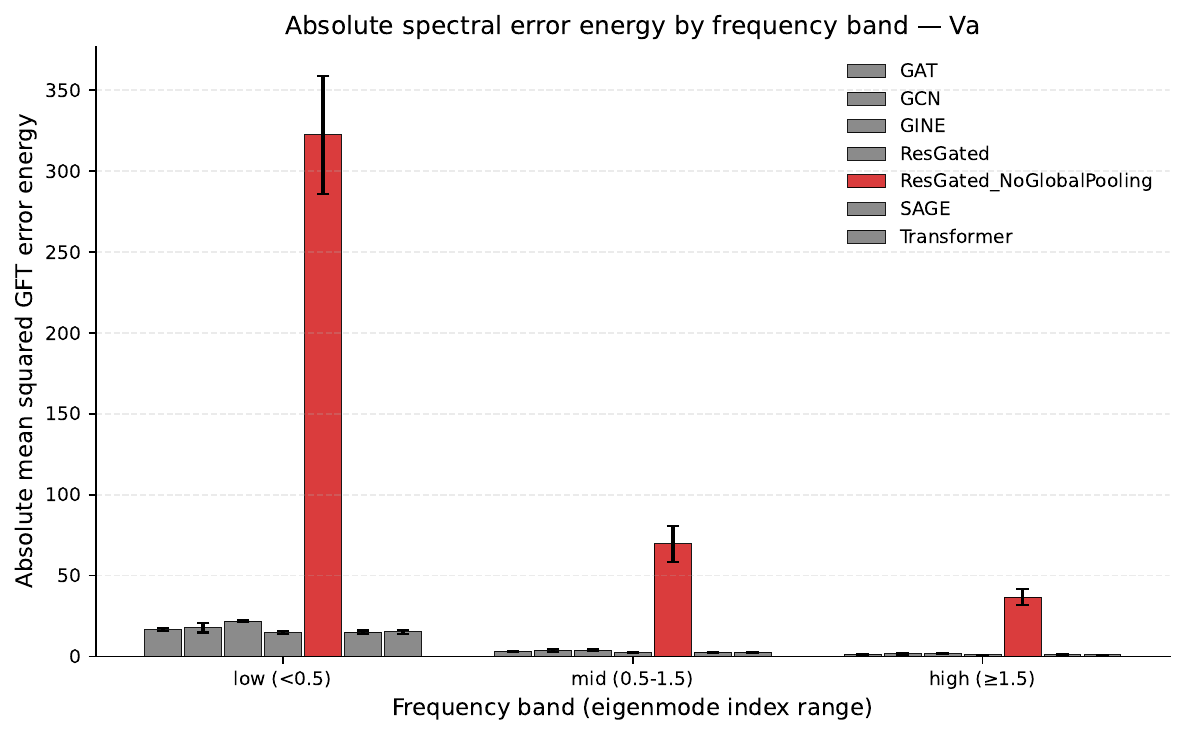}
        \caption{Voltage Angle ($V_a$)}
        \label{fig:abs_spectral_error_va}
    \end{subfigure}
    \caption{Mean squared GFT residual energy partitioned into low-, mid-, and high-frequency spectral bands for $V_m$ and $V_a$. The red bar highlights performance loss when removing global pooling from the \gls{resgated} baseline.}
    \label{fig:abs_spectral_error}
\end{figure}

As shown in Figure~\ref{fig:abs_spectral_error}, examining the absolute residual energy reveals a massive degradation in $V_a$ when global pooling is removed. This failure is most pronounced in the low- and mid-frequency bands, suggesting that fixed-depth local message passing alone cannot reliably propagate the system-wide slack bus reference or coordinate inter-area power transfers. For $V_m$, overall residual energies are substantially smaller (peaking below \(4 \times 10^{-5}\)). Nevertheless, removing global pooling still roughly quadruplicates the low- and mid-frequency error, indicating that the lack of global context impairs both the broad system voltage profile and the resolution of highly localized reactive power balances.

While the preceding discussion concerns differences \emph{between} architectures at fixed depth, the global-pooling ablation instead probes what happens \emph{within} a single architecture when its only long-range correction mechanism is removed. This spectral perspective also offers a principled lens on the oversmoothing/oversquashing trade-off inherent to fixed-depth message passing. \cite{giraldo2023trade} show analytically that both phenomena are jointly governed by the spectral gap of the graph Laplacian: a small gap between consecutive low-frequency eigenvalues permits long-range information to mix without collapsing node representations, whereas a large gap forces a trade-off in which deeper propagation needed to bridge distant clusters increasingly smooths local distinctions. The step structure visible in Figure~\ref{fig:eig_mag} is consistent with this framework, as the plateaus in $\lambda_k$ correspond to near-degenerate eigenvalue clusters separated by comparatively large gaps -- precisely the topological signature associated with bottlenecked information flow \citep{topping2021understanding, alon2020bottleneck}. This suggests that the residual error inflation observed for $V_a$ when global pooling is removed (Figure~\ref{fig:abs_spectral_error}) reflects, at least in part, a topological bottleneck intrinsic to the IEEE-118 admittance graph---consistent with its small algebraic connectivity reported in Table~\ref{tab:grid_topology}---rather than a deficiency specific to the \gls{resgated} architecture. We note, however, that this spectral-gap argument characterizes the topology's \emph{propensity} for oversquashing; it does not by itself attribute the observed error to oversquashing as opposed to model capacity or depth. Disentangling these two effects would require jointly analyzing the spectral gap against the depth-wise ($T$) error curves for each architecture -- a diagnostic we leave for future work, given the added complexity of decoupling the receptive-field depth $T$ from parameter budget under our controlled experimental protocol (Section~\ref{sec:benchmark_structure}).

\end{document}